\documentclass[3p,twocolumn]{elsarticle}

\usepackage{hyperref}

\usepackage{multirow}
\usepackage{color,soul}
\usepackage{amssymb,adjustbox}
\usepackage{amsmath,graphicx,epstopdf,siunitx,caption}

\makeatletter
\def\ps@pprintTitle{%
	\let\@oddhead\@empty
	\let\@evenhead\@empty
	\def\@oddfoot{\footnotesize\itshape Published in Pattern Recognition Letters 2020.\hfill \today}%
	\let\@evenfoot\@oddfoot}
\makeatother

\journal{Pattern Recognition Letters}

\begin{document}

\begin{frontmatter}

\title{EGO-CH: Dataset and Fundamental Tasks for Visitors Behavioral Understanding using Egocentric Vision}

\author[1,2]{Francesco Ragusa\corref{}}

\author[1]{Antonino Furnari\corref{}}

\author[1]{Sebastiano Battiato\corref{}}

\author[3]{Giovanni Signorello}

\author[1,3]{Giovanni Maria Farinella\corref{mycorrespondingauthor}}
\cortext[mycorrespondingauthor]{Corresponding author}
\ead{gfarinella@dmi.unict.it}

\address[1]{DMI-IPLab, University of Catania}
\address[2]{XGD - XENIA s.r.l., Acicastello, Catania, Italy}
\address[3]{CUTGANA, University of Catania}

\begin{abstract}
Equipping visitors of a cultural site with a wearable device allows to easily collect information about their preferences which can be exploited to improve the fruition of cultural goods with augmented reality.
Moreover, egocentric video can be processed using computer vision and machine learning to enable an automated analysis of visitors' behavior. 
The inferred information can be used both online to assist the visitor and offline to support the manager of the site. 
Despite the positive impact such technologies can have in cultural heritage, the topic is currently understudied due to the limited number of public datasets suitable to study the considered problems.
To address this issue, in this paper we propose EGOcentric-Cultural Heritage (EGO-CH), the first dataset of egocentric videos for visitors' behavior understanding in cultural sites. 
The dataset has been collected in two cultural sites and includes more than $27$ hours of video acquired by $70$ subjects, with labels for $26$ environments and over $200$ different Points of Interest. 
A large subset of the dataset, consisting of $60$ videos, is associated with surveys filled out by real visitors.
To encourage research on the topic, we propose $4$ challenging tasks (room-based localization, point of interest/object recognition, object retrieval and survey prediction) useful to understand visitors' behavior and report baseline results on the dataset.
\end{abstract}


\end{frontmatter}

\section{\uppercase{Introduction}}

Cultural sites receive many visitors every day. 
For a cultural site manager, it is hence paramount to 1) provide services able to assist the visitors, and 2) analyze their behavior to measure the performance of the site and understand what can be improved. For example using indicators \cite{EmpiricalBollo} such as: a) Attraction index: to measure how much a point of interest attracts the visitors, b) Retention index: to measure the average time spent observing information element (e.g., a caption, a video a panel, etc.), c) Sweep Rate Index (SRI): it is used to calculate if visitors move slowly or quickly through the exhibition, d) Diligent Visitor Index (DVI): the percentage of visitors who stopped in front of more than half of the points of interest.
Classic approaches addressed the former task through the delivery of printed material (e.g., maps of the museum), the use of audio-guides and the installation of informative panels. Similarly, the analysis of visitors' behavior has generally been performed through the administration of questionnaires. It should be noted that such approaches often require manual intervention and are limited especially when the number of visitors is large.
Recent works~\cite{cucchiara2014visions,ragusavedi2018,ragusasigmap2018} have highlighted that the use of wearable devices such as smart glasses can provide a convenient platform to tackle the considered tasks in an automated fashion. 
Using such technology, it is possible to provide to the user services such as automated localization (e.g., to help visitors navigating the site) and recognition of currently observed Ponts Of Interest (POIs)\footnote{In this work, we refer to the definition of Point Of Interest (POI) given in~\cite{ragusavisapp}, as an element which can attract the attention of visitors. Most POIs are objects such as paintings and statues, but architectural elements such as pavements can qualify as POIs, despite not being objects. Therefore, in this paper the notations ``Point Of Interest'' and ``object'' are not used interchangeably.} 
to provide more information on relevant objects and suggest what to see next.
Conveniently, localization and POI recognition can be used by the manager of the cultural site to obtain information about the visitors and understand their behavior by inferring where they have been, how much time they have spent in a specific environment and what POIs have been liked most.

\begin{figure}
	\centering
	\includegraphics[width=\columnwidth]{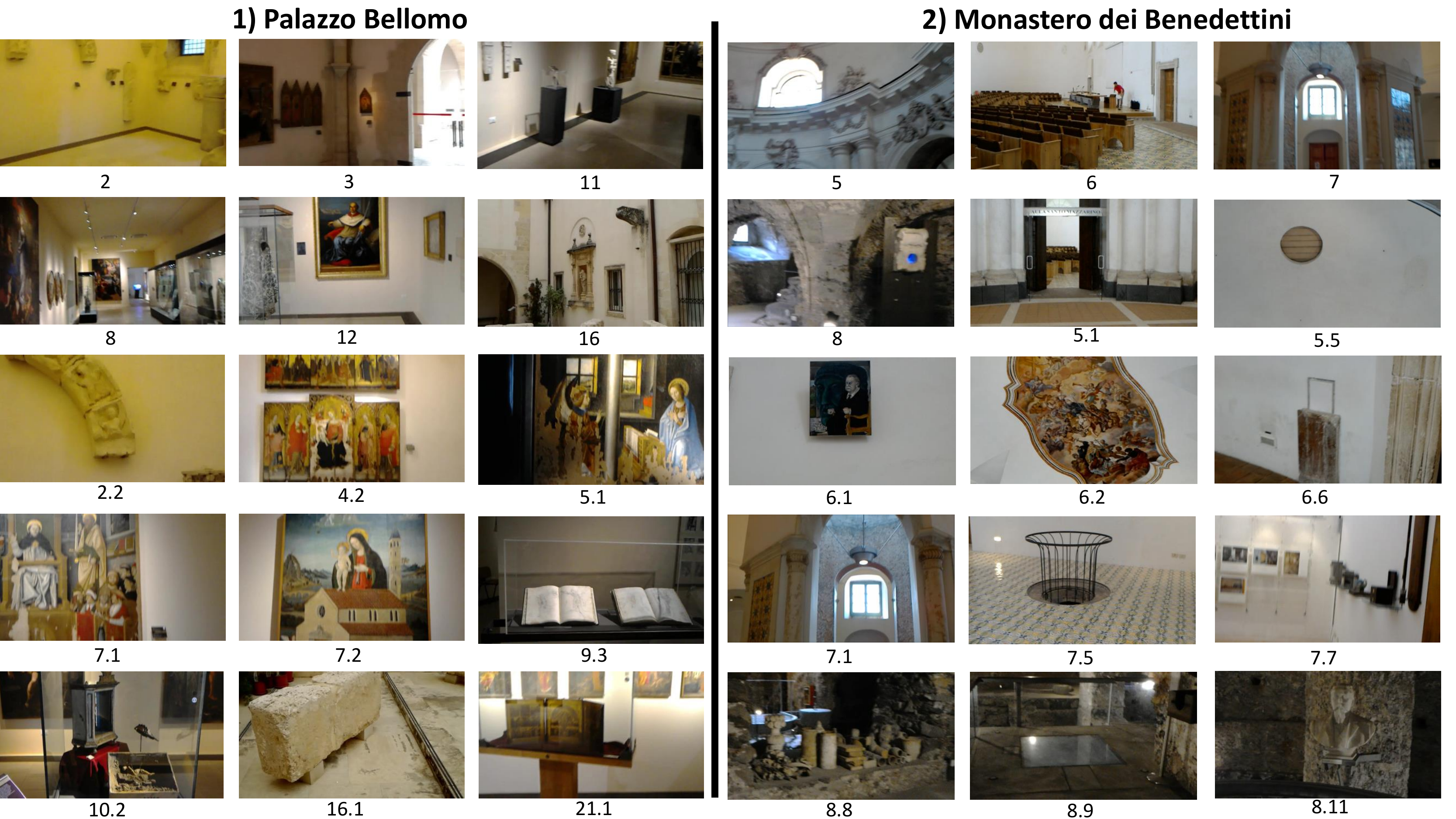}
	\caption{Sample frames from the two cultural sites belonging to  EGO-CH: 1) Palazzo Bellomo, 2) Monastero dei Benedettini. The first two rows show frames extracted from the training videos and related to the environments, whereas the remaining rows show frames  of the training videos related to POIs. See Section~\ref{sec:dataset} for more details.} \label{fig:dataset}
\end{figure}

Despite the aforementioned technologies can have a significant impact on cultural heritage, they are currently under-explored due to the lack of public benchmark datasets. 
To address this issue, in this paper we propose EGOcentric-Cultural Heritage (EGO-CH), the first large dataset of egocentric videos for visitors behavioral understanding in cultural sites. 
The dataset has been collected in two cultural sites located in Sicily, Italy: Galleria Regionale di Palazzo Bellomo\footnote{\url{http://www.regione.sicilia.it/beniculturali/palazzobellomo/}.} and Monastero dei Benedettini\footnote{\url{http://www.monasterodeibenedettini.it/}}. The overall dataset contains more than 27 hours of video, including 26 environments, over $200$ Points of Interest and $70$ visits. 
We release EGO-CH with a set of annotations useful to tackle fundamental tasks related to visitors behavior understanding in cultural sites, and specifically, temporal labels specifying the location of the visitor as well as the currently observed POI, bounding box annotations around POIs, surveys filled out by visitors at the end of each tour in the cultural site.
Figure~\ref{fig:dataset} reports some sample frames from the proposed dataset. The dataset can be publicly accessed upon request to the authors from our webpage \textit{http://iplab.dmi.unict.it/EGO-CH/}.

We  propose $4$ fundamental tasks for visitors behavioral understanding using egocentric vision: 1) \textit{room-based localization}, consisting in recognizing the environment in which the visitor is located in each frame of the video, 2) \textit{Point of Interest recognition}, which consists in correctly detecting and localizing all objects in the image frames, 3) \textit{object retrieval}, which consists in matching an observed object from the egocentric point of view to a reference  image contained in the museum catalogue of all artworks, 4) \textit{survey prediction}, which consists in generating the survey associated to a visit from video. 
We also provide baseline results for each task on the proposed dataset. The experimental results suggest that the proposed dataset is a challenging benchmark for visitors behavioral understanding using egocentric vision.

In sum, the contributions of this work are: 1) we present EGO-CH, a new challenging dataset of egocentric videos acquired in two cultural sites, 2) the dataset has been labeled to tackle $4$ main tasks useful to understand visitors behavior, 3) we report  baseline results for each task. 

\section{\uppercase{Related Work}}
\label{sec:prediction_related}
{\textbf{Visitors Behavioural Understanding and Site Manager Assistance in Cultural Sites \hspace{1mm}}
Several works investigated the use of wearable systems to augment the fruition in cultural sites \cite{cucchiara2014visions}. 
Razavian et al.~\cite{Razavian2014EstimatingAI} proposed a method to estimate the attention of the visitors of an exhibition, whereas in~\cite{seidenari2017deep} a CNN to perform localization and object recognition is introduced in order to develop a context aware audio guide.
Raptis et al.~\cite{Raptis2005} studied the design of mobile applications in museum environments and highlighted that context influences interaction. In~\cite{ragusavedi2018,ragusasigmap2018}, the problem of localizing the visitors of a museum from egocentric videos is considered. The inferred localization can be used to provide behavioral information to the manager of the site.
Past works investigated specific applications, generally relying on data collected on purpose and not publicly released. In this work, we aim at standardizing the fundamental problems of visitors behavioral understanding in cultural sites by proposing a public dataset and a series of tasks.

\textbf{Datasets on Cultural Heritage \hspace{1mm}}
Few image-based datasets focusing on cultural heritage have been proposed in past works. 
Koniusz et al.~\cite{museumexhibit2018} proposed the OpenMIC dataset containing photos captured in ten different exhibition spaces of several museums and explored the problem of artwork identification. 
DelChiaro et al. \cite{DelChiaro2019} proposed NoisyArt, a dataset composed of artwork images collected from Google Images and Flickr correlated by metadata gathered from DBpedia. 
In contrast with the aforementioned works, we propose the first dataset composed of  egocentric videos, and release it publicly. 
The dataset can be used to address different tasks related to visitors behavioral understanding in cultural sites.
A significative part of the proposed dataset has been collected by real visitors (i.e., 60 visits) and hence it is a realistic set of data for benchmarking.

\textbf{Localization \hspace{1mm}}
Ahmetovic et al.~\cite{Ahmetovic2016} presented NavCog, a system to navigate with a smartphone in complex indoor and outdoor environments exploiting Bluetooth Low Energy beacons. Kendall et al.~\cite{kendall2015posenet} proposed to infer the 6 Degrees of Freedom pose of a camera from egocentric images using a CNN. In~\cite{ragusavedi2018}, it has been considered the problem of localizing a visitor in a cultural site from egocentric images to provide behavioral information to the site manager. In this work, we consider the work presented in~\cite{ragusavedi2018} as a baseline for the localization task.

\textbf{Point Of Interest/Object Recognition \hspace{1mm}}
Seidenari et. al~\cite{seidenari2017deep} and Taverriti et al.~\cite{taverriti2016real} proposed to perform object classification and artwork recognition to assist tourists with additional information about the observed objects. In general, object detectors (e.g., YOLOv3~\cite{yolov3}) have been used to detect artworks in cultural sites. However, it should be noted that, as pointed out in~\cite{ragusavisapp}, depending on the cultural site, not all Points Of Interest are objects. For instance, a point of interest can be an architectural element such as a pavement, or even a corridor. In this case, it should be considered that object detectors can be limited.
In this work, we consider the YOLOv3 object detector~\cite{yolov3} as  baseline for Point Of Interest/Object recognition.

\textbf{Object Retrieval \hspace{1mm}}
Many previous works investigated approaches to image retrieval. 
Rubhasy et al.~\cite{Rubhasy2014ManagementAR} used an ontology-based approach to retrieval in multimedia cultural heritage collections. 
The goal is to enable the integration of different types of cultural heritage media and to retrieve relevant heritage media given a query. 
Kwan et al.~\cite{KwanRetrieval} proposed matrix of visual perspectives to address Content-based Image Retrieval (CBIR) of cultural heritage symbols, whereas Iakovidis et al.~\cite{PatternRetrieval} perform pattern-based Content-based Image Retrieval. The work of~\cite{Makantasis2016} focused on discarding image outliers using Content-based Image Retrieval. 
Despite the availability of advanced approaches, for generality and ease of comparison, in this paper we consider simple baselines based on image representation and nearest neighbor search to address the object retrieval task.

\section{\uppercase{THE EGO-CH DATASET}}
\label{sec:dataset}

\begin{figure}
	\centering
	\includegraphics[width=8cm]{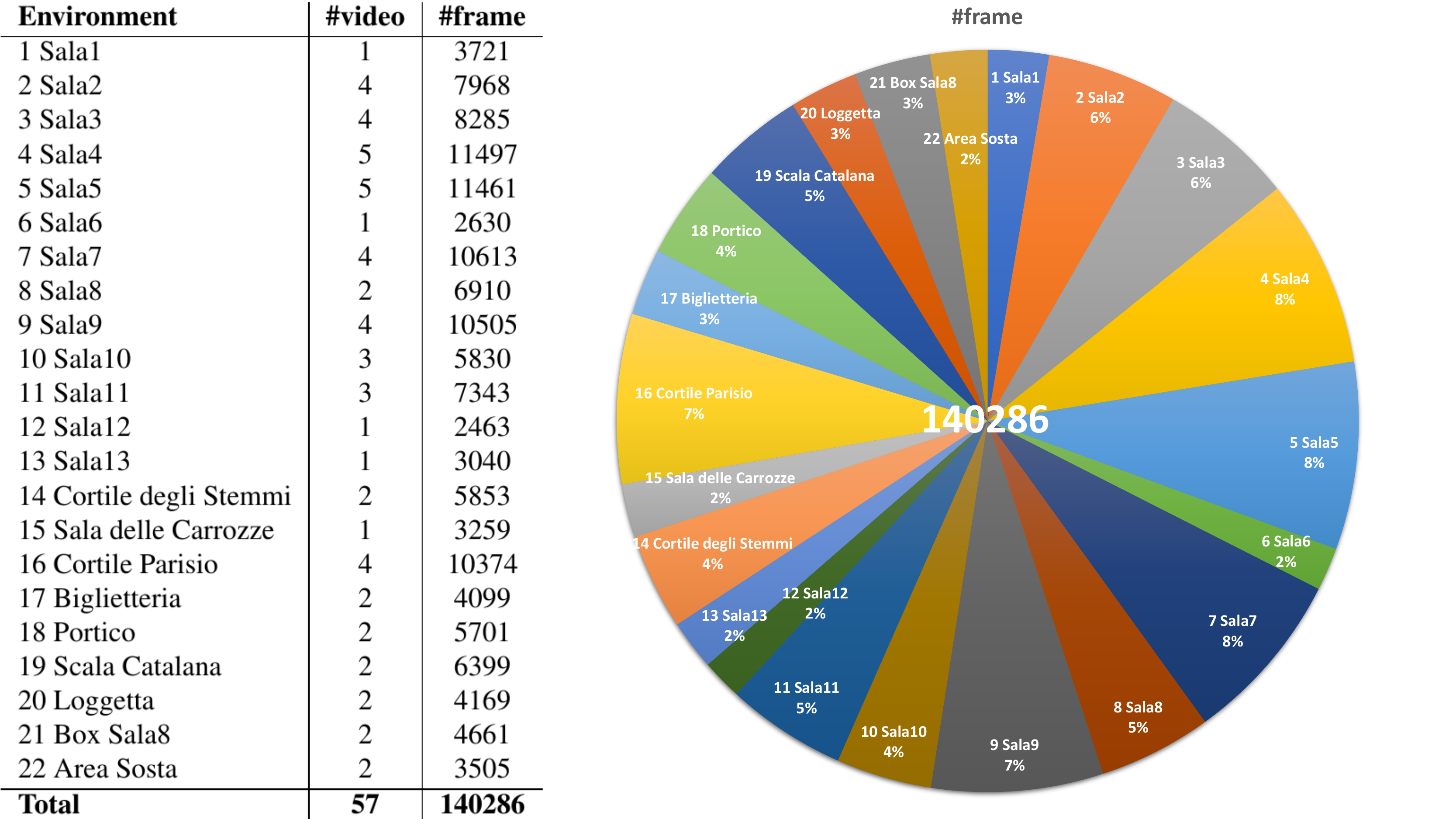}
	\caption{Number of training videos collected in each environment and corresponding number of frames for the cultural site ``Palazzo Bellomo'' (left), along with a pie chart representation of the same data (right).}
	\label{fig:bellomostats}
\end{figure}

\subsection{Data Collection}
The dataset has been acquired using a head-mounted Microsoft HoloLens device in two cultural sites located in Sicily, Italy: 1) Palazzo Bellomo (Table~\ref{tab:Bellomo_set}), located in Siracusa\footnote{http://www.regione.sicilia.it/beniculturali/palazzobellomo/}, and 2) Monastero dei Benedettini (Table~\ref{tab:Monastero_set}), located in Catania\footnote{http://monasterodeibenedettini.it/}. 

\begin{table*}[]
\caption{Details regarding the cultural site "Palazzo Bellomo".}
\label{tab:Bellomo_set}
\scriptsize
\centering
\begin{tabular}{cllccccc}
\textbf{Subset} & \multicolumn{1}{c}{\textbf{Resolution}} & \multicolumn{1}{c}{\textbf{FPS}} & \textbf{AVG Time (min)} & \textbf{\# POIs} & \textbf{\#environments} & \textbf{bbox annotations} & \textbf{temporal segments} \\ \hline
\multicolumn{1}{l}{Training} & 1280x720 & 29.97 & 1.4 & 191 & 22 & 56686 & 57 \\
Test & 1280x720 & 29.97 & 31.27 & 191 & 22 & 13402 & 340
\end{tabular}
\end{table*}

\begin{table*}[]
\centering
\caption{Details regarding the cultural site "Monastero dei Benedettini".}
\label{tab:Monastero_set}
\scriptsize
\begin{tabular}{cccccccc}
\textbf{Subset} & \textbf{Resolution} & \textbf{FPS} & \textbf{AVG Time (min)} & \textbf{\# POIs} & \textbf{\#environments} & \textbf{bbox annotations} & \textbf{temporal segments} \\ \hline
Training & 1216x684 & 24.00 & 2.2 & 35 & 4 & 33366 & 48 \\
Validation & 1216x684 & 24.00 & 3.5 & 35 & 4 & 2235 & 20 \\
Test & 1408x792 & 30.03 & 21 & 35 & 4 & 71310 & 455
\end{tabular}
\end{table*}

\textbf{Palazzo Bellomo \hspace{1mm}}
This cultural site is composed of $22$ environments and contains $191$ Points of Interest (e.g., statues, paintings, etc.).\footnote{See the supplementary material for the list of environments and POIs.} Training videos have been collected by operators instructed to walk around in order to capture images of each environment from different points of view. To simplify labeling, each training video contains only frames from a given environment. At least one training video has been collected per environment. In the case of outdoor environments (e.g., courtyards), we collected multiple videos to include different lighting conditions. We have  collected a total of $57$ training video in this cultural site.
Figure~\ref{fig:dataset}(left) shows some frames acquired in the considered cultural site, whereas Figure~\ref{fig:bellomostats} reports the number/percentage of frames acquired in each environment. Ten
 test videos have been collected separately asking $10$ volunteers to visit the cultural site. One of the $10$ videos (i.e., ``Test 3'') was selected randomly and used as validation set, whereas the remaining $9$ videos are used for evaluation purposes. No specific instructions on where to go, what to look at and how much time to spend in a specific environment/POI has been provided to the visitors. Most of the subjects had limited confidence with the cultural site. This provided a natural means to collect realistic data of visitors exploring the environments and observing Points of Interest.
All the videos have a resolution of $1280 \times 720$ pixels and a frame-rate of $29.97$ fps. The average duration of test videos is $31.27\ min$, with the longest one being $50.23\ min$. See the supplementary material for more details about training/test videos.
We also include $191$ reference images related to the considered POIs to be used for one-shot image retrieval. The images are akin to the images generally included in museum catalogs.\footnote{\label{HD}Examples reference images for both cultural sites are included in the supplementary material.}

	\begin{figure}[t]
	\centering
	\includegraphics[width=8cm]{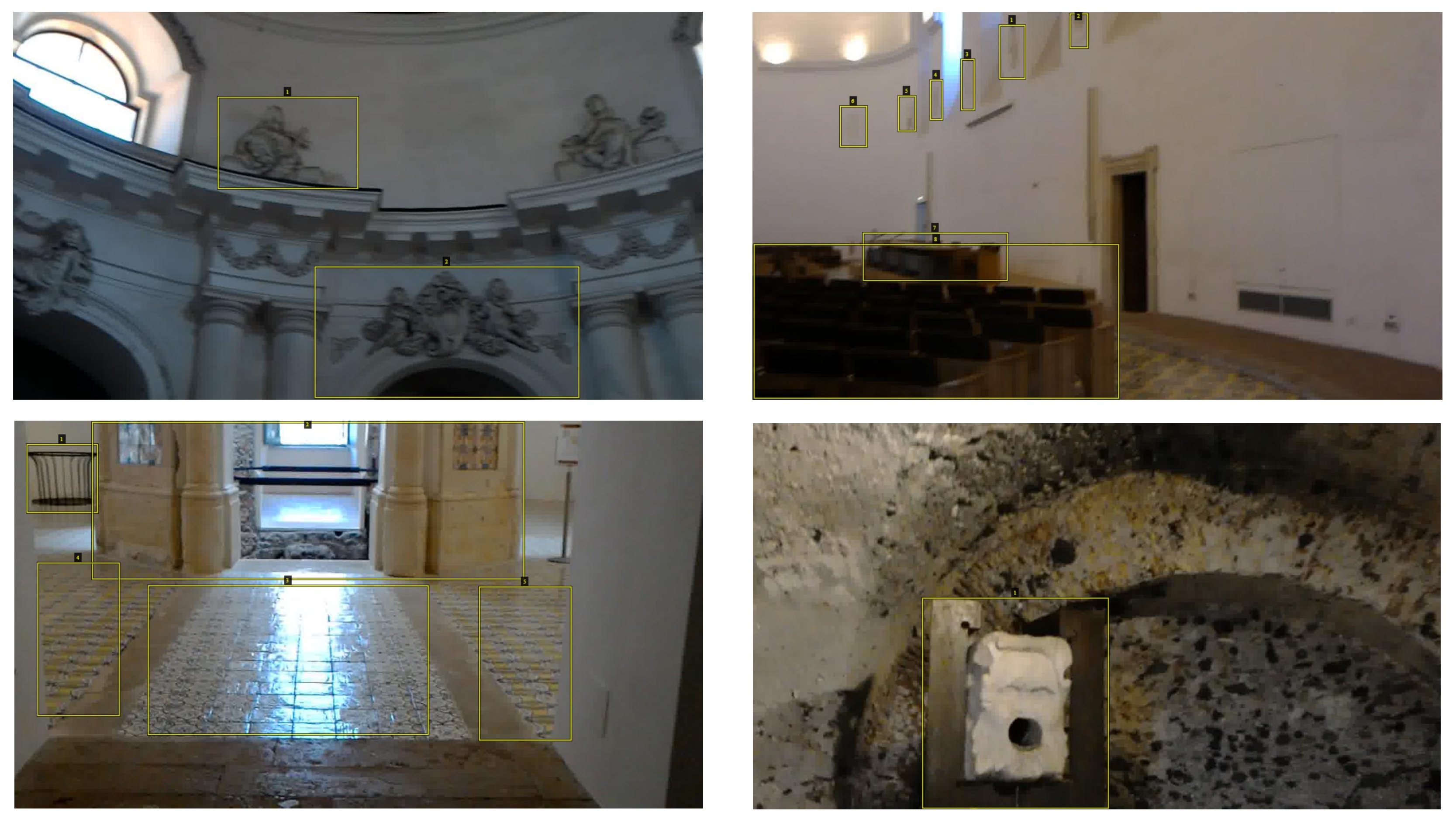}
	\caption{Some example bounding box annotations from the cultural site ``Monastero dei Benedettini''.} \label{fig:bboxann}
\end{figure}
\textbf{Monastero dei Benedettini \hspace{1mm}}
This dataset is composed of $4$ environments and contains $35$ Points Of Interest.\footnote{See the supplementary material for the list of environments and POIs.}
Differently from ``Palazzo Bellomo'', the POIs belonging to this cultural site include both objects such as paintings and statues as well as architectural elements, such as pavements, which cannot be easily recognized using object detection techniques as noted in~\cite{ragusavisapp}. See Figure~\ref{fig:dataset}(right) for some qualitative examples of the considered objects.
Training videos have been collected with the same acquisition modality considered for the ``Palazzo Bellomo'' cultural site. Figure~\ref{fig:montrainstats} reports the number/percentage of frames acquired in each environment.
Training and validation videos have a resolution of $1216\times 684$ pixels and a frame-rate of $24$ fps. Five validation videos have been collected by asking volunteers to visit the cultural site following the same protocol used for ``Palazzo Bellomo''.
Additionally, we collected $60$ test videos by asking real visitors inexperienced with both the research project and its goals and the HoloLens device to freely visit the cultural site. No specific instructions have been given to the visitors, who were free to explore the $4$ environments and the $35$ POIs.
This allowed us to obtain realistic data of how a visitor would move in a cultural site.
Test videos have been collected over a period of three months.
Moreover, at the end of the visit, we administered the visitor a survey, the content of which is described in Section~\ref{sec:data_annotations}. 
The $60$ test videos have a resolution of $1408 \times 792$ pixels and a frame-rate of $30.03\ fps$. The average video length is $21\ min$, with the maximum length being $42\ min$.
See the supplementary material for more details about training/validation/test videos.
Similarly to ``Palazzo Bellomo'', we include $35$ reference images related to the considered POIs for one-shot image retrieval\footnotemark[\getrefnumber{HD}].
Please note that this set of data is adapted from and extends significantly the dataset proposed in~\cite{ragusavedi2018}, introducing $60$ new labelled videos collected by real visitors. Specifically, the overall dataset presented in this work contains $+1600$ minutes of video, data from $+70$ more subjects, $+91369$ bounding box annotations and an additional cultural site ``Palazzo Bellomo'' comprising $22$ environments and $191$ points of interest.

\begin{figure}
\centering
\includegraphics[width=8cm]{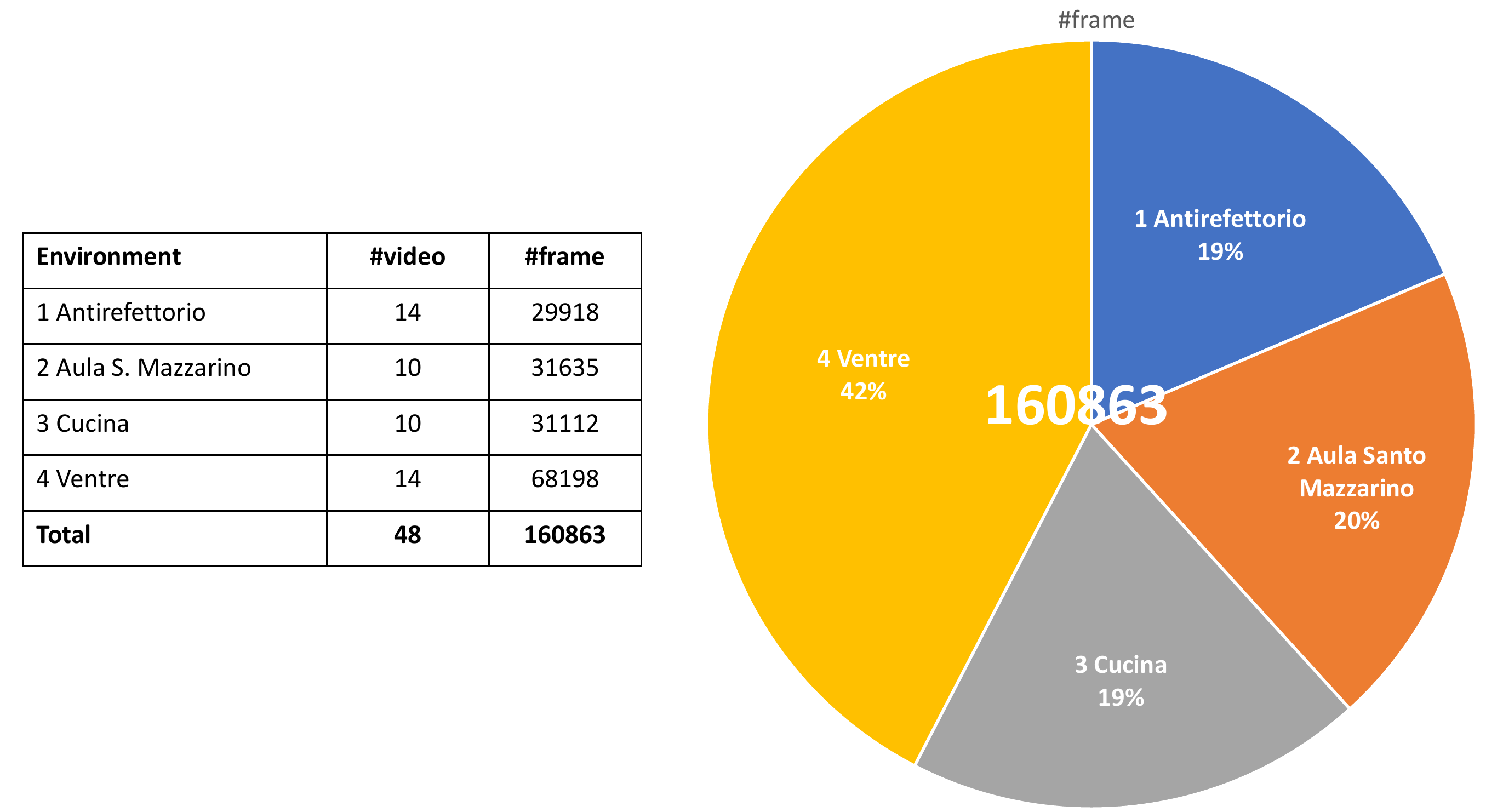}
\caption{Number of training videos collected in each environment and corresponding number of frames for the cultural site ``Monastero dei Benedettini'' (left), along with a pie chart representation of the same data (right).}
\label{fig:montrainstats}
\end{figure}

\subsection{Annotations} 
\label{sec:data_annotations}
\textbf{Temporal Labels \hspace{1mm}}
All test and validation videos have been temporally labeled to indicate in every frame the environment in which the visitor is located and the observed point of interest, if any.
If the visitor is not located in one of the considered environment (e.g., a stair), the frame is marked as  ``negative"\footnote{\label{negatives} Examples of ``negative" frames are reported in the supplementary material.}.
It is worth noting that there are no negative frames in ``Palazzo Bellomo'' since all environments are part of the museum, whereas negative frames are contained in ``Monastero dei Benedettini''.
This is due to the different nature of the two sites: ``Palazzo Bellomo'' is a museum, consisting in a limited set of rooms, whereas ``Monastero dei Benedettini'' is a much more complex environment including many corridors and stairs which have not been labeled as locations of interest for visitors.
Similarly, we mark as ``negative'' all frames in which the visitor is not observing any of the considered POIs.
Each location is identified by a number that denotes a specific environment ($1-22$ for ``Palazzo Bellomo'' and $1-4$ for ``Monastero dei Benedettini''). Each point of interest is denoted by a code in the form X.Y (e.g., 3.5) where ``X'' denotes the environment in which the point of interest is located and ``Y'' identifies the point of interest. See Figure~\ref{fig:dataset} for some examples. 
\textbf{Bounding Box Annotations \hspace{1mm}}
A subset of frames from the dataset (sampled at at $1$ fps) has been labeled with bounding boxes indicating the presence and locations of all POIs. 
Specifically, each POI has been labeled with a tuple $(class, x, y, w, h)$ indicating the class of the POI and its bounding box information. 
It is worth mentioning that, as noted in~\cite{ragusavisapp}, a POI can be an object (e.g., a painting or a statue) or a different element (e.g., a pavement or a specific location), which cannot be strictly defined as an object.
Indeed, the kind of POIs contained in a cultural site depends on the nature of the site itself.
In EGO-CH, ``Palazzo Bellomo'' contains only objects as POIs, whereas ``Monastero dei Benedettini'' contains both objects and other elements.
Nevertheless, all elements are labeled with class type and bounding box annotations.
Figure~\ref{fig:bboxann} shows examples of labeled frames from the $60$ visits of ``Monastero dei Benedettini''. 
\textbf{Surveys \hspace{1mm}}
The $60$ test videos collected in the ``Monastero dei Benedettini'' are associated with surveys which have been administered to the visitors at the end of the visits.
Specifically, the visitors are asked to rate a subset of $33$ out of the $35$ Points Of Interest (a picture of each point is shown) or specify if any of them had not been seen it during the visit. 
The rating is expressed as a number ranging from $-7$ (not liked) to $+7$ (liked). %

The EGO-CH dataset is publicy available at our website: \textit{http://iplab.dmi.unict.it/EGO-CH/}. The reader is referred to the supplementary material for more details about the dataset and the experiments. 
The dataset can be used only for research purposes and is available upon the acceptance of an agreement.

\section{\uppercase{Proposed Tasks and Baselines}}
\label{sec:challenges}
In this Section, we propose four tasks which can be addressed using the proposed dataset. The tasks are related to problems investigated in previous works on cultural heritage~\cite{ragusasigmap2018,ragusavedi2018,seidenari2017deep,taverriti2016real}. We believe that solving these tasks can bring useful information about the behavior of the visitors of a cultural site.

\subsection{Room-based Localization}
\textbf{Task:} The task consists in determining the room in which the visitor of a cultural site is located from egocentric images collected using a wearable device. Localization information can be used both to provide a ``where am I'' service to the visitor and to collect behavioral information useful for the site manager to understand what paths do visitors prefer and where they spend more time in the cultural site. 

\textbf{Baseline:} As a baseline for this task, we consider the approach proposed in~\cite{ragusavedi2018,furnari2018personal}. This approach is selected as a baseline due to the limited work on room-based localization in the cultural heritage domain~\cite{ragusavedi2018} and due to the state-of-the-art performance of the approach shown in~\cite{furnari2018personal}.
Given a set of locations, the considered approach allows to segment a given video into video shots related to the specified locations.
If a given shot is not related to any of the locations, the algorithm automatically labels it as a ``negative segment'' through a ``negative rejection'' stage.
The method is composed by three steps, as illustrated in Figure~\ref{fig:loc}. For each cultural site, we trained a VGG-19 CNN to discriminate between locations (``Discrimination'' stage). 
The ``Negative Rejection" step has been considered only for the data of ``Monastero dei Benedettini'', since ``Palazzo Bellomo'' does not contain negative locations. The ``Sequential Modeling'' stage allows to obtain a temporal segmentation of the input video where each segment is associated to one of the considered environments.  
This algorithm is chosen as it achieves state-of-the-art performance in the task of location-based egocentric video segmentation~\cite{furnari2018personal,ragusavedi2018}.
Two hyper-parameters are involved in the algorithm: $K$, related to the ``negative rejection'' stage and $\epsilon$, which regulates the amount of temporal smoothing applied to the predictions.
The reader is referred to~\cite{ragusavedi2018} for more details.
	\begin{figure}
	\centering
	\includegraphics[width=8cm]{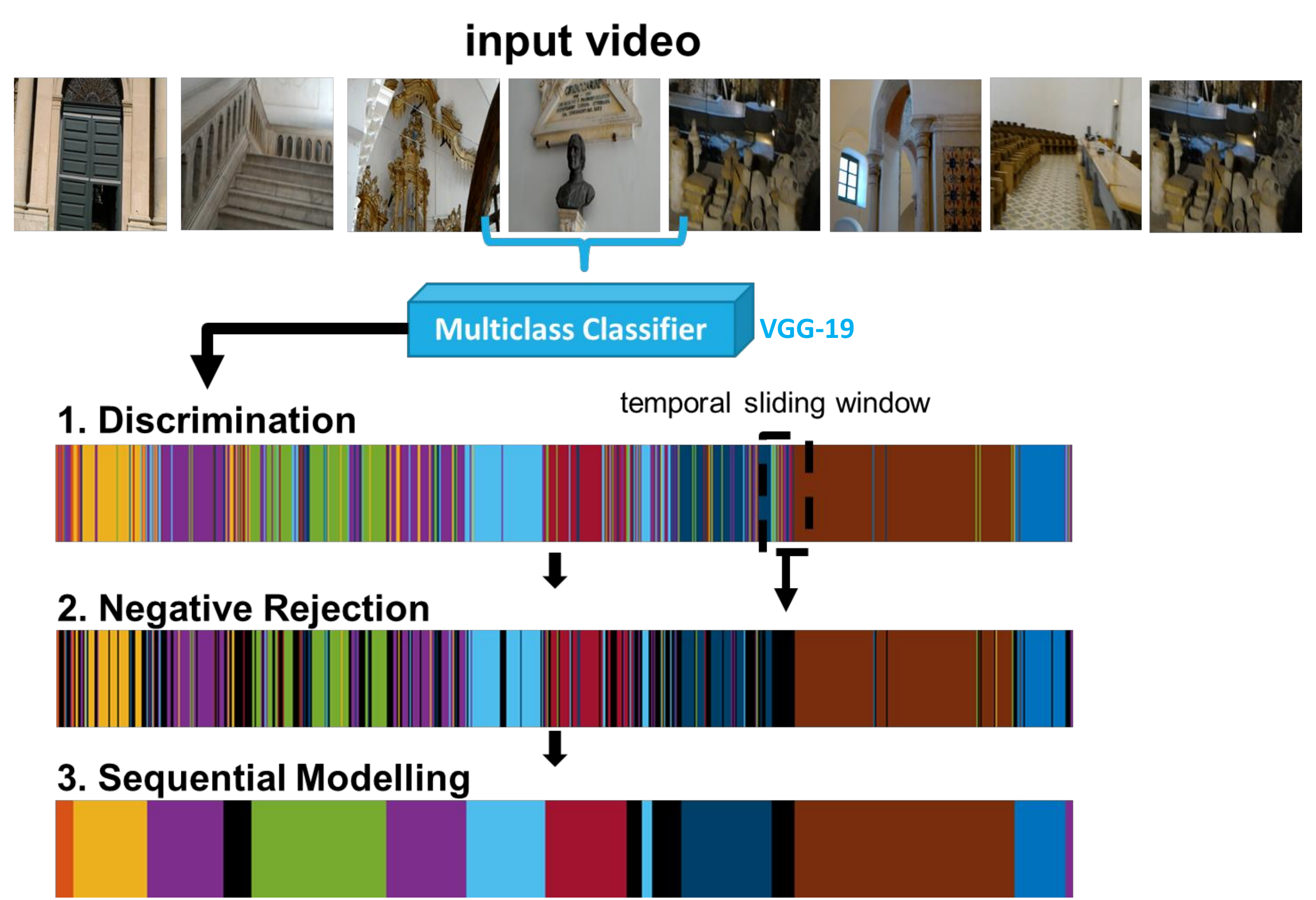}
	\caption{The method used to perform room-based localization. The method is composed by three steps: 1) Discrimination, 2) Negative Rejection, 3) Sequential Modeling. See~\cite{furnari2018personal,ragusavedi2018} for more details.} \label{fig:loc}
\end{figure}

\textbf{Implementation Details and Evaluation Measures:} We evaluated our method following~\cite{ragusavedi2018} using $FF_1$ score and $ASF_1$ score. Specifically, the $FF_1$ score is the $F_1$ score applied to individual frames and, as such, it does not evaluate the ability of the methods to produce a temporally coherent segmentation. $ASF_1$ is the $F_1$ score applied to temporal segments rather than frames and measures the ability to detect video segments coherent with the ground truth. Both scores are normalized between $0$ and $1$. The hyper-parameters of the algorithm $K$ and $\epsilon$ are tuned on the validation sets of the proposed dataset. Specifically, $\epsilon=10^{-273}$ is found by optimizing the validation $ASF_1$ score with a grid search in the range \begin{math}[10^{-1} : 10^{-299}]\end{math} on ``Palazzo Bellomo'' (see \cite{ragusavedi2018} for details). Since no negative locations are contained in ``Palazzo Bellomo'', the ``negative rejection'' stage is not performed and hence the parameter $K$ is not optimized. Similarly, we find $\epsilon = 10^{-89}$ and $K = 100$ on ``Monastero dei Benedettini''.\footnote{The supplementary material reports more implementation details.}

\textbf{Results:} 
\tablename~\ref{tab:results1} reports the results obtained by the baseline in the two cultural sites\footnote{Extended tables, qualitative results and confusion matrix are included in the supplementary material.}.
On ``Palazzo Bellomo'', the baseline achieves good $FF_1$ scores for most rooms, obtaining an average value of $0.81$. Much lower results are observed when the $ASF_1$ score is considered. In this case, an average value of $0.59$ is reached. Lower results equal to $0.68$ and $0.40$ are obtained in the ``Monastero dei Bendettini''. This is partly due to the presence of negatives, which are not included in ``Palazzo Bellomo'' and to the more challenging nature of the test set of ``Monastero dei Benedettini'', which contains $60$ videos collected by real visitors within 3 months with different lighting condition and blur as shown in Figure~\ref{fig:lights}. The overall results highlight that addressing the considered task on the proposed dataset is challenging. In particular, issues such as varying lighting conditions and the presence of negatives need to be addressed in task-specific investigations.

\begin{table}[t]
	\centering
	\caption{Room-based localization results.For each cultural site, the last row reports the Average (AVG) of the $FF_1$ and $ASF_1$ scores.}
	\label{tab:results1}
	\scriptsize
	\centering
	\begin{tabular}{lcc}
		& \multicolumn{1}{l}{}                  & \multicolumn{1}{l}{}                   \\
		\multicolumn{3}{c}{\textbf{1) Palazzo Bellomo}}                                                                  \\ \hline
		\textbf{Room}       & \multicolumn{1}{l}{\textbf{$FF_1$ score}} & \multicolumn{1}{l}{\textbf{$ASF_1$ score}} \\ \hline
		Sala1                & 0.71                                  & 0.48                                   \\
		Sala2                & 0.92                                  & 0.79                                   \\
		Sala3                & 0.84                                  & 0.50                                    \\
		Sala4                & 0.92                                  & 0.59                                   \\
		Sala5                & 0.94                                  & 0.64                                   \\
		Sala6                & 0.77                                  & 0.52                                   \\
		Sala7                & 0.94                                  & 0.61                                   \\
		Sala8                & 0.89                                  & 0.64                                   \\
		Sala9                & 0.91                                  & 0.47                                   \\
		Sala10               & 0.84                                  & 0.69                                   \\
		Sala11               & 0.84                                  & 0.58                                   \\
		Sala12               & 0.80                                  & 0.66                                   \\
		Sala13               & 0.80                                  & 0.66                                   \\
		Cortile degli Stemmi & 0.85                                  & 0.64                                   \\
		Sala Carrozze        & 0.91                                  & 0.67                                   \\
		Cortile Parisio      & 0.75                                  & 0.50                                   \\
		Biglietteria         & 0.65                                  & 0.44                                   \\
		Portico              & 0.69                                  & 0.51                                   \\
		Scala Catalana       & 0.76                                  & 0.63                                   \\
		Loggetta             & 0.71                                  & 0.51                                   \\
		Box Sala8            & 0.94                                  & 0.79                                   \\
		Area Sosta           & 0.43                                  & 0.47                                   \\ \hline
		\textbf{AVG}         & \textbf{0.81}                         & \textbf{0.59}                          \\ \hline
		& \multicolumn{1}{l}{}                  & \multicolumn{1}{l}{}                   \\
		\multicolumn{3}{c}{\textbf{2) Monastero dei Benedettini}}                                                                \\ \hline
		\textbf{Class}       & \textbf{$FF_1$ score}                     & \multicolumn{1}{l}{\textbf{$ASF_1$ score}} \\ \hline
		Antirefettorio       & 0.75                                  & 0.54                                   \\
		Aula S. Mazzarino    & 0.33                                  & 0.12                                   \\
		Cucina               & 0.79                                  & 0.34                                   \\
		Ventre               & 0.97                                  & 0.60                                    \\
		Negative             & 0.54                                  & 0.33                                   \\ \hline
		\textbf{AVG}         & \textbf{0.68}                         & \textbf{0.40}                          \\ \hline
		
	\end{tabular}
\end{table}

	\begin{figure}
	\centering
	\includegraphics[width=8cm]{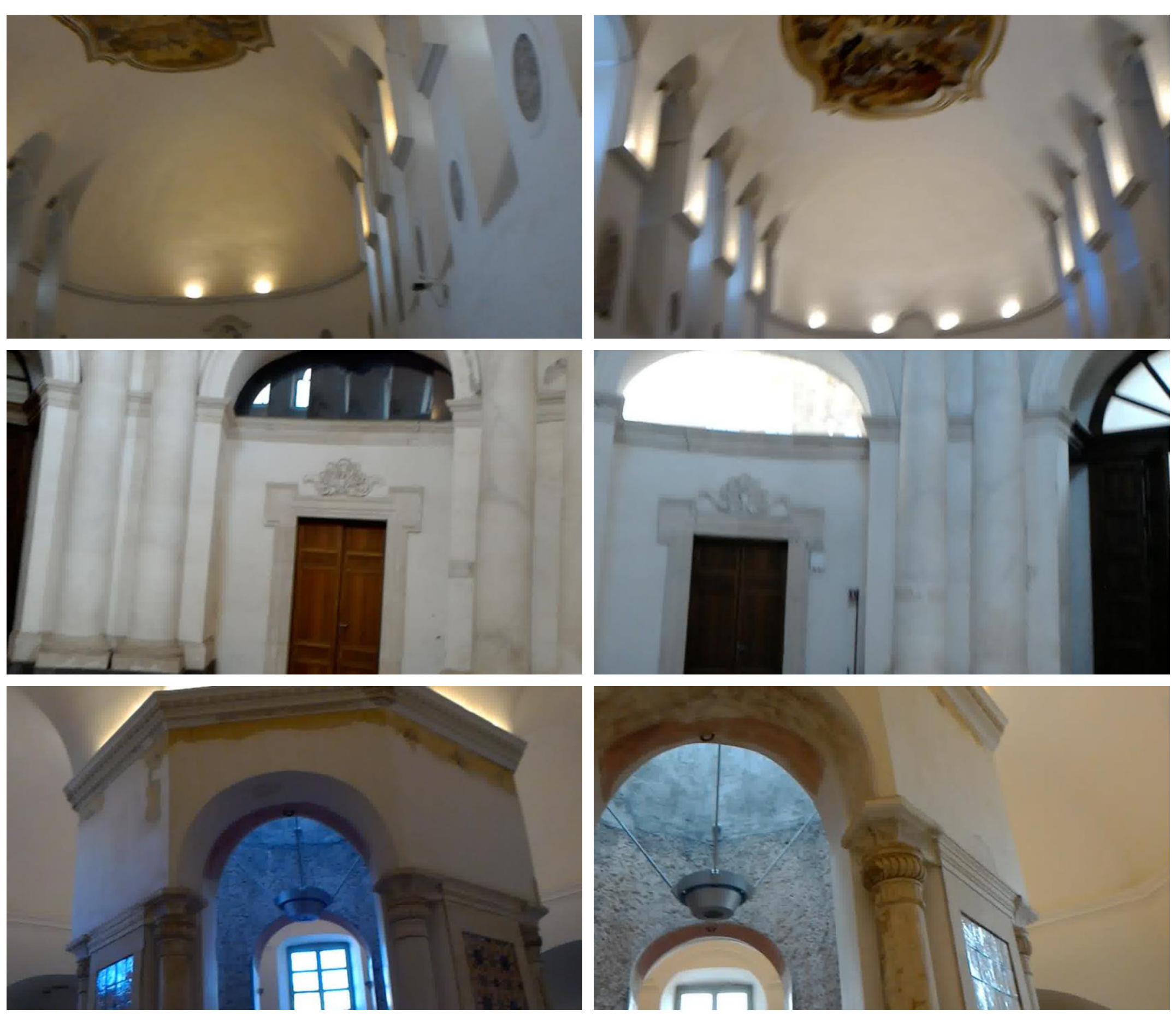}
	\caption{Some sample frames from different visits acquired within 3 months. Each row represents similar positions in the same environment with different lighting conditions.} \label{fig:lights}
\end{figure}

\subsection{Point of Interest/Object Recognition}
\textbf{Task:} This tasks consists in recognizing the points of interest which the user is looking at. 
This can be useful to understand the visitor's behavior and answer questions as ``What are the most viewed points of interest?'' and ``How long have they been observed?''. 
Moreover, a system able to recognize points of interest could suggest the visitor what to see next, as well as provide information with Augmented Reality. {The dataset could be used to perform standard object detection task.

\textbf{Baseline:} Due to its real-time performance and to its popularity in the cultural heritage domain~\cite{ragusavisapp,seidenari2017deep,taverriti2016real}, we consider a YOLOv3~\cite{yolov3} object detector as a baseline for the task. The detector has been trained on the training sets of ``Palazzo Bellomo'' and ``Monastero dei Benedettini''.

\textbf{Implementation Details and Evaluation Metrics:} We trained YOLOv3 using the standard anchors provided by the authors for the COCO dataset. We use mean Average Precision (mAP) with threshold on IoU equal to $0.5$ for the evaluations. 
In order to use YOLOv3 to detect artworks, a detection threshold is specified to discard detections with low confidence scores. 
For each cultural site, we tuned this threshold on the validation sets by choosing the value which maximizes mAP in the range \begin{math}[5^{-4};1^{-3};5^{-3};1^{-2};3^{-2};5^{-2};0.10;0.15;0.2;0.25;0.3;\\0.35;0.40]\end{math}. 
To train the detector on ``Palazzo Bellomo'', we set the initial learning rate to $0.001$ and the detection threshold to $0.01$. 
On ``Monastero dei Benedettini'', we set the initial learning rate to $0.01$ and the detection threshold to $0.001$.

\textbf{Results:} Table~\ref{tab:results2} reports the results obtained in the two cultural sites. The results obtained on ``Palazzo Bellomo'' are much lower than the ones obtained on ``Monastero dei Benedettini'' mainly because of the larger set of POIs contained in the former site ($191$) versus the lower number of POIs contained in the latter ($35$). 
In both cases, the results are in general very low, which highlights the challenging nature of the proposed dataset and tasks. Among the challenges of the dataset, as previously discussed, it should be considered that some of the points of interest represent architectural elements such as corridors or pavements, which might be challenging to detect with a simple object detector, as pointed out in~\cite{ragusavisapp}. Moreover, differently from other object detection tasks, POIs here need to be recognized at the instance level. For instance, the dataset contains multiple paintings which should be recognized as separate objects. We leave the investigation of more specific approaches to future investigations.

\begin{figure}[t]
	\centering
	\includegraphics[width=7cm]{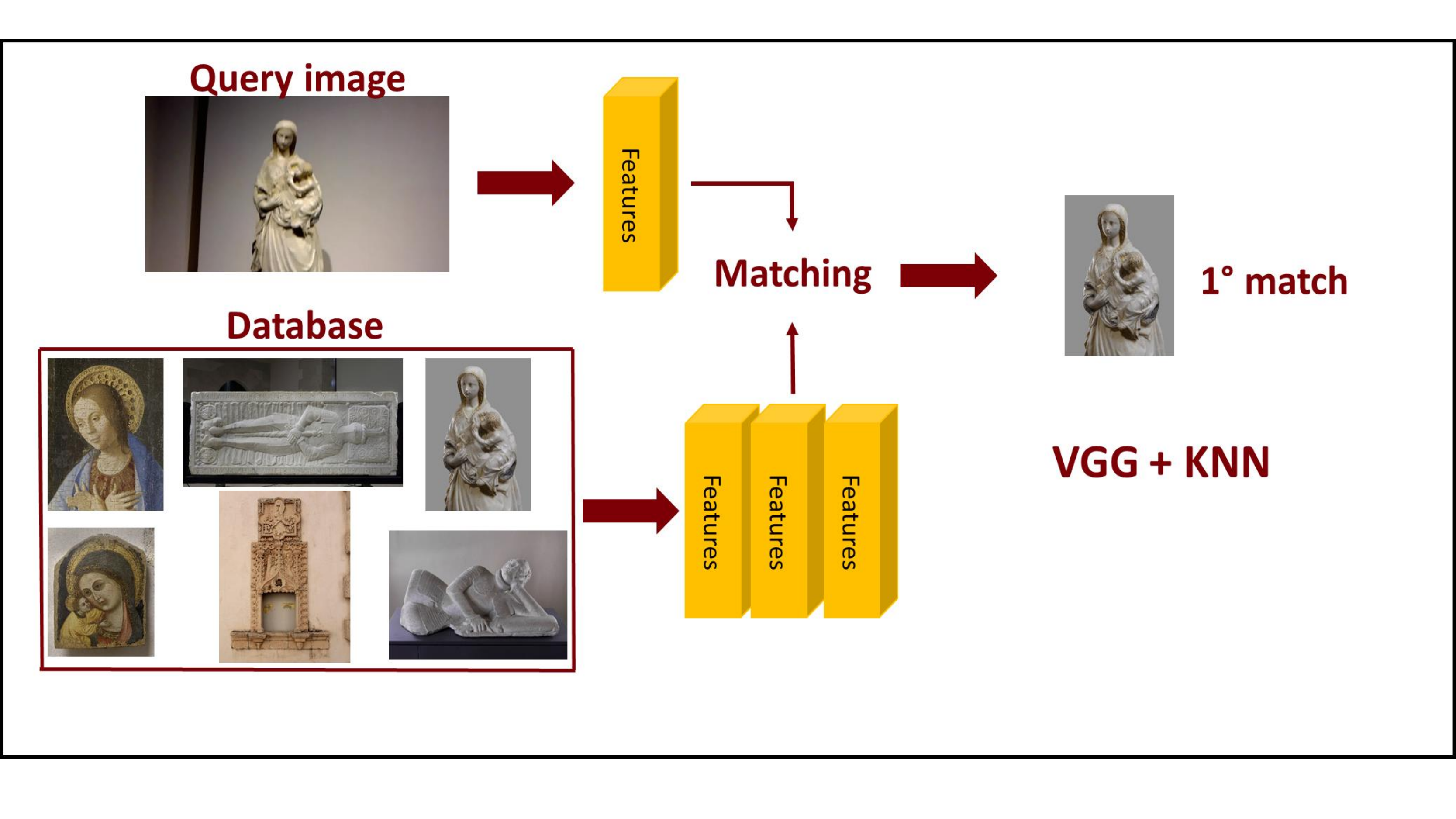}
	\caption{Diagram of the baseline for the object retrieval task.} \label{fig:3_task}
\end{figure}

\begin{table}[t]
	\caption{Object detection results. The reported mean Average Precision (mAP) is averaged over all test videos. Per-class Average Precision (AP) values are reported in the supplementary material.}
	\label{tab:results2}
	\centering
	\scriptsize
	\begin{tabular}{lc}
		\multicolumn{2}{l}{}                                             \\ \hline
		\multicolumn{1}{c|}{\textbf{Cultural Site}}       & \textbf{mAP} \\ \hline
		\multicolumn{1}{l|}{1) Palazzo Bellomo}           & 10.59\%           \\
		\multicolumn{1}{l|}{2) Monastero dei Benedettini} & 15.45\%     
	\end{tabular}
\end{table}
\subsection{Object Retrieval}
\textbf{Task:} Given a query image containing an object, the task consists in retrieving an image of the same object from a database. 
This task can be useful to perform automatic recognition of artworks when detection can be bypassed, i.e., when the user places the artwork in the center of the field of view using a wearable or mobile device.
Moreover, the task is particularly of interest especially considering that artwork detection is a hard task, as highlighted in the previous section. 
We obtain a set of query images by extracting image patches from the bounding boxes annotated in the test set and consider two variants of the task.
This accounts to \textit{23727} image patches for ``Palazzo Bellomo'' and \textit{44978} image patches for ``Monastero dei Benedettini''.\footnote{\label{patches}The supplementary material reports examples of extracted image patches.} %
We consider two variants of this task. In the first variant, object retrieval is framed as a one-shot retrieval problem.
In this case, the database contains only the reference images associated to each POI, whereas the whole set of image patches is used as the test set, i.e., only a single labeled sample is assumed to be available for each object. 
In the second variant, we split the set of image patches into a training set ($70\%$ - used as DB) and a test set ($30\%$). 
It should be noted that the first variant of the task is much more challenging both due to the presence of few labeled samples and to the domain shift which affects the two sets of images: reference images for the POIs and image patches cropped from egocentric images. \figurename~\ref{fig:patch} shows an example of image patches cropped from the egocentric images using bounding box annotations.

\begin{figure}[t]
	\centering
	\includegraphics[width=7cm]{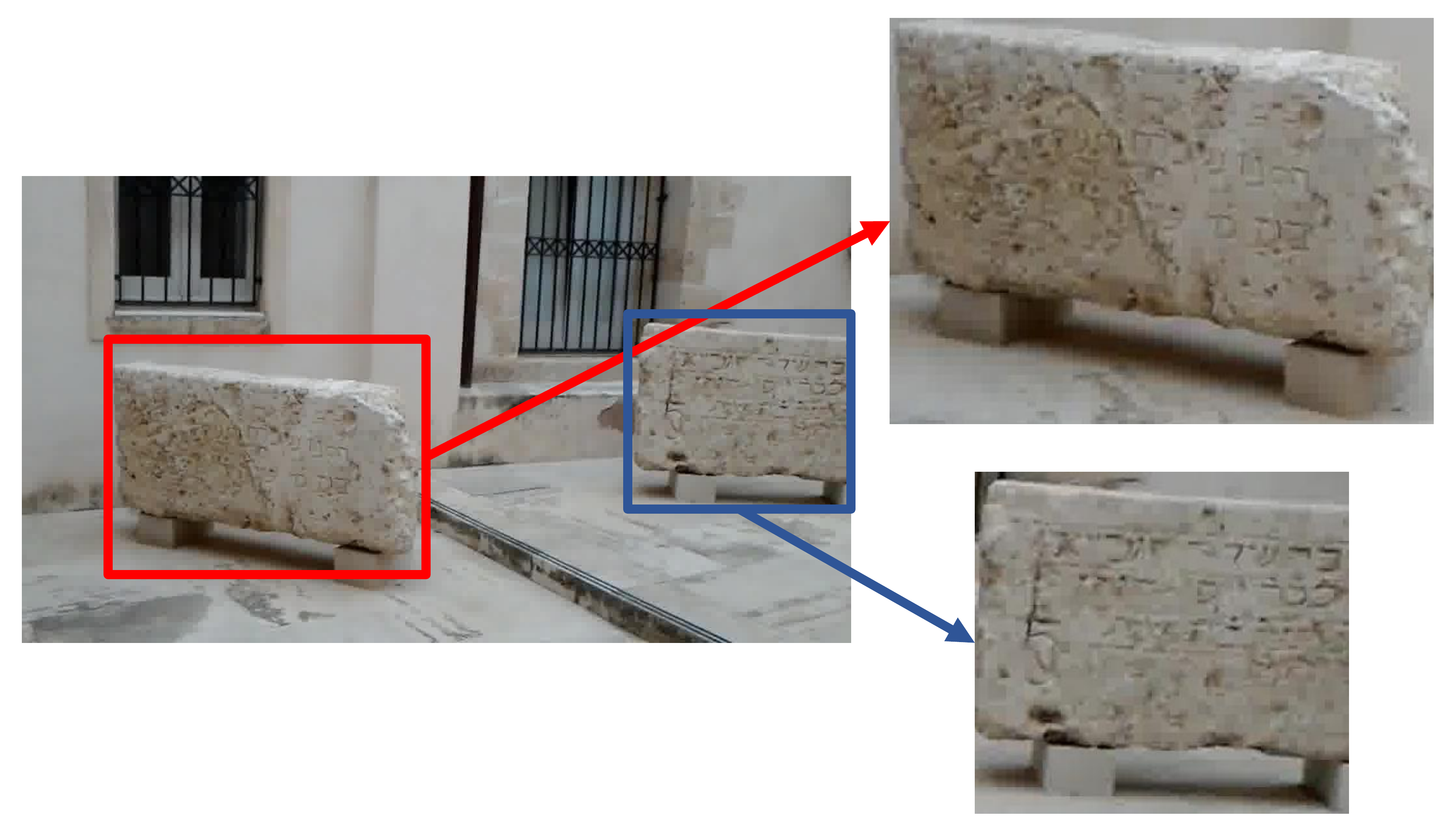}
	\caption{Example of patches extracted using bounding boxes annotations.} 
	\label{fig:patch}
\end{figure}

\textbf{Baseline}:
Given the lack of investigation of approaches for retrieval in the scenario of First-Person vision in the cultural heritage domain, we consider a simple image-retrieval pipeline for both variants of the task.
The pipeline uses VGG19 CNN pre-trained on ImageNet to represent image patches, while matching is perfor and matched using a K-NN. A scheme of the considered baseline is shown in Figure~\ref{fig:3_task}.

\textbf{Implementation Details and Evaluation Measures:} 
We have extracted all features from the FC7 layer of the VGG19 network pre-trained on ImageNet. %
When the second variant of the task is considered, we perform K-NN using $K = [1;3;5]$. We evaluated the performance of our baseline using standard metrics for image-retrival: precision, recall and $F_1$ score.

\textbf{Results:} Table~\ref{tab:results3} shows the results of the baseline on the image retrieval variants. In both cultural sites, one-shot retrieval does not achieve good results. This is probably due to the fact that one-shot retrieval relies on a limited number of training samples, which are drawn from a different distribution as compared to test samples. This suggests that dedicated methodologies should be considered to tackle one-shot retrieval and the domain shift problem.
Better results are obtained on both sites in the second variant of the task, when the effect of one-shot retrieval and domain shift is reduced.
Best results are obtained in ``Palazzo Bellomo'' for $K=1$ ($F_1$ score of $0.67$) and in ``Monastero dei Benedettini'' for $K=5$ ($F_1$ score of $0.88$).

\begin{table}[t]
	\caption{Object retrieval results for the two variant of the task.}
	\label{tab:results3}
	\scriptsize
	\centering
\begin{tabular}{ccccc}
\multicolumn{5}{c}{\textbf{Points of Interest Retrieval}} \\ \hline
\multicolumn{1}{l}{} &  & \multicolumn{1}{l}{} & \multicolumn{1}{l}{} & \multicolumn{1}{l}{} \\
\multicolumn{5}{c}{\textbf{1) Palazzo Bellomo}} \\ \hline
\multicolumn{1}{c|}{\textbf{Variant}} & \multicolumn{1}{c|}{\textbf{K}} & \multicolumn{1}{c|}{\textbf{Precision}} & \multicolumn{1}{c|}{\textbf{Recall}} & \textbf{$\mathbf{F_1}$ score} \\ \hline
\multicolumn{1}{c|}{1 - One Shot} & \multicolumn{1}{c|}{1} & \multicolumn{1}{c|}{0.004} & \multicolumn{1}{c|}{0.007} & 0.001 \\ \hline
\multicolumn{1}{c|}{} & \multicolumn{1}{c|}{1} & \multicolumn{1}{c|}{0.69} & \multicolumn{1}{c|}{0.66} & \textbf{0.67} \\
\multicolumn{1}{c|}{} & \multicolumn{1}{c|}{3} & \multicolumn{1}{c|}{0.69} & \multicolumn{1}{c|}{0.62} & 0.62 \\
\multicolumn{1}{c|}{2 - Many Shots} & \multicolumn{1}{c|}{5} & \multicolumn{1}{c|}{0.69} & \multicolumn{1}{c|}{0.62} & 0.62 \\
\multicolumn{1}{l|}{} & \multicolumn{1}{c|}{7} & \multicolumn{1}{c|}{0.68} & \multicolumn{1}{c|}{0.62} & 0.62 \\
\multicolumn{1}{l|}{} & \multicolumn{1}{c|}{9} & \multicolumn{1}{c|}{0.67} & \multicolumn{1}{c|}{0.61} & 0.62 \\
\multicolumn{1}{l|}{} & \multicolumn{1}{c|}{11} & \multicolumn{1}{c|}{0.67} & \multicolumn{1}{c|}{0.61} & 0.61 \\
\multicolumn{1}{l}{} & \multicolumn{1}{l}{} & \multicolumn{1}{l}{} & \multicolumn{1}{l}{} & \multicolumn{1}{l}{} \\
\multicolumn{5}{c}{\textbf{2) Monastero dei Benedettini}} \\ \hline
\multicolumn{1}{c|}{\textbf{Variant}} & \multicolumn{1}{c|}{\textbf{K}} & \multicolumn{1}{c|}{\textbf{Precision}} & \multicolumn{1}{c|}{\textbf{Recall}} & \textbf{$\mathbf{F_1}$ score} \\ \hline
\multicolumn{1}{l|}{1 - One shot} & \multicolumn{1}{l|}{1} & \multicolumn{1}{c|}{0.29} & \multicolumn{1}{c|}{0.07} & 0.08 \\ \hline
\multicolumn{1}{c|}{} & \multicolumn{1}{c|}{1} & \multicolumn{1}{c|}{0.87} & \multicolumn{1}{c|}{0.87} & 0.87 \\
\multicolumn{1}{c|}{} & \multicolumn{1}{c|}{3} & \multicolumn{1}{c|}{0.88} & \multicolumn{1}{c|}{0.87} & 0.87 \\
\multicolumn{1}{c|}{2 - Many Shots} & \multicolumn{1}{c|}{5} & \multicolumn{1}{c|}{0.88} & \multicolumn{1}{c|}{0.88} & \textbf{0.88} \\
\multicolumn{1}{l|}{} & \multicolumn{1}{c|}{7} & \multicolumn{1}{c|}{0.88} & \multicolumn{1}{c|}{0.87} & 0.87 \\
\multicolumn{1}{l|}{} & \multicolumn{1}{c|}{9} & \multicolumn{1}{c|}{0.87} & \multicolumn{1}{c|}{0.87} & 0.87 \\
\multicolumn{1}{l|}{} & \multicolumn{1}{c|}{11} & \multicolumn{1}{c|}{0.87} & \multicolumn{1}{c|}{0.86} & 0.86
\end{tabular}
\end{table}

\subsection{Survey Prediction}
\textbf{Task:} Each test video of the ``Monastero dei Benedettini'' is associated to a survey collected from visitors at the end of the visit.
We define this task as predicting the content of a survey from the analysis of the related egocentric video.
We deem this to be possible as the egocentric video contains information on what the visitor has seen during the visit.
In particular, the task consists in predicting for each POI 1) if the POI has been remembered by the visitor and 2) how the POI would be rated by the visitor in a $[-7,7]$ scale.
This task investigates automatic algorithms for automatically ``filling in'' surveys from videos.

\textbf{Baseline:} 
Since the proposed task is novel and very challenging, as a proof of concept, we propose a baseline which takes as input the temporal annotations annotations indicating the objects observed by the visitors in the $60$ visits.
To obtain fixed-length descriptors for each video, we accumulate the number of frames in which a given POI has been observed in a Bag Of Word representation. %
In such representation, each component of the fixed-length vector indicates the total time in which a specific point of interest has been observed by the visitor. 
The vector is hence sum-normalized to reduce the influence of videos with different lengths.
The whole training set is normalized with z-scoring and classification is performed using K-NN.
We consider two baselines.
The first one simply performs a binary classification to predict whether a POI has been remembered by the visitor or not.
The second one predicts both if the POI has been seen and what score has been assigned to it.
This is tackled as a $15$-class classification problem, where class $-8$ indicates that the POI has not been remembered, whereas the other $14$ classes represent the scores from $-7$ to $7$ assigned by the visitors to POIs. We would like to note that we treat the problem as a classification task, as the scores assigned by the visitors are discrete integer numbers. Also, the dataset contains a limited set of data-points, which would prevent the algorithm from generalizing beyond the discrete set of labels available at training time.

\textbf{Implementation Details and Evaluation Measures:} 
We perform our experiments using a leave-one-out strategy. We tested different values for $k$ ranging from $1$ to $9$ and chose $K=9$ which resulted to be optimal in our experiments.
We evaluate results with weighted precision, recall and $F_1$ score.

\textbf{Results:}
Table~\ref{tab:binary} reports the results obtained in the case of binary classification (remembered vs not remembered)\footnote{\label{survey} See the supplementary material for the extended tables.}. 
The number of instances belonging to each class is reported in the last column. 
The results suggest that this task is very challenging.
Indeed, even if a POI appears in some frames, this does not imply that the visitor remembers it. 
Table~\ref{tab:multiclass} shows that the multi-class task\footnotemark[\getrefnumber{survey}] is even more challenging, with classes containing fewer examples (e.g., $-6, -5, -4, -3$) hard to recognize.
As a final remark, it is worth noting that the results suggest that the task can be addressed to some degree.
We expect that more complex approaches leveraging the analysis of the semantics of the input videos and the estimation of the attention of the visitor can achieve much better performance.

\begin{table*}[t]
\centering
\caption{Survey prediction results - binary classification task.}
\label{tab:binary}
\scriptsize
\centering
\begin{tabular}{lcccc}
\textbf{Class}    & \multicolumn{1}{l}{\textbf{Precision}} & \multicolumn{1}{l}{\textbf{Recall}} & \multicolumn{1}{l}{\textbf{$\mathbf{F_1}$ score}} & \multicolumn{1}{l}{\textbf{support}} \\ \hline
\textbf{Not Remembered} & 0,43                                   & 0,2                                 & 0,27                                   & 561                                  \\
\textbf{Remembered}     & 0,74                                   & 0,89                                & 0,81                                   & 1419                                 \\ \hline
\textbf{AVG}      & 0,65                                   & 0,7                                 & \textbf{0,66}                          & 1980                                
\end{tabular}
\end{table*}

\begin{table*}[t]
\centering
\caption{Survey prediction results - multi-class classification. ``Weighted AVG'' reports the average scores weighted by the number of samples in each class.}
\label{tab:multiclass}
\scriptsize
\begin{tabular}{lcccc}
\textbf{Class}        & \multicolumn{1}{l}{\textbf{Precision}} & \multicolumn{1}{l}{\textbf{Recall}} & \multicolumn{1}{l}{\textbf{$\mathbf{F_1}$ score}} & \multicolumn{1}{l}{\textbf{Support}} \\ \hline
\textbf{Not Remem.}           & 0,32                                   & 0,63                                & 0,43                                  & 561                                  \\
\textbf{-7}           & 0,52                                   & 0,24                                & 0,33                                  & 49                                   \\
\textbf{-6}           & 0                                      & 0                                   & 0                                     & 8                                    \\
\textbf{-5}           & 0                                      & 0                                   & 0                                     & 8                                    \\
\textbf{-4}           & 0                                      & 0                                   & 0                                     & 5                                    \\
\textbf{-3}           & 0                                      & 0                                   & 0                                     & 5                                    \\
\textbf{-2}           & 0,09                                   & 0,08                                & 0,08                                  & 13                                   \\
\textbf{-1}           & 0                                      & 0                                   & 0                                     & 10                                   \\
\textbf{0}            & 0,18                                   & 0,15                                & 0,17                                  & 104                                  \\
\textbf{1}            & 0                                      & 0                                   & 0                                     & 36                                   \\
\textbf{2}            & 0,02                                   & 0,02                                & 0,02                                  & 65                                   \\
\textbf{3}            & 0,12                                   & 0,02                                & 0,04                                  & 91                                   \\
\textbf{4}            & 0,1                                    & 0,04                                & 0,06                                  & 181                                  \\
\textbf{5}            & 0,13                                   & 0,07                                & 0,09                                  & 213                                  \\
\textbf{6}            & 0,14                                   & 0,09                                & 0,11                                  & 248                                  \\
\textbf{7}            & 0,33                                   & 0,29                                & 0,31                                  & 383                                  \\ \hline
\textbf{weighted AVG} & 0,23                                   & 0,27                                & \textbf{0,23}                         & 1980                                
\end{tabular}
\end{table*}

The code of our baselines is public available. See our web-page for the details: https://iplab.dmi.unict.it/EGO-CH/\#code.\hspace{2mm}

\section{\uppercase{Conclusion}}
\label{sec:conclusions}
We presented EGO-CH, a dataset for visitors behavioral understanding using egocentric vision. The dataset includes more than $27$ hours of video, $70$ visits acquired by real visitors, $26$ environments and over $200$ different points of interest related to two different cultural sites.
We publicly release the dataset along with temporal labels for locations and observed points of interest, bounding box annotations for objects, and surveys associated to $60$ visits.
Baseline results on the challenging tasks of Room-based Localization, Point of Interest/Object Recognition, Object Retrieval and Survey Prediction  show the potential of the dataset for visitors behavioral understanding. 
We believe that EGO-CH can be a valuable benchmark to tackle the proposed tasks, as well as others not investigated in this paper. Future works can address the evaluation considering more advanced baselines and investigate specialized approach to the four proposed tasks.

\section*{Acknowledgment}
This research is part of the project VALUE - Visual Analysis for Localization and Understanding of Environments (N. 08CT6209090207, CUP G69J18001060007) supported by PO FESR 2014/2020 - Azione 1.1.5. - ``Sostegno all'avanzamento tecnologico delle imprese attraverso il finanziamento di linee pilota e azioni di validazione precoce dei prodotti e di dimostrazioni su larga scala'' del PO FESR Sicilia 2014/2020, and Piano della Ricerca 2016-2018 linea di Intervento 2 of DMI, University of Catania. The authors would like to thank Regione Siciliana Assessorato dei Beni Culturali dell'Identità Siciliana - Dipartimento dei Beni Culturali e dell'Identità Siciliana and Polo regionale di Siracusa per i siti culturali - Galleria Regionale di Palazzo Bellomo.

\bibliography{refs}

\begin{thebibliography}{10}
\expandafter\ifx\csname url\endcsname\relax
  \def\url#1{\texttt{#1}}\fi
\expandafter\ifx\csname urlprefix\endcsname\relax\def\urlprefix{URL }\fi
\expandafter\ifx\csname href\endcsname\relax
  \def\href#1#2{#2} \def\path#1{#1}\fi

\bibitem{EmpiricalBollo}
A.~Bollo, L.~Pozzolo, Analysis of visitor behaviour inside the museum: An
  empirical study, in: Arts \& Cultural Management, 2005.

\bibitem{cucchiara2014visions}
R.~Cucchiara, A.~Del~Bimbo, Visions for augmented cultural heritage experience,
  IEEE MultiMedia 21~(1) (2014) 74--82.

\bibitem{ragusavedi2018}
F.~Ragusa, A.~Furnari, S.~Battiato, G.~Signorello, G.~M. Farinella, Egocentric
  visitors localization in cultural sites, J. Comput. Cult. Herit. 12~(2)
  (2019) 11:1--11:19.

\bibitem{ragusasigmap2018}
F.~Ragusa, L.~Guarnera, A.~Furnari, S.~Battiato, G.~Signorello, G.~M.
  Farinella, Localization of visitors for cultural sites management, in:
  International Joint Conference on e-Business and Telecommunications - Volume
  2: ICETE, 2018, pp. 407--413.

\bibitem{ragusavisapp}
F.~Ragusa, A.~Furnari, S.~Battiato, G.~Signorello, G.~M. Farinella, Egocentric
  point of interest recognition in cultural sites, in: VISAPP, 2019.

\bibitem{Razavian2014EstimatingAI}
A.~S. Razavian, O.~Aghazadeh, J.~Sullivan, S.~Carlsson, Estimating attention in
  exhibitions using wearable cameras, ICPR (2014) 2691--2696.

\bibitem{seidenari2017deep}
L.~Seidenari, C.~Baecchi, T.~Uricchio, A.~Ferracani, M.~Bertini, A.~D. Bimbo,
  Deep artwork detection and retrieval for automatic context-aware audio
  guides, TOMM 13~(3s) (2017) 35.

\bibitem{Raptis2005}
D.~Raptis, N.~K. Tselios, N.~M. Avouris, Context-based design of mobile
  applications for museums: a survey of existing practices, in: Mobile HCI,
  2005.

\bibitem{museumexhibit2018}
P.~Koniusz, Y.~Tas, H.~Zhang, M.~Harandi, F.~Porikli, R.~Zhang, Museum exhibit
  identification challenge for domain adaptation and beyond (2018).
\newblock \href {http://arxiv.org/abs/1802.01093} {\path{arXiv:1802.01093}}.

\bibitem{DelChiaro2019}
R.~D. Chiaro, A.~Bagdanov, A.~D. Bimbo, {{\{}NoisyArt{\}}: A Dataset for
  Webly-supervised Artwork Recognition}, in: International Joint Conference on
  Computer Vision, Imaging and Computer Graphics Theory and Applications, 2019.

\bibitem{Ahmetovic2016}
D.~Ahmetovic, C.~Gleason, K.~M. Kitani, H.~Takagi, C.~Asakawa, Navcog:
  Turn-by-turn smartphone navigation assistant for people with visual
  impairments or blindness, in: Web for All Conference, W4A '16, 2016, pp.
  9:1--9:2.

\bibitem{kendall2015posenet}
A.~Kendall, M.~Grimes, R.~Cipolla, Posenet: A convolutional network for
  real-time 6-dof camera relocalization, in: ICCV, 2015, pp. 2938--2946.

\bibitem{taverriti2016real}
G.~Taverriti, S.~Lombini, L.~Seidenari, M.~Bertini, A.~Del~Bimbo, Real-time
  wearable computer vision system for improved museum experience, in: ACM
  Multimedia, 2016, pp. 703--704.

\bibitem{yolov3}
J.~Redmon, A.~Farhadi, Yolov3: An incremental improvement, CoRR abs/1804.02767
  (2018).
\newblock \href {http://arxiv.org/abs/1804.02767} {\path{arXiv:1804.02767}}.

\bibitem{Rubhasy2014ManagementAR}
A.~Rubhasy, A.~A. G.~Y. Paramartha, I.~Budi, Z.~A. Hasibuan, Management and
  retrieval of cultural heritage multimedia collection using ontology,
  International Conference on Information Technology, Computer, and Electrical
  Engineering (2014) 255--259.

\bibitem{KwanRetrieval}
P.~Kwan, K.~Kameyama, J.~Gao, K.~Toraichi, Content-based image retrieval of
  cultural heritage symbols by interaction of visual perspectives., IJPRAI 25
  (2011) 643--673.

\bibitem{PatternRetrieval}
D.~K. Iakovidis, E.~E. Kotsifakos, N.~Pelekis, H.~Karanikas, I.~Kopanakis,
  T.~Mavroudakis, Y.~Theodoridis, Pattern-based retrieval of cultural heritage
  images, 2007.

\bibitem{Makantasis2016}
K.~Makantasis, A.~Doulamis, N.~Doulamis, M.~Ioannides, In the wild image
  retrieval and clustering for 3d cultural heritage landmarks reconstruction,
  Multimedia Tools and Applications 75~(7) (2016) 3593--3629.

\bibitem{furnari2018personal}
A.~Furnari, S.~Battiato, G.~M. Farinella, Personal-location-based temporal
  segmentation of egocentric videos for lifelogging applications, Journal of
  Visual Communication and Image Representation 52 (2018) 1 -- 12.

\bibitem{bellomo}
\href{http://www.regione.sicilia.it/beniculturali/palazzobellomo/}{Galleria
  regionale di palazzo bellomo} (2007).
\newline\urlprefix\url{http://www.regione.sicilia.it/beniculturali/palazzobellomo/}

\bibitem{Benedettini}
\href{http://www.monasterodeibenedettini.it/}{Monastero dei benedettini}.
\newline\urlprefix\url{http://www.monasterodeibenedettini.it/}

\bibitem{huang2017densely}
G.~Huang, Z.~Liu, L.~van~der Maaten, K.~Q. Weinberger, Densely connected
  convolutional networks, in: Proceedings of the IEEE Conference on Computer
  Vision and Pattern Recognition, 2017.

\end{thebibliography}

\pagebreak
\section*{\uppercase{Supplementary Material}}
\label{sec:supp_material}
This supplementary material complements the submitted paper. It reports additional details on the EGO-CH dataset and experiments.

\subsection{\uppercase{Introduction}}
\label{sec:introduction}

\begin{table*}[]
\centering
\caption{$FF_1$ Results for Room- Based Localization on Palazzo Bellomo.}
\label{tab:F1_Dense}
\scriptsize
\begin{tabular}{lcccccccccc}
\multicolumn{11}{c}{\textbf{$FF_1$ score}} \\ \hline
\multicolumn{1}{l|}{Class} & \multicolumn{1}{l}{Test1} & \multicolumn{1}{l}{Test2} & \multicolumn{1}{l}{Test4} & \multicolumn{1}{l}{Test5} & \multicolumn{1}{l}{Test6} & \multicolumn{1}{l}{Test7} & \multicolumn{1}{l}{Test8} & \multicolumn{1}{l}{Test9} & \multicolumn{1}{l|}{Test10} & \multicolumn{1}{l}{AVG} \\ \hline
\multicolumn{1}{l|}{1\_Sala1} & 0,00 & 0,96 & 0,96 & 0,95 & 0,99 & 0,94 & 0,97 & 0,97 & \multicolumn{1}{c|}{0,91} & 0,85 \\
\multicolumn{1}{l|}{2\_Sala2} & 0,82 & 0,91 & 0,96 & 0,95 & 0,99 & 0,99 & 0,99 & 0,97 & \multicolumn{1}{c|}{0,95} & 0,95 \\
\multicolumn{1}{l|}{3\_Sala3} & 0,65 & 0,84 & 0,84 & 0,83 & 0,94 & 0,88 & 0,61 & 0,85 & \multicolumn{1}{c|}{0,83} & 0,81 \\
\multicolumn{1}{l|}{4\_Sala4} & 0,84 & 0,96 & 0,98 & 0,93 & / & 0,96 & 0,78 & 0,97 & \multicolumn{1}{c|}{0,93} & 0,92 \\
\multicolumn{1}{l|}{5\_Sala5} & 0,98 & 0,88 & 0,96 & 0,91 & 0,65 & 0,91 & 0,97 & 0,97 & \multicolumn{1}{c|}{0,99} & 0,91 \\
\multicolumn{1}{l|}{6\_Sala6} & 0,84 & 0,80 & 0,43 & 0,00 & 0,86 & 0,00 & 0,18 & 0,83 & \multicolumn{1}{c|}{0,73} & 0,52 \\
\multicolumn{1}{l|}{7\_Sala7} & 0,67 & 0,93 & 0,22 & 0,00 & 0,38 & 0,88 & 0,88 & 0,40 & \multicolumn{1}{c|}{0,96} & 0,59 \\
\multicolumn{1}{l|}{8\_Sala8} & 0,64 & 0,75 & 0,60 & 0,52 & 0,56 & 0,56 & 0,64 & 0,60 & \multicolumn{1}{c|}{0,77} & 0,63 \\
\multicolumn{1}{l|}{9\_Sala9} & 0,90 & 0,89 & 0,73 & 0,88 & 0,41 & 0,92 & 0,98 & 0,99 & \multicolumn{1}{c|}{0,88} & 0,84 \\
\multicolumn{1}{l|}{10\_Sala10} & 0,96 & 0,98 & 0,92 & 0,72 & 0,00 & 0,98 & 0,85 & 0,97 & \multicolumn{1}{c|}{0,78} & 0,80 \\
\multicolumn{1}{l|}{11\_Sala11} & 0,93 & 0,96 & 0,96 & 0,97 & 0,00 & 0,97 & 0,97 & 0,98 & \multicolumn{1}{c|}{0,95} & 0,85 \\
\multicolumn{1}{l|}{12\_Sala12} & 0,82 & 0,88 & 0,87 & 0,90 & 0,00 & 0,87 & 0,85 & 0,92 & \multicolumn{1}{c|}{0,87} & 0,78 \\
\multicolumn{1}{l|}{13\_Sala13} & 0,96 & 0,95 & 0,95 & 0,76 & 0,00 & 0,93 & 0,95 & 0,94 & \multicolumn{1}{c|}{0,96} & 0,82 \\
\multicolumn{1}{l|}{14\_CortiledegliStemmi} & 0,78 & 0,68 & 0,74 & 0,88 & 0,17 & 0,90 & 0,79 & 0,92 & \multicolumn{1}{c|}{0,00} & 0,65 \\
\multicolumn{1}{l|}{15\_SalaCarrozze} & 0,84 & 0,89 & 0,95 & 0,93 & 0,91 & 0,84 & 0,97 & 0,95 & \multicolumn{1}{c|}{/} & 0,91 \\
\multicolumn{1}{l|}{16\_CortileParisio} & 0,33 & 0,51 & 0,52 & 0,45 & 0,35 & 0,60 & 0,81 & 0,91 & \multicolumn{1}{c|}{0,80} & 0,59 \\
\multicolumn{1}{l|}{17\_Biglietteria} & 0,25 & 0,56 & 0,00 & 0,26 & 0,00 & 0,00 & 0,00 & 0,36 & \multicolumn{1}{c|}{0,00} & 0,16 \\
\multicolumn{1}{l|}{18\_Portico} & 0,68 & 0,67 & 0,78 & 0,54 & 0,69 & 0,69 & 0,62 & 0,78 & \multicolumn{1}{c|}{0,62} & 0,67 \\
\multicolumn{1}{l|}{19\_ScalaCatalana} & 0,00 & 0,00 & 0,54 & 0,55 & 0,78 & 0,67 & 0,58 & 0,85 & \multicolumn{1}{c|}{0,49} & 0,50 \\
\multicolumn{1}{l|}{20\_Loggetta} & 0,00 & 0,50 & 0,80 & 0,44 & 0,72 & 0,53 & 0,55 & 0,83 & \multicolumn{1}{c|}{0,68} & 0,56 \\
\multicolumn{1}{l|}{21\_BoxSala8} & 0,96 & 0,97 & 0,97 & / & 0,00 & 0,97 & 0,99 & 0,81 & \multicolumn{1}{c|}{0,94} & 0,83 \\
\multicolumn{1}{l|}{22\_AreaSosta} & 0,74 & 0,85 & 0,45 & 0,83 & 0,00 & 0,56 & 0,72 & 0,85 & \multicolumn{1}{c|}{0,00} & 0,56 \\ \hline
\multicolumn{1}{c|}{$mFF_1$} & 0,66 & 0,79 & 0,73 & 0,68 & 0,45 & 0,75 & 0,76 & 0,85 & \multicolumn{1}{c|}{0,72} & 0,71
\end{tabular}
\end{table*}

\begin{table*}[]
\centering
\caption{$ASF_1$ Results for Room- Based Localization on Palazzo Bellomo.}
\label{tab:ASF1_Dense}
\scriptsize
\begin{tabular}{lcccccccccc}
\multicolumn{11}{c}{\textbf{$ASF_1$ score}} \\ \hline
\multicolumn{1}{l|}{Class} & \multicolumn{1}{l}{Test1} & \multicolumn{1}{l}{Test2} & \multicolumn{1}{l}{Test4} & \multicolumn{1}{l}{Test5} & \multicolumn{1}{l}{Test6} & \multicolumn{1}{l}{Test7} & \multicolumn{1}{l}{Test8} & \multicolumn{1}{l}{Test9} & \multicolumn{1}{l|}{Test10} & \multicolumn{1}{l}{AVG} \\ \hline
\multicolumn{1}{l|}{1\_Sala1} & 0,00 & 0,93 & 0,93 & 0,91 & 0,97 & 0,89 & 0,94 & 0,94 & \multicolumn{1}{c|}{0,84} & 0,82 \\
\multicolumn{1}{l|}{2\_Sala2} & 0,64 & 0,83 & 0,91 & 0,90 & 0,97 & 0,97 & 0,97 & 0,94 & \multicolumn{1}{c|}{0,90} & 0,89 \\
\multicolumn{1}{l|}{3\_Sala3} & 0,36 & 0,53 & 0,51 & 0,53 & 0,61 & 0,63 & 0,29 & 0,59 & \multicolumn{1}{c|}{0,72} & 0,53 \\
\multicolumn{1}{l|}{4\_Sala4} & 0,39 & 0,92 & 0,96 & 0,86 & / & 0,92 & 0,31 & 0,94 & \multicolumn{1}{c|}{0,87} & 0,77 \\
\multicolumn{1}{l|}{5\_Sala5} & 0,94 & 0,64 & 0,66 & 0,62 & 0,34 & 0,63 & 0,54 & 0,95 & \multicolumn{1}{c|}{0,97} & 0,70 \\
\multicolumn{1}{l|}{6\_Sala6} & 0,46 & 0,43 & 0,18 & 0,00 & 0,42 & 0,00 & 0,09 & 0,59 & \multicolumn{1}{c|}{0,51} & 0,30 \\
\multicolumn{1}{l|}{7\_Sala7} & 0,50 & 0,86 & 0,13 & 0,00 & 0,10 & 0,78 & 0,28 & 0,25 & \multicolumn{1}{c|}{0,92} & 0,42 \\
\multicolumn{1}{l|}{8\_Sala8} & 0,36 & 0,44 & 0,42 & 0,35 & 0,35 & 0,41 & 0,49 & 0,53 & \multicolumn{1}{c|}{0,57} & 0,44 \\
\multicolumn{1}{l|}{9\_Sala9} & 0,81 & 0,80 & 0,51 & 0,78 & 0,22 & 0,84 & 0,95 & 0,97 & \multicolumn{1}{c|}{0,78} & 0,74 \\
\multicolumn{1}{l|}{10\_Sala10} & 0,92 & 0,96 & 0,85 & 0,19 & 0,00 & 0,95 & 0,27 & 0,94 & \multicolumn{1}{c|}{0,64} & 0,64 \\
\multicolumn{1}{l|}{11\_Sala11} & 0,53 & 0,92 & 0,91 & 0,93 & 0,00 & 0,93 & 0,65 & 0,95 & \multicolumn{1}{c|}{0,91} & 0,75 \\
\multicolumn{1}{l|}{12\_Sala12} & 0,38 & 0,79 & 0,76 & 0,81 & 0,00 & 0,77 & 0,30 & 0,85 & \multicolumn{1}{c|}{0,76} & 0,60 \\
\multicolumn{1}{l|}{13\_Sala13} & 0,93 & 0,90 & 0,90 & 0,62 & 0,00 & 0,87 & 0,91 & 0,89 & \multicolumn{1}{c|}{0,92} & 0,77 \\
\multicolumn{1}{l|}{14\_CortiledegliStemmi} & 0,57 & 0,53 & 0,54 & 0,77 & 0,08 & 0,81 & 0,49 & 0,67 & \multicolumn{1}{c|}{0,00} & 0,50 \\
\multicolumn{1}{l|}{15\_SalaCarrozze} & 0,72 & 0,80 & 0,91 & 0,87 & 0,83 & 0,72 & 0,94 & 0,90 & \multicolumn{1}{c|}{/} & 0,84 \\
\multicolumn{1}{l|}{16\_CortileParisio} & 0,30 & 0,63 & 0,43 & 0,65 & 0,38 & 0,47 & 0,59 & 0,83 & \multicolumn{1}{c|}{0,66} & 0,55 \\
\multicolumn{1}{l|}{17\_Biglietteria} & 0,26 & 0,43 & 0,00 & 0,11 & 0,00 & 0,00 & 0,00 & 0,39 & \multicolumn{1}{c|}{0,00} & 0,13 \\
\multicolumn{1}{l|}{18\_Portico} & 0,51 & 0,47 & 0,60 & 0,45 & 0,48 & 0,52 & 0,42 & 0,64 & \multicolumn{1}{c|}{0,47} & 0,51 \\
\multicolumn{1}{l|}{19\_ScalaCatalana} & 0,00 & 0,00 & 0,43 & 0,51 & 0,61 & 0,55 & 0,42 & 0,75 & \multicolumn{1}{c|}{0,39} & 0,41 \\
\multicolumn{1}{l|}{20\_Loggetta} & 0,00 & 0,31 & 0,54 & 0,28 & 0,46 & 0,38 & 0,48 & 0,72 & \multicolumn{1}{c|}{0,44} & 0,40 \\
\multicolumn{1}{l|}{21\_BoxSala8} & 0,92 & 0,95 & 0,95 & / & 0,00 & 0,93 & 0,98 & 0,67 & \multicolumn{1}{c|}{0,88} & 0,79 \\
\multicolumn{1}{l|}{22\_AreaSosta} & 0,59 & 0,73 & 0,29 & 0,71 & 0,00 & 0,75 & 0,57 & 0,74 & \multicolumn{1}{c|}{0,00} & 0,49 \\ \hline
\multicolumn{1}{c|}{$mASF_1$} & 0,50 & 0,67 & 0,61 & 0,56 & 0,32 & 0,67 & 0,54 & 0,76 & \multicolumn{1}{c|}{0,63} & 0,58
\end{tabular}
\end{table*}

\begin{table*}[]
\centering
\caption{$FF_1$ Results for Room- Based Localization on Monastero dei Benedettini.}
\label{tab:my-table}
\scriptsize
\begin{tabular}{lccccccll|lllll|l}
\multicolumn{1}{l|}{\textbf{ID\_visit}} & \textbf{1} & \textbf{2} & \textbf{3} & \textbf{4} & \multicolumn{1}{c|}{\textbf{Neg.}} & \textbf{AVG} &  & \textbf{ID\_visit} & \textbf{1} & \textbf{2} & \textbf{3} & \textbf{4} & \textbf{Neg.} & \textbf{AVG} \\ \cline{1-7} \cline{9-15} 
\multicolumn{1}{l|}{4805} & 0,79 & 0,82 & 0,74 & 0,95 & \multicolumn{1}{c|}{0,45} & 0,75 &  & 2043 & 0,52 & 0,44 & 0,62 & 0,91 & 0,33 & 0,564 \\
\multicolumn{1}{l|}{1804} & 0,32 & 0,36 & 0,79 & 0,98 & \multicolumn{1}{c|}{0,55} & 0,6 &  & 3996 & 0,17 & 0,68 & 0,57 & 0,95 & 0,72 & 0,618 \\
\multicolumn{1}{l|}{4377} & 0,46 & 0,53 & 0,8 & 0,96 & \multicolumn{1}{c|}{0,32} & 0,614 &  & 3455 & 0,78 & 0,65 & 0,8 & 0,99 & 0,55 & 0,754 \\
\multicolumn{1}{l|}{1669} & 0,75 & 0,72 & 0,9 & 0,99 & \multicolumn{1}{c|}{0,45} & 0,762 &  & 4785 & 0,02 & 0 & 0,64 & 0,95 & 0,33 & 0,388 \\
\multicolumn{1}{l|}{1791} & 0,49 & 0,5 & 0,84 & 0,95 & \multicolumn{1}{c|}{0,43} & 0,642 &  & 2047 & 0,95 & 0,84 & 0,84 & 0,98 & 0,38 & 0,798 \\
\multicolumn{1}{l|}{3948} & 0 & / & 0,59 & 0,98 & \multicolumn{1}{c|}{0,49} & 0,515 &  & 1912 & 0,66 & 0,51 & 0,69 & 0,94 & 0,46 & 0,652 \\
\multicolumn{1}{l|}{3152} & 0,39 & 0,72 & 0,87 & 0,97 & \multicolumn{1}{c|}{0,76} & 0,742 &  & 3232 & 0,66 & 0,53 & 0,85 & 0,99 & 0,49 & 0,704 \\
\multicolumn{1}{l|}{4361} & 0,55 & 0,78 & 0,82 & 0,97 & \multicolumn{1}{c|}{0,28} & 0,68 &  & 4442 & 0,8 & 0,77 & 0,82 & 0,98 & 0,23 & 0,72 \\
\multicolumn{1}{l|}{3976} & 0,87 & 0,43 & 0,81 & 0,93 & \multicolumn{1}{c|}{0,66} & 0,74 &  & 3646 & 0,63 & 0,29 & 0,66 & 0,97 & 0,4 & 0,59 \\
\multicolumn{1}{l|}{3527} & 0,86 & 0,79 & 0,85 & 0,99 & \multicolumn{1}{c|}{0,56} & 0,81 &  & 4833 & 0,67 & 0,82 & 0,67 & 0,88 & 0,24 & 0,656 \\
\multicolumn{1}{l|}{4105} & 0,62 & 0 & 0,77 & 0,97 & \multicolumn{1}{c|}{0,09} & 0,49 &  & 3478 & 0,71 & 0,75 & 0,84 & 0,98 & 0,37 & 0,73 \\
\multicolumn{1}{l|}{1399} & 0,46 & 0,23 & 0,79 & 0,98 & \multicolumn{1}{c|}{0,43} & 0,578 &  & 4396 & 0,74 & 0,69 & 0,89 & 0,96 & 0,26 & 0,708 \\
\multicolumn{1}{l|}{3836} & 0,51 & 0 & 0,72 & 0,99 & \multicolumn{1}{c|}{0,55} & 0,554 &  & 2894 & 0,83 & 0,8 & 0,66 & 0,92 & 0,48 & 0,738 \\
\multicolumn{1}{l|}{4006} & 0,57 & 0,78 & 0,78 & 0,99 & \multicolumn{1}{c|}{0,37} & 0,698 &  & 4414 & 0,75 & 0,7 & 0,91 & 0,95 & 0,22 & 0,706 \\
\multicolumn{1}{l|}{4415} & 0,86 & 0,61 & 0,82 & 0,97 & \multicolumn{1}{c|}{0,35} & 0,722 &  & 4639 & 0,57 & 0,52 & 0,85 & 0,99 & 0,39 & 0,664 \\
\multicolumn{1}{l|}{3008} & 0,69 & 0 & 0,7 & 0,99 & \multicolumn{1}{c|}{0,51} & 0,578 &  & 1004 & 0,08 & 0,72 & 0,49 & 0,97 & 0,26 & 0,504 \\
\multicolumn{1}{l|}{4660} & 0,62 & 0,81 & 0,83 & 0,98 & \multicolumn{1}{c|}{0,57} & 0,762 &  & 1917 & 0,41 & 0,71 & 0,36 & 0,87 & 0,29 & 0,528 \\
\multicolumn{1}{l|}{2826} & 0,31 & 0,52 & 0,76 & 0,97 & \multicolumn{1}{c|}{0,61} & 0,634 &  & 1153 & 0,62 & 0,7 & 0,71 & 0,91 & 0,32 & 0,652 \\
\multicolumn{1}{l|}{1099} & 0,62 & 0,42 & 0,85 & 0,98 & \multicolumn{1}{c|}{0,49} & 0,672 &  & 2244 & 0,94 & 0,56 & 0,74 & 0,99 & 0,43 & 0,732 \\
\multicolumn{1}{l|}{4391} & 0,74 & 0,72 & 0,72 & 0,98 & \multicolumn{1}{c|}{0,33} & 0,698 &  & 2614 & 0,88 & 0,56 & 0,83 & 0,99 & 0,4 & 0,732 \\
\multicolumn{1}{l|}{3929} & 0,26 & 0 & 0,8 & 0,99 & \multicolumn{1}{c|}{0,43} & 0,496 &  & 1624 & 0,33 & 0,8 & 0,62 & 0,99 & 0,33 & 0,614 \\
\multicolumn{1}{l|}{3362} & 0,34 & 0,68 & 0,68 & 0,95 & \multicolumn{1}{c|}{0,21} & 0,572 &  & 3441 & 0,61 & 0,25 & 0,84 & 0,99 & 0,41 & 0,62 \\
\multicolumn{1}{l|}{1379} & 0,41 & 0 & 0,81 & 0,96 & \multicolumn{1}{c|}{0,45} & 0,526 &  & 4793 & 0,52 & / & 0,68 & 0,99 & 0,33 & 0,63 \\
\multicolumn{1}{l|}{2600} & 0,26 & 0 & 0,63 & 0,96 & \multicolumn{1}{c|}{0,3} & 0,43 &  & 4083 & 0,93 & / & 0,73 & 0,99 & 0,71 & 0,84 \\
\multicolumn{1}{l|}{1430} & 0,94 & 0 & 0,69 & 0,87 & \multicolumn{1}{c|}{0,58} & 0,616 &  & 4906 & 0,46 & 0,36 & 0,59 & 0,94 & 0,31 & 0,532 \\
\multicolumn{1}{l|}{2956} & 0,32 & 0,45 & 0,33 & 0,91 & \multicolumn{1}{c|}{0,65} & 0,532 &  & 1160 & 0,88 & 0,84 & 0,72 & 0,98 & 0,46 & 0,776 \\
\multicolumn{1}{l|}{4742} & 0,13 & 0,58 & 0,59 & 0,84 & \multicolumn{1}{c|}{0,5} & 0,528 &  & 3416 & 0,56 & / & 0,5 & 0,82 & 0,2 & 0,52 \\
\multicolumn{1}{l|}{3651} & 0,77 & 0,41 & 0,88 & 0,98 & \multicolumn{1}{c|}{0,41} & 0,69 &  & 1051 & 0,78 & 0,76 & 0,64 & 0,95 & 0,57 & 0,74 \\
\multicolumn{1}{l|}{1064} & 0,77 & 0,23 & 0,83 & 0,99 & \multicolumn{1}{c|}{0,22} & 0,608 &  & 2580 & 0,71 & 0,45 & 0,63 & 0,96 & 0,44 & 0,638 \\
\multicolumn{1}{l|}{3818} & 0,68 & 0,64 & 0,73 & 0,99 & \multicolumn{1}{c|}{0,47} & 0,702 &  & 1109 & 0,91 & 0,28 & 0,85 & 0,97 & 0,36 & 0,674 \\ \cline{1-7} \cline{9-15} 
 & \multicolumn{1}{l}{} & \multicolumn{1}{l}{} & \multicolumn{1}{l}{} & \multicolumn{1}{l}{} & \multicolumn{1}{l}{} & \multicolumn{1}{l}{} &  & \textbf{mFF1} & \textbf{0.59} & \textbf{0.51} & \textbf{0.73} & \textbf{0.96} & \textbf{0.42} & \textbf{0.64}
\end{tabular}
\end{table*}

\begin{table*}[]
\centering
\caption{$ASF_1$ Results for Room- Based Localization on Monastero dei Benedettini.}
\label{tab:my-table}
\scriptsize
\begin{tabular}{lccccccll|lllll|l}
\multicolumn{1}{l|}{\textbf{ID\_visit}} & \textbf{1} & \textbf{2} & \textbf{3} & \textbf{4} & \multicolumn{1}{c|}{\textbf{Neg.}} & \textbf{AVG} &  & \textbf{ID\_visit} & \textbf{1} & \textbf{2} & \textbf{3} & \textbf{4} & \textbf{Neg.} & \textbf{AVG} \\ \cline{1-7} \cline{9-15} 
\multicolumn{1}{l|}{4805} & 0,55 & 0,62 & 0,4 & 0,12 & \multicolumn{1}{c|}{0,36} & 0,41 &  & 2043 & 0,4 & 0,27 & 0,28 & 0,13 & 0,27 & 0,27 \\
\multicolumn{1}{l|}{1804} & 0,32 & 0,1 & 0,39 & 0,05 & \multicolumn{1}{c|}{0,21} & 0,214 &  & 3996 & 0,23 & 0,3 & 0,22 & 0,05 & 0,38 & 0,236 \\
\multicolumn{1}{l|}{4377} & 0,31 & 0,12 & 0,26 & 0,08 & \multicolumn{1}{c|}{0,27} & 0,208 &  & 3455 & 0,63 & 0,38 & 0,65 & 0,99 & 0,48 & 0,626 \\
\multicolumn{1}{l|}{1669} & 0,68 & 0,34 & 0,67 & 0,9 & \multicolumn{1}{c|}{0,33} & 0,584 &  & 4785 & 0,06 & 0 & 0,25 & 0,45 & 0,07 & 0,166 \\
\multicolumn{1}{l|}{1791} & 0,41 & 0,41 & 0,25 & 0,06 & \multicolumn{1}{c|}{0,22} & 0,27 &  & 2047 & 0,9 & 0,72 & 0,72 & 0,37 & 0,34 & 0,61 \\
\multicolumn{1}{l|}{3948} & 0 & / & 0,33 & 0,26 & \multicolumn{1}{c|}{0,49} & 0,27 &  & 1912 & 0,42 & 0,25 & 0,2 & 0,08 & 0,3 & 0,25 \\
\multicolumn{1}{l|}{3152} & 0,39 & 0,43 & 0,64 & 0,44 & \multicolumn{1}{c|}{0,58} & 0,496 &  & 3232 & 0,52 & 0,1 & 0,34 & 0,13 & 0,35 & 0,288 \\
\multicolumn{1}{l|}{4361} & 0,26 & 0,35 & 0,32 & 0,08 & \multicolumn{1}{c|}{0,2} & 0,242 &  & 4442 & 0,61 & 0,22 & 0,33 & 0,06 & 0,12 & 0,268 \\
\multicolumn{1}{l|}{3976} & 0,68 & 0,26 & 0,35 & 0,03 & \multicolumn{1}{c|}{0,5} & 0,364 &  & 3646 & 0,62 & 0,3 & 0,26 & 0,19 & 0,22 & 0,318 \\
\multicolumn{1}{l|}{3527} & 0,65 & 0,37 & 0,38 & 0,28 & \multicolumn{1}{c|}{0,44} & 0,424 &  & 4833 & 0,54 & 0,55 & 0,26 & 0,1 & 0,21 & 0,332 \\
\multicolumn{1}{l|}{4105} & 0,49 & 0 & 0,49 & 0,16 & \multicolumn{1}{c|}{0,06} & 0,24 &  & 3478 & 0,57 & 0,4 & 0,47 & 0,14 & 0,24 & 0,364 \\
\multicolumn{1}{l|}{1399} & 0,55 & 0,1 & 0,33 & 0,23 & \multicolumn{1}{c|}{0,29} & 0,3 &  & 4396 & 0,41 & 0,1 & 0,2 & 0,07 & 0,12 & 0,18 \\
\multicolumn{1}{l|}{3836} & 0,47 & 0 & 0,35 & 0,98 & \multicolumn{1}{c|}{0,34} & 0,428 &  & 2894 & 0,55 & 0,67 & 0,36 & 0,13 & 0,31 & 0,404 \\
\multicolumn{1}{l|}{4006} & 0,47 & 0,49 & 0,49 & 0,55 & \multicolumn{1}{c|}{0,26} & 0,452 &  & 4414 & 0,59 & 0,26 & 0,83 & 0,11 & 0,08 & 0,374 \\
\multicolumn{1}{l|}{4415} & 0,74 & 0,07 & 0,35 & 0,14 & \multicolumn{1}{c|}{0,23} & 0,306 &  & 4639 & 0,5 & 0,35 & 0,67 & 0,39 & 0,27 & 0,436 \\
\multicolumn{1}{l|}{3008} & 0,57 & 0 & 0,5 & 0,99 & \multicolumn{1}{c|}{0,41} & 0,494 &  & 1004 & 0,22 & 0,47 & 0,17 & 0,11 & 0,17 & 0,228 \\
\multicolumn{1}{l|}{4660} & 0,45 & 0,52 & 0,17 & 0,05 & \multicolumn{1}{c|}{0,27} & 0,292 &  & 1917 & 0,39 & 0,19 & 0,03 & 0,03 & 0,21 & 0,17 \\
\multicolumn{1}{l|}{2826} & 0,31 & 0,29 & 0,33 & 0,06 & \multicolumn{1}{c|}{0,25} & 0,248 &  & 1153 & 0,47 & 0,33 & 0,21 & 0,14 & 0,16 & 0,262 \\
\multicolumn{1}{l|}{1099} & 0,52 & 0,24 & 0,55 & 0,96 & \multicolumn{1}{c|}{0,17} & 0,488 &  & 2244 & 0,77 & 0,35 & 0,49 & 0,35 & 0,32 & 0,456 \\
\multicolumn{1}{l|}{4391} & 0,54 & 0,13 & 0,14 & 0,2 & \multicolumn{1}{c|}{0,24} & 0,25 &  & 2614 & 0,43 & 0,06 & 0,39 & 0,99 & 0,26 & 0,426 \\
\multicolumn{1}{l|}{3929} & 0,15 & 0 & 0,31 & 0,22 & \multicolumn{1}{c|}{0,39} & 0,214 &  & 1624 & 0,3 & 0,33 & 0,24 & 0,29 & 0,29 & 0,29 \\
\multicolumn{1}{l|}{3362} & 0,2 & 0,35 & 0,38 & 0,05 & \multicolumn{1}{c|}{0,22} & 0,24 &  & 3441 & 0,49 & 0,11 & 0,44 & 0,88 & 0,42 & 0,468 \\
\multicolumn{1}{l|}{1379} & 0,22 & 0 & 0,43 & 0,05 & \multicolumn{1}{c|}{0,35} & 0,21 &  & 4793 & 0,35 & / & 0,36 & 0,28 & 0,37 & 0,34 \\
\multicolumn{1}{l|}{2600} & 0,32 & 0 & 0,39 & 0,31 & \multicolumn{1}{c|}{0,26} & 0,256 &  & 4083 & 0,88 & / & 0,41 & 0,25 & 0,48 & 0,505 \\
\multicolumn{1}{l|}{1430} & 0,43 & 0 & 0,1 & 0,02 & \multicolumn{1}{c|}{0,21} & 0,152 &  & 4906 & 0,36 & 0,1 & 0,31 & 0,06 & 0,21 & 0,208 \\
\multicolumn{1}{l|}{2956} & 0,41 & 0,2 & 0,16 & 0,29 & \multicolumn{1}{c|}{0,31} & 0,274 &  & 1160 & 0,79 & 0,5 & 0,39 & 0,14 & 0,28 & 0,42 \\
\multicolumn{1}{l|}{4742} & 0,15 & 0,18 & 0,25 & 0,11 & \multicolumn{1}{c|}{0,17} & 0,172 &  & 3416 & 0,32 & / & 0,25 & 0,42 & 0,34 & 0,3325 \\
\multicolumn{1}{l|}{3651} & 0,32 & 0,25 & 0,74 & 0,48 & \multicolumn{1}{c|}{0,28} & 0,414 &  & 1051 & 0,5 & 0,31 & 0,21 & 0,05 & 0,4 & 0,294 \\
\multicolumn{1}{l|}{1064} & 0,63 & 0,09 & 0,52 & 0,97 & \multicolumn{1}{c|}{0,19} & 0,48 &  & 2580 & 0,36 & 0,29 & 0,33 & 0,05 & 0,28 & 0,262 \\
\multicolumn{1}{l|}{3818} & 0,43 & 0,49 & 0,2 & 0,06 & \multicolumn{1}{c|}{0,29} & 0,294 &  & 1109 & 0,82 & 0,17 & 0,46 & 0,11 & 0,28 & 0,368 \\ \cline{1-7} \cline{9-15} 
 & \multicolumn{1}{l}{} & \multicolumn{1}{l}{} & \multicolumn{1}{l}{} & \multicolumn{1}{l}{} & \multicolumn{1}{l}{} & \multicolumn{1}{l}{} &  & \textbf{$mASF_1$} & \textbf{0,46} & \textbf{0,26} & \textbf{0,37} & \textbf{0,28} & \textbf{0,28} & \textbf{0.40}
\end{tabular}
\end{table*}

\begin{table}[]
\centering
\caption{Point of Interest Retrieval related to Palazzo Bellomo cultural site.}
\label{tab:my-table}
\scriptsize
\begin{tabular}{ccccc}
\multicolumn{5}{c}{\textbf{1) Palazzo Bellomo}} \\ \hline
\multicolumn{1}{c|}{\textbf{Variant}} & \multicolumn{1}{c|}{\textbf{K}} & \multicolumn{1}{c|}{\textbf{Precision}} & \multicolumn{1}{c|}{\textbf{Recall}} & \textbf{$F_1$ score} \\ \hline
\multicolumn{1}{c|}{1 - One Shot} & \multicolumn{1}{c|}{1} & \multicolumn{1}{c|}{0.02} & \multicolumn{1}{c|}{0.01} & 0.00 \\ \hline
\multicolumn{1}{c|}{} & \multicolumn{1}{c|}{1} & \multicolumn{1}{c|}{\textbf{0.62}} & \multicolumn{1}{c|}{\textbf{0.59}} & \textbf{0.6} \\
\multicolumn{1}{c|}{} & \multicolumn{1}{c|}{3} & \multicolumn{1}{c|}{0.62} & \multicolumn{1}{c|}{0.56} & 0.56 \\
\multicolumn{1}{c|}{2 - Many Shots} & \multicolumn{1}{c|}{5} & \multicolumn{1}{c|}{0.62} & \multicolumn{1}{c|}{0.56} & 0.56 \\
\multicolumn{1}{l|}{} & \multicolumn{1}{c|}{7} & \multicolumn{1}{c|}{0,61} & \multicolumn{1}{c|}{0,56} & 0,56 \\
\multicolumn{1}{l|}{} & \multicolumn{1}{c|}{9} & \multicolumn{1}{c|}{0,61} & \multicolumn{1}{c|}{0,55} & 0,56 \\
\multicolumn{1}{l|}{} & \multicolumn{1}{c|}{11} & \multicolumn{1}{c|}{0,61} & \multicolumn{1}{c|}{0,55} & 0,55
\end{tabular}
\end{table}

\begin{table}[]
\centering
\caption{Point of Interest Retrieval related to Monastero dei Benedettini cultural site.}
\label{tab:my-table}
\scriptsize
\begin{tabular}{ccccc}
\multicolumn{5}{c}{\textbf{2) Monastero dei Benedettini}} \\ \hline
\multicolumn{1}{c|}{\textbf{Variant}} & \multicolumn{1}{c|}{\textbf{K}} & \multicolumn{1}{c|}{\textbf{Precision}} & \multicolumn{1}{c|}{\textbf{Recall}} & \textbf{$F_1$ score} \\ \hline
\multicolumn{1}{l|}{1 - One shot} & \multicolumn{1}{l|}{1} & \multicolumn{1}{c|}{0.38} & \multicolumn{1}{c|}{0.07} & 0.09 \\ \hline
\multicolumn{1}{c|}{} & \multicolumn{1}{c|}{1} & \multicolumn{1}{c|}{0,83} & \multicolumn{1}{c|}{0,83} & 0,83 \\
\multicolumn{1}{c|}{} & \multicolumn{1}{c|}{3} & \multicolumn{1}{c|}{0,84} & \multicolumn{1}{c|}{0,83} & 0,83 \\
\multicolumn{1}{c|}{2 - Many Shots} & \multicolumn{1}{c|}{5} & \multicolumn{1}{c|}{\textbf{0,84}} & \multicolumn{1}{c|}{\textbf{0,84}} & \textbf{0.83} \\
\multicolumn{1}{l|}{} & \multicolumn{1}{c|}{7} & \multicolumn{1}{c|}{0,84} & \multicolumn{1}{c|}{0,83} & 0,83 \\
\multicolumn{1}{l|}{} & \multicolumn{1}{c|}{9} & \multicolumn{1}{c|}{0,84} & \multicolumn{1}{c|}{0,83} & 0,83 \\
\multicolumn{1}{l|}{} & \multicolumn{1}{c|}{11} & \multicolumn{1}{c|}{0,83} & \multicolumn{1}{c|}{0,83} & 0,82
\end{tabular}
\end{table}

\noindent
This document is intended for the convenience of the reader and reports additional information about the proposed dataset and the performed experiments. This supplementary material is related to the following submission:
\begin{itemize}
    \item F. Ragusa, A. Furnari, S. Battiato, G. Signorello, G. M. Farinella, ``EGO-CH: Dataset and Challenges for Visitors Behavioral Understanding from Egocentric Vision'', Pattern Recognition Letters, DOI: 10.1016/j.patrec.2019.12.016
\end{itemize}
The reader is referred to the manuscript and to our web page \textit{http://iplab.dmi.unict.it/EGO-CH/} for further information.

\subsection{\uppercase{THE EGO-CH DATASET}}
\label{sec:dataset}

\subsubsection{Data Collection}
The dataset has been acquired using a head-mounted Microsoft HoloLens device in two cultural sites located in Sicily, Italy: 1) ``Palazzo Bellomo'', located in Siracusa~\cite{bellomo}, and 2) ``Monastero dei Benedettini'', located in Catania~\cite{Benedettini}. 
\paragraph{1) Palazzo Bellomo}
This cultural site is composed by $22$ environments and contains $191$ Points of Interest (e.g., statues, paintings, etc.). Figure~\ref{fig:bellomoframes} and Figure~\ref{fig:bellomoPOI} report some frames related to the different environments and some points of interest. Table~\ref{tab:bellomotrain} details the list of the acquired training video. Some of the videos are related to the $22$ rooms of the cultural site, whereas other are related to specific points of interest. 
For each video we report its total duration, the amount of required storage, the number of frames, as well as the percentage of frames with respect to the whole training set.

\begin{figure}
\centering
\includegraphics[width=8cm]{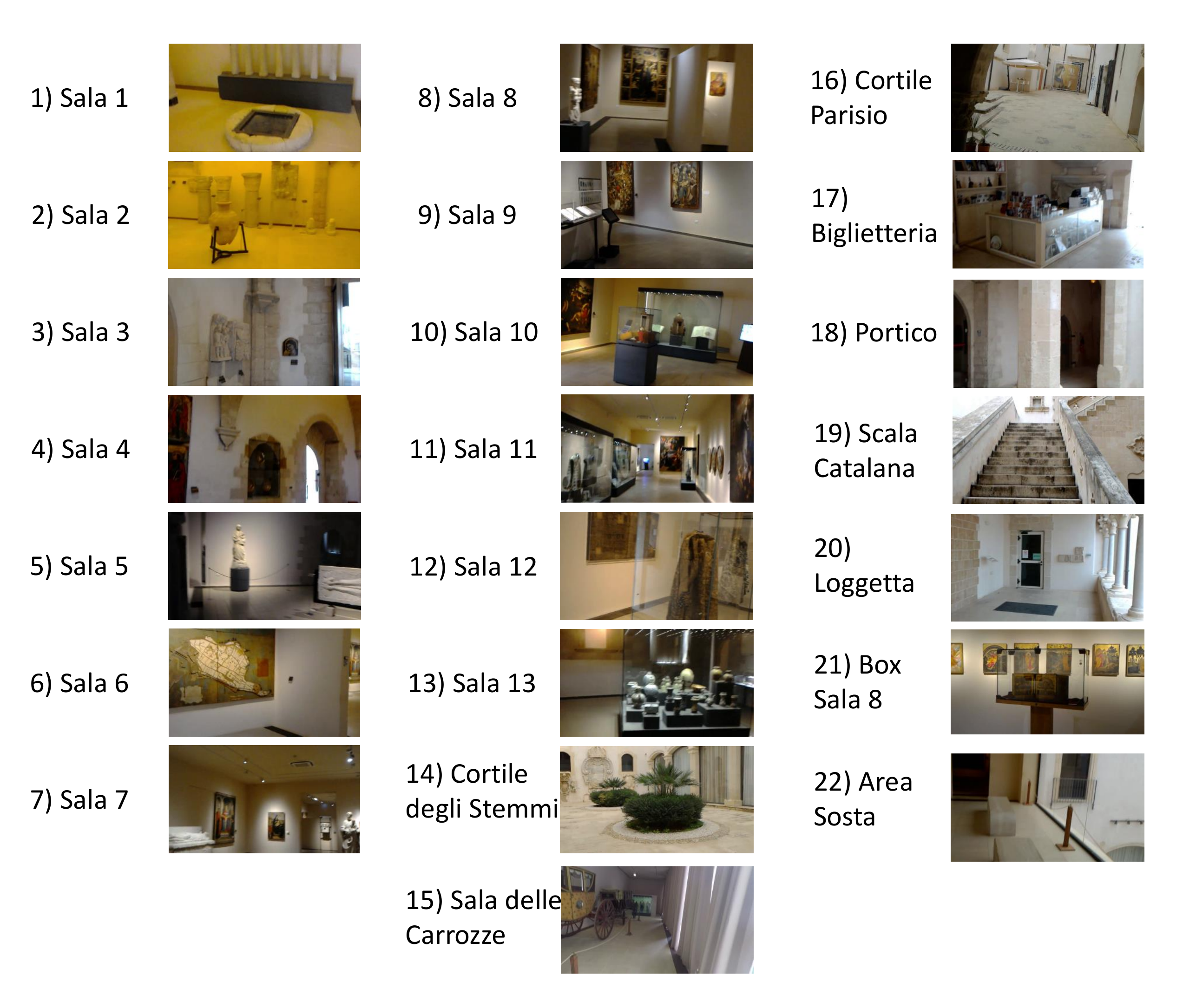}
\caption{Sample frames for each of the $22$ considered environments of ``Palazzo Bellomo''.}
\label{fig:bellomoframes}
\end{figure}

\begin{figure}
\centering
\includegraphics[width=8cm]{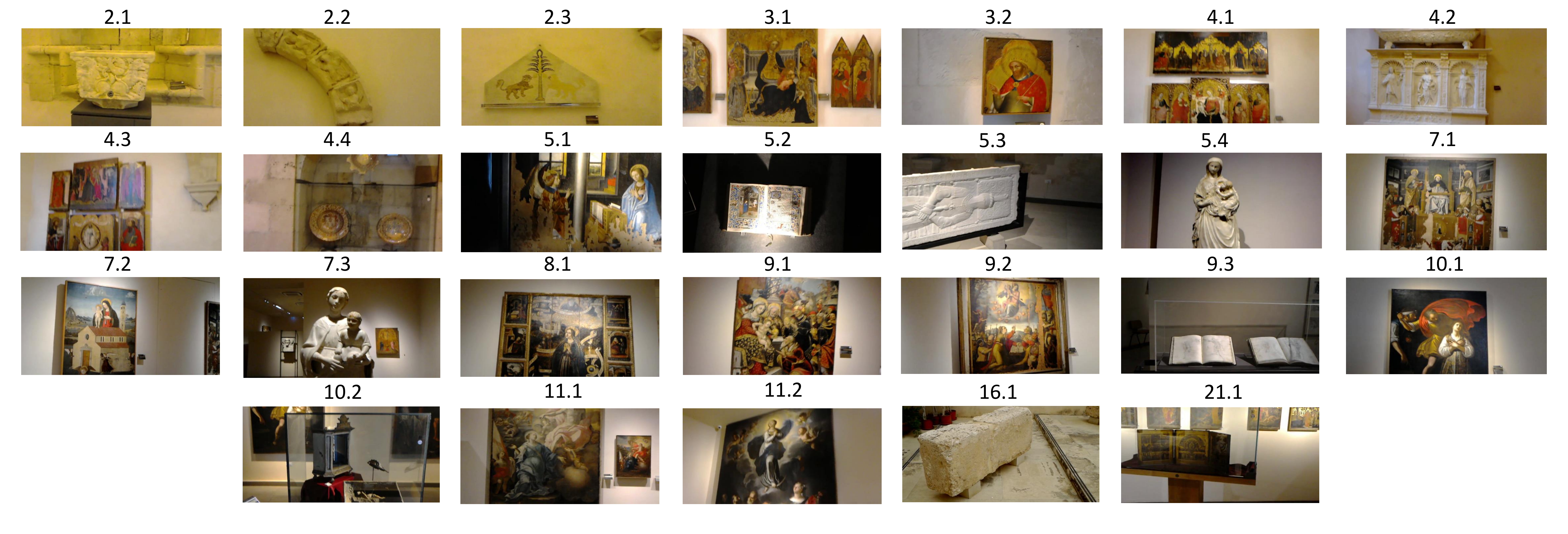}
\caption{Sample frames of points of interest of ``Palazzo Bellomo''.}
\label{fig:bellomoPOI}
\end{figure}

\begin{table*}[]
\centering
\caption{List of training videos of ``Palazzo Bellomo''.}
\label{tab:bellomotrain}
\scriptsize
\begin{tabular}{l|c|c|c|c}
\hline
\multicolumn{1}{c|}{\textbf{Name}} & \textbf{Time (s)} & \textbf{Storage (MB)} & \textbf{\#frame} & \textbf{\%frame} \\ \hline
1.0\_Sala1 & 124 & 156.229 & 3721 & 3,13\% \\
2.0\_Sala2 & 117 & 148.480 & 3525 & 2,96\% \\
3.0\_Sala3 & 100 & 125.924 & 3000 & 2,52\% \\
3.0\_Sala3\_S & 73 & 92.589 & 2200 & 1,85\% \\
4.0\_Sala4 & 97 & 122.941 & 2914 & 2,45\% \\
5.0\_Sala5 & 99 & 126.213 & 2992 & 2,51\% \\
6.0\_Sala6 & 87 & 110.451 & 2630 & 2,21\% \\
7.0\_Sala7 & 113 & 143.257 & 3402 & 2,86\% \\
8.0\_Sala8 & 147 & 186.470 & 4427 & 3,72\% \\
9.0\_Sala9 & 143 & 180.971 & 4298 & 3,61\% \\
10.0\_Sala10 & 71 & 90.697 & 2154 & 1,81\% \\
11.0\_Sala11 & 104 & 131.983 & 3145 & 2,64\% \\
12.0\_Sala12 & 82 & 103.785 & 2463 & 2,07\% \\
13.0\_Sala13 & 101 & 128.013 & 3040 & 2,55\% \\
14.0\_CortiledegliStemmi & 104 & 131.962 & 3131 & 2,63\% \\
14.0\_CortiledegliStemmi\_S & 90 & 113.822 & 2722 & 2,29\% \\
15.0\_SalaCarrozze & 108 & 136.968 & 3259 & 3,12\% \\
16.0\_CortileParisio & 124 & 156.605 & 3718 & 3,12\% \\
16.0\_Cortile\_Parisio\_S & 68 & 86.646 & 2062 & 1,73\% \\
17.0\_Biglietteria & 83 & 104.532 & 2489 & 2,09\% \\
17.0\_Biglietteria\_S & 53 & 68.071 & 1610 & 1,35\% \\
18.0\_Portico & 126 & 159.800 & 3791 & 3,18\% \\
18.0\_Portico\_S & 63 & 80.044 & 1910 & 1,60\% \\
19.0\_ScalaCatalana & 97 & 123.010 & 2918 & 2,45\% \\
19.0\_ScalaCatalana\_S & 116 & 110.063 & 3481 & 2,92\% \\
20.0\_Loggetta & 80 & 101.584 & 2425 & 2,04\% \\
20.0\_Loggetta\_S & 58 & 73.722 & 1744 & 1,46\% \\
21.0\_BoxSala8 & 85 & 107.540 & 2562 & 2,15\% \\
22.0\_AreaSosta & 64 & 81.934 & 1945 & 1,63\% \\
22.0\_Area\_Sosta\_S & 52 & 65.340 & 1560 & 1,31\% \\
2.1\_Sala2\_Acquasantiera & 54 & 68.445 & 1623 & 2,94\% \\
2.2\_Sala2\_FrammentiArchitett. & 46 & 58.265 & 1393 & 2,53\% \\
2.3\_Sala2\_LastraconLeoni & 47 & 60.199 & 1427 & 2,59\% \\
3.1\_Sala3\_MadonnainTrono & 65 & 83.198 & 1972 & 3,58\% \\
3.2\_Sala3\_FrammentoS.Leonardo & 37 & 47.061 & 1113 & 2,02\% \\
4.1\_Sala4\_MadonnainTrono & 75 & 94.767 & 2252 & 4,08\% \\
4.2\_Sala4\_MonumentoE.d'Aragona & 86 & 108.296 & 2580 & 4,68\% \\
4.3\_Sala4\_TrasfigurazioneCristo & 76 & 96.106 & 2277 & 4,13\% \\
4.4\_Sala4\_Piatti & 49 & 62.281 & 1474 & 2,67\% \\
5.1\_Sala5\_Annunciazione & 76 & 96.952 & 2295 & 4,16\% \\
5.2\_Sala5\_LibroD'OreMiniato & 46 & 59.011 & 1406 & 2,55\% \\
5.3\_Sala5\_LastraG.Cabastida & 100 & 127.023 & 3017 & 5,47\% \\
5.4\_Sala5\_MadonnadelCardillo & 61 & 77.568 & 1829 & 3,32\% \\
7.1\_Sala7\_DisputaS.Tommaso & 74 & 94.188 & 2234 & 4,05\% \\
7.2\_Sala7\_TraslazioneSantaCasa & 76 & 96.045 & 2281 & 4,14\% \\
7.3\_Sala7\_MadonnacolBambino & 90 & 113.202 & 2696 & 4,89\% \\
8.1\_Sala8\_ImmacolataConcezione & 82 & 104.570 & 2483 & 4,50\% \\
9.1\_Sala9\_AdorazionedeiMagi & 60 & 76.171 & 1803 & 3,27\% \\
9.2\_Sala9\_S.ElenaCostantinoeMadonna & 76 & 96.227 & 2283 & 4,14\% \\
9.3\_Sala9\_TaccuinidiDisegni & 70 & 89.647 & 2121 & 3,85\% \\
10.1\_Sala10\_MartirioS.Lucia & 58 & 74.248 & 1759 & 3,19\% \\
10.2\_Sala10\_VoltodiCristo & 64 & 80.896 & 1917 & 3,48\% \\
11.1\_Sala11\_MiracolodiS.Orsola & 66 & 84.297 & 2002 & 3,63\% \\
11.2\_Sala11\_Immacolata & 73 & 92.424 & 2196 & 3,98\% \\
16.1\_CortileParisio\_LapidiEbraiche & 85 & 108.098 & 2563 & 4,65\% \\
16.1\_CortileParisio\_LapidiEbraiche\_S & 67 & 85.173 & 2031 & 3,68\% \\
21.1\_BoxSala8\_StoriedellaGenesi & 70 & 88.300 & 2099 & 3,81\% \\ \hline
\textbf{AVG} & \textbf{81.72} & \textbf{103.02} & \textbf{2462.53} & \textbf{2.07\%} \\ \hline
\end{tabular}
\end{table*}

Table~\ref{tab:bellomotest} reports the list of the $10$ test videos acquired by volunteers visiting the cultural site. 
For each video, we report its total duration, the amount of required storage, the number of frames, the number of environments encountered in the video, as well as the sequence of environments, as visited by the subject acquiring the video.
All the training/test videos have a resolution of $1280 \times 720$ pixels and a frame-rate of $29.97$ fps.
We also include $191$ reference images related to the considered POIs to be used for one-shot image retrieval. 
The images are akin to pictures generally included in museums catalog. 
Figure~\ref{fig:bellomoHD} shows some examples of such reference images.

\begin{table*}[]
	\caption{List of test videos of ``Palazzo Bellomo''.}
	\label{tab:bellomotest}
	\centering
	\resizebox{\textwidth}{!}{%
		\begin{tabular}{l|c|c|c|c|c|c|c|c}
			\hline
			\textbf{Name}        & \textbf{Time (s)}                      & \textbf{MB}        & \multicolumn{1}{l|}{\textbf{\#Frame}} & \multicolumn{1}{l|}{\textbf{\%Frame}} & \multicolumn{1}{l|}{\textbf{\#Environments}}  & \textbf{Environments - Temporal sequence}                                                                                                                                                                                                                                                                                                                                                                                                                                                                                                                                                                                                                                                                                                                                                                                                                                                                                                                                                                                                                                                                                                                                                                                                                                                                                           \\ \hline
			Test1   & 1906                      & 2.400.360 & 57123                        & 11,13\%                      & 22                                                      & \begin{tabular}[c]{@{}l@{}}16-\textgreater{}17-\textgreater{}18-\textgreater{}1-\textgreater{}18-\textgreater{}3-\textgreater
			{}2-\textgreater{}3-\textgreater{}18-\textgreater{}4-\textgreater{}18-\textgreater{}14-\textgreater
			{}15-\textgreater{}14-\textgreater{}19-\textgreater{}20-\textgreater\\ 6-\textgreater{}5-\textgreater{}6-\textgreater{}7-\textgreater{}21-\textgreater{}8-\textgreater{}9-\textgreater{}22-\textgreater{}10-\textgreater{}11-\textgreater{}12-\textgreater{}13-\textgreater{}5-\textgreater{}6-\textgreater{}20-\textgreater{}19-\textgreater{}18-\textgreater 17\end{tabular}                                                                                                                                                                                                     \\ \hline
			Test2   & 1413                      & 1.435.096 & 42348                        & 8,25\%                       & 22                                                              & \begin{tabular}[c]{@{}l@{}}16-\textgreater{}17-\textgreater{}18-\textgreater{}1-\textgreater{}18-\textgreater{}3-\textgreater 2-\textgreater{}3-\textgreater{}18-\textgreater{}4-\textgreater{}18-\textgreater{}14-\textgreater 15-\textgreater{}14-\textgreater{}19-\textgreater{}20-\textgreater{}6-\textgreater{}5-\textgreater\\ 6-\textgreater{}7-\textgreater{}8-\textgreater{}21-\textgreater{}8-\textgreater{}9-\textgreater{}22-\textgreater 10-\textgreater{}11-\textgreater{}12-\textgreater{}13-\textgreater{}5-\textgreater{}6-\textgreater 20-\textgreater{}19-\textgreater{}18-\textgreater{}17\end{tabular}                                                                    \\ \hline
			Test3   & 1830                      & 2.304.410   & 54845                        & 10,69\%                      & 22                                                              & \begin{tabular}[c]{@{}l@{}}16-\textgreater{}17-\textgreater{}18-\textgreater{}1-\textgreater{}18-\textgreater{}3-\textgreater 2-\textgreater{}3-\textgreater{}18-\textgreater{}4-\textgreater{}18-\textgreater{}14-\textgreater 15-\textgreater{}14-\textgreater{}19-\textgreater{}20-\textgreater{}6-\textgreater{}5-\textgreater\\ 13-\textgreater{}12-\textgreater{}11-\textgreater{}10-\textgreater{}22-\textgreater{}9-\textgreater 8-\textgreater{}21-\textgreater{}8-\textgreater{}7-\textgreater{}6-\textgreater{}20-\textgreater{}19-\textgreater 18-\textgreater{}17\end{tabular}                                                                                                                                                                         \\ \hline
			Test4   & 1542                      & 1.942.200   & 46214                        & 5,49\%                       & 22                                                              & \begin{tabular}[c]{@{}l@{}}16-\textgreater{}17-\textgreater{}18-\textgreater{}1-\textgreater{}18-\textgreater{}3-\textgreater 2-\textgreater{}3-\textgreater{}18-\textgreater{}4-\textgreater{}18-\textgreater{}14-\textgreater 15-\textgreater{}14-\textgreater{}19-\textgreater{}20-\textgreater{}6-\textgreater{}5-\textgreater\\ 6-\textgreater{}7-\textgreater{}8-\textgreater{}21-\textgreater{}8-\textgreater{}9-\textgreater 22-\textgreater{}10-\textgreater{}11-\textgreater{}12-\textgreater{}13-\textgreater{}5-\textgreater 6-\textgreater{}20-\textgreater{}19-\textgreater{}18-\textgreater{}17\end{tabular}                                                                                                                                                                                 \\ \hline
			Test5  & 1034                      & 1.302.612   & 30989                        & 9,00\%                       & 22                                                             & \begin{tabular}[c]{@{}l@{}}16-\textgreater{}17-\textgreater{}18-\textgreater{}1-\textgreater{}18-\textgreater{}3-\textgreater 2-\textgreater{}3-\textgreater{}18-\textgreater{}4-\textgreater{}18-\textgreater{}14-\textgreater 15-\textgreater{}14-\textgreater{}19-\textgreater{}20-\textgreater{}6-\textgreater{}5-\textgreater\\ 6-\textgreater{}7-\textgreater{}8-\textgreater{}9-\textgreater{}22-\textgreater{}10-\textgreater 11-\textgreater{}12-\textgreater{}13-\textgreater{}5-\textgreater{}6-\textgreater{}20-\textgreater 19-\textgreater{}18-\textgreater{}17\end{tabular}                                                                                                                                                                                                                                                                                                           \\ \hline
			Test6  & 1949                      & 2.273.926   & 58411                        & 11,38\%                      & 22                                                               & \begin{tabular}[c]{@{}l@{}}16-\textgreater{}17-\textgreater{}8-\textgreater{}1-\textgreater{}18-\textgreater{}3-\textgreater 2-\textgreater{}3-\textgreater{}18-\textgreater{}14-\textgreater{}15-\textgreater{}14-\textgreater 19-\textgreater{}20-\textgreater{}6-\textgreater{}5-\textgreater{}13-\textgreater{}5-\textgreater\\ 6-\textgreater{}7-\textgreater{}8-\textgreater{}21-\textgreater{}8-\textgreater{}9-\textgreater{}22-\textgreater 10-\textgreater{}11-\textgreater{}12-\textgreater{}11-\textgreater{}10-\textgreater{}22-\textgreater 9-\textgreater{}8-\textgreater{}7-\textgreater{}6-\textgreater{}20-\textgreater\\{}19-\textgreater 18-\textgreater{}17\end{tabular}                                                                                                                                                                                                                                                                                   \\ \hline
			Test7  & 1332                      & 1.677.047   & 39920                        & 7,78\%                       & 22                                                             & \begin{tabular}[c]{@{}l@{}}16-\textgreater{}17-\textgreater{}14-\textgreater{}15-\textgreater{}14-\textgreater{}18-\textgreater 1-\textgreater{}18-\textgreater{}3-\textgreater{}2-\textgreater{}3-\textgreater{}18-\textgreater{}4-\textgreater 18-\textgreater{}19-\textgreater{}20-\textgreater{}6-\textgreater{}5-\textgreater\\{}6-\textgreater{}7-\textgreater 8-\textgreater{}21-\textgreater{}8-\textgreater{}9-\textgreater{}22-\textgreater{}10-\textgreater 11-\textgreater{}12-\textgreater{}13-\textgreater{}5-\textgreater{}6-\textgreater{}20-\textgreater 19-\textgreater{}18-\textgreater{}17-\textgreater{}16\end{tabular}                                                                                                                            \\ \hline
			Test8   & 3023                      & 3.806.383   & 90599                        & 17,65\%                      & 22                                                              & \begin{tabular}[c]{@{}l@{}}16-\textgreater{}17-\textgreater{}14-\textgreater{}5-\textgreater{}14-\textgreater{}18-\textgreater 4-\textgreater{}18-\textgreater{}3-\textgreater{}2-\textgreater{}3-\textgreater{}18-\textgreater{}1-\textgreater 18-\textgreater{}19-\textgreater{}20-\textgreater{}6-\textgreater{}5-\textgreater\\{}13-\textgreater 12-\textgreater{}11-\textgreater{}10-\textgreater{}22-\textgreater{}9-\textgreater{}8-\textgreater 21-\textgreater{}8-\textgreater{}7-\textgreater{}6-\textgreater{}20-\textgreater{}19-\textgreater 14\end{tabular}                                                                                                                                                                                                                \\ \hline
			Test9   & 2236 & 2.815.878   & 67013                        & 13,05\%                      & 22                                                               & \begin{tabular}[c]{@{}l@{}}16-\textgreater{}17-\textgreater{}18-\textgreater{}1-\textgreater{}18-\textgreater{}3-\textgreater 2-\textgreater{}3-\textgreater{}18-\textgreater{}4-\textgreater{}18-\textgreater{}14-\textgreater 15-\textgreater{}14-\textgreater{}19-\textgreater{}20-\textgreater{}6-\textgreater{}5-\textgreater\\ 13-\textgreater{}12-\textgreater{}11-\textgreater{}10-\textgreater{}22-\textgreater 9-\textgreater{}8-\textgreater{}21-\textgreater{}8-\textgreater{}7-\textgreater{}6-\textgreater{}20-\textgreater 19-\textgreater{}14-\textgreater{}17\end{tabular}                                                                                                                                                                                                                                                     \\ \hline
			Test10 & 858  & 1.080.389   & 25714                        & 5,01\%                       & 22                                                              & \begin{tabular}[c]{@{}l@{}}16-\textgreater{}17-\textgreater{}14-\textgreater{}19-\textgreater{}20-\textgreater{}6-\textgreater 7-\textgreater{}8-\textgreater{}21-\textgreater{}8-\textgreater{}9-\textgreater{}22-\textgreater{}10-\textgreater 11-\textgreater{}12-\textgreater{}13-\textgreater{}5-\textgreater{}6-\textgreater\\{}20-\textgreater 19-\textgreater{}18-\textgreater{}4-\textgreater{}18-\textgreater{}1-\textgreater{}18-\textgreater 3-\textgreater{}2-\textgreater{}3-\textgreater{}18\end{tabular}                                                                                                                                                                                                                                                                                                     
		\end{tabular}%
	}
\end{table*}

	\begin{figure}[t]
	\centering
	\includegraphics[width=8cm]{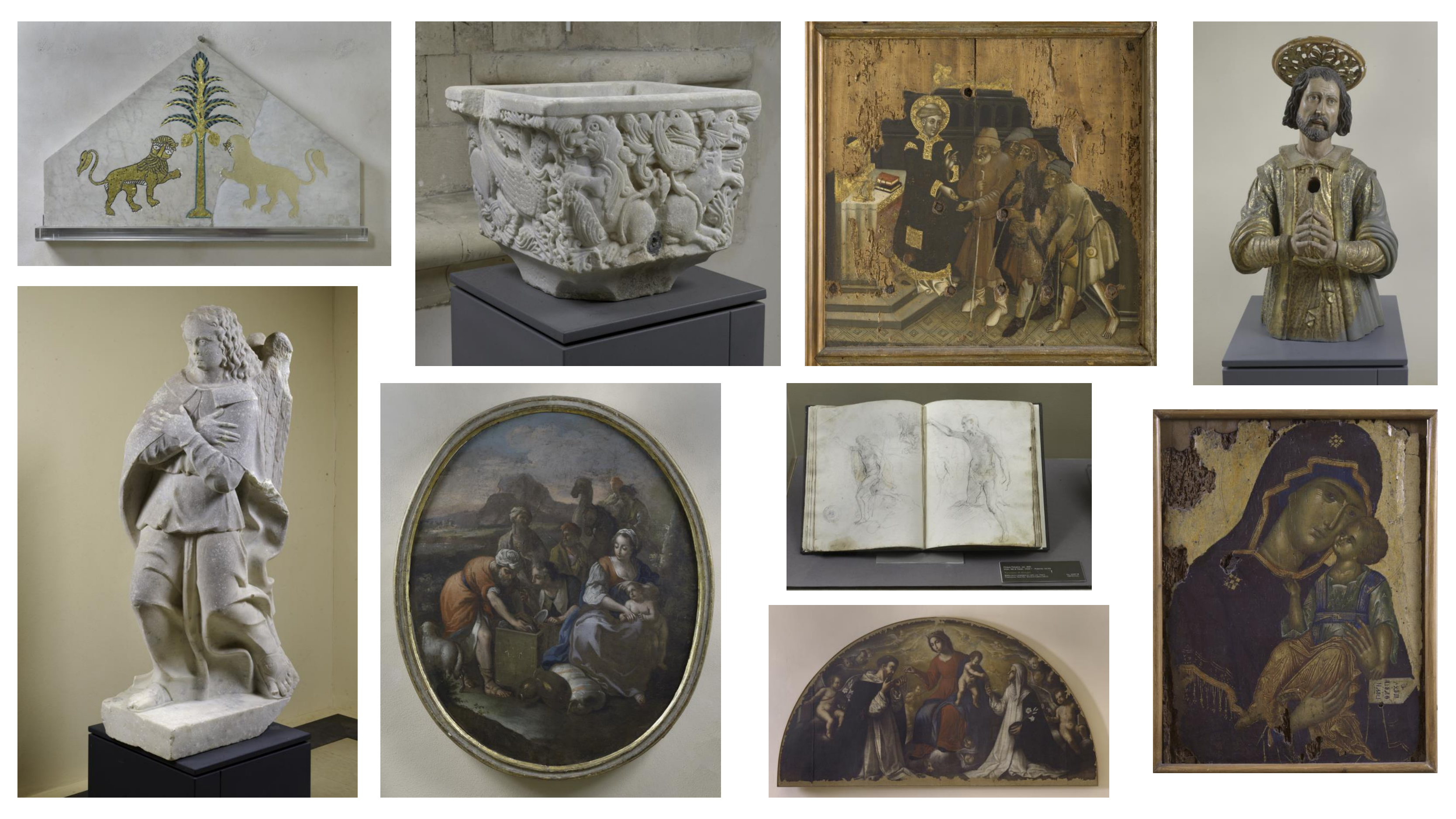}
	\caption{Sample references images related to the cultural site ``Palazzo Bellomo''.}
	\label{fig:bellomoHD}
\end{figure}

\paragraph{2) Monastero dei Benedettini}
This dataset is composed by $4$ environments and contains $35$ Points of Interest. Figure~\ref{fig:monenv} and Figure~\ref{fig:monPOI} report some frames related to the 4 different environments and some of the points of interest. Table~\ref{tab:montrainstats} reports details on the acquired training videos, highlighting the total duration of the videos, the required storage, the number of frames and the percentage of frames with respect to the whole training set. 
Training and validation videos have a resolution of $1216\times 684$ pixels and a frame-rate of $24$ fps. 
Five validation videos have been collected by asking volunteers to visit the cultural site with rules similarly to the one used for ``Palazzo Bellomo''. 
Table~\ref{tab:monval} shows the number of frames belonging to each video (left) and the number of frames belonging for each class (right).

	\begin{figure}[t]
	\centering
	\includegraphics[width=8cm]{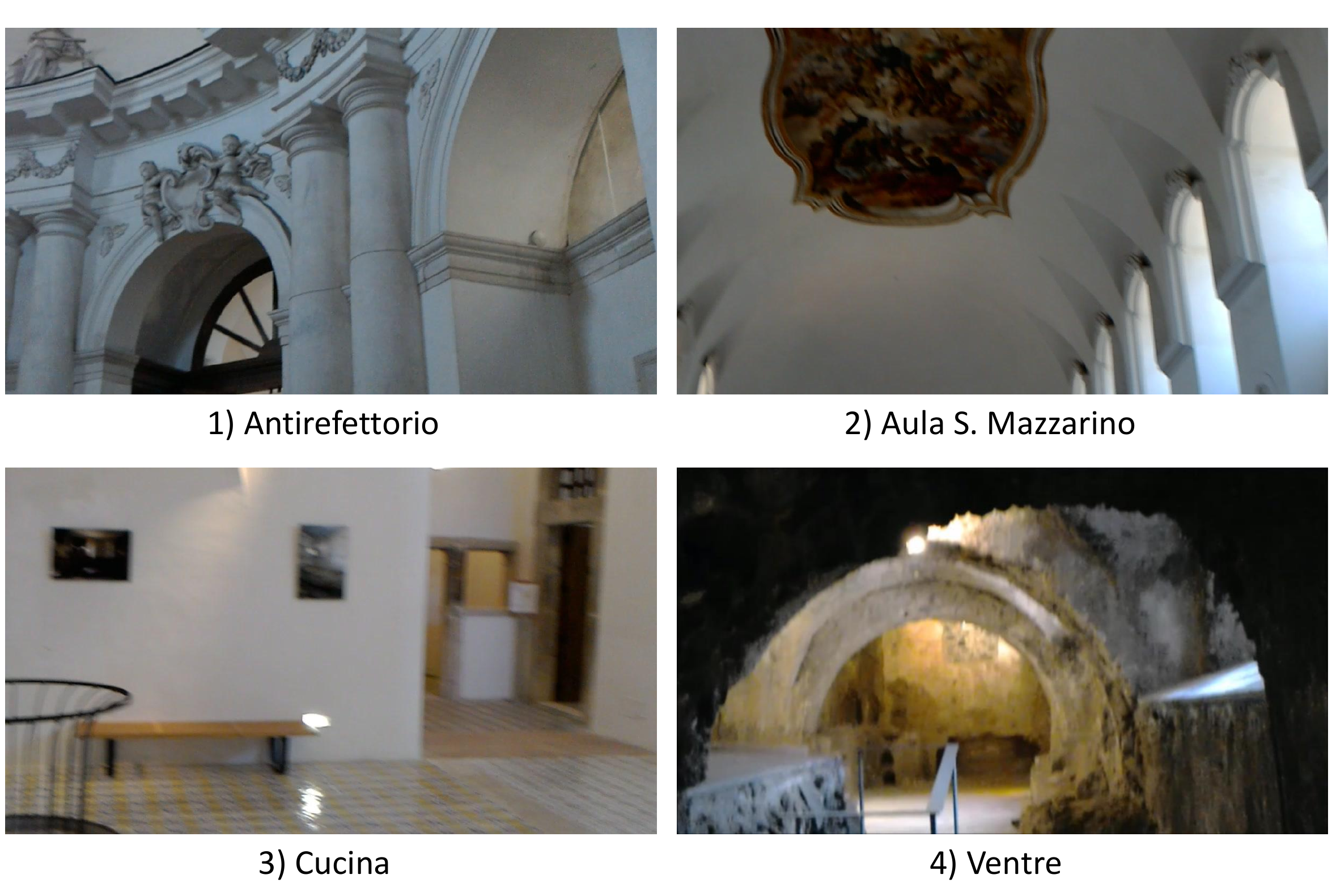}
	\caption{Sample frames from the 4 considered environments of ``Monastero dei Benedettini''.} \label{fig:monenv}
\end{figure}

	\begin{figure}[t]
	\centering
	\includegraphics[width=8cm]{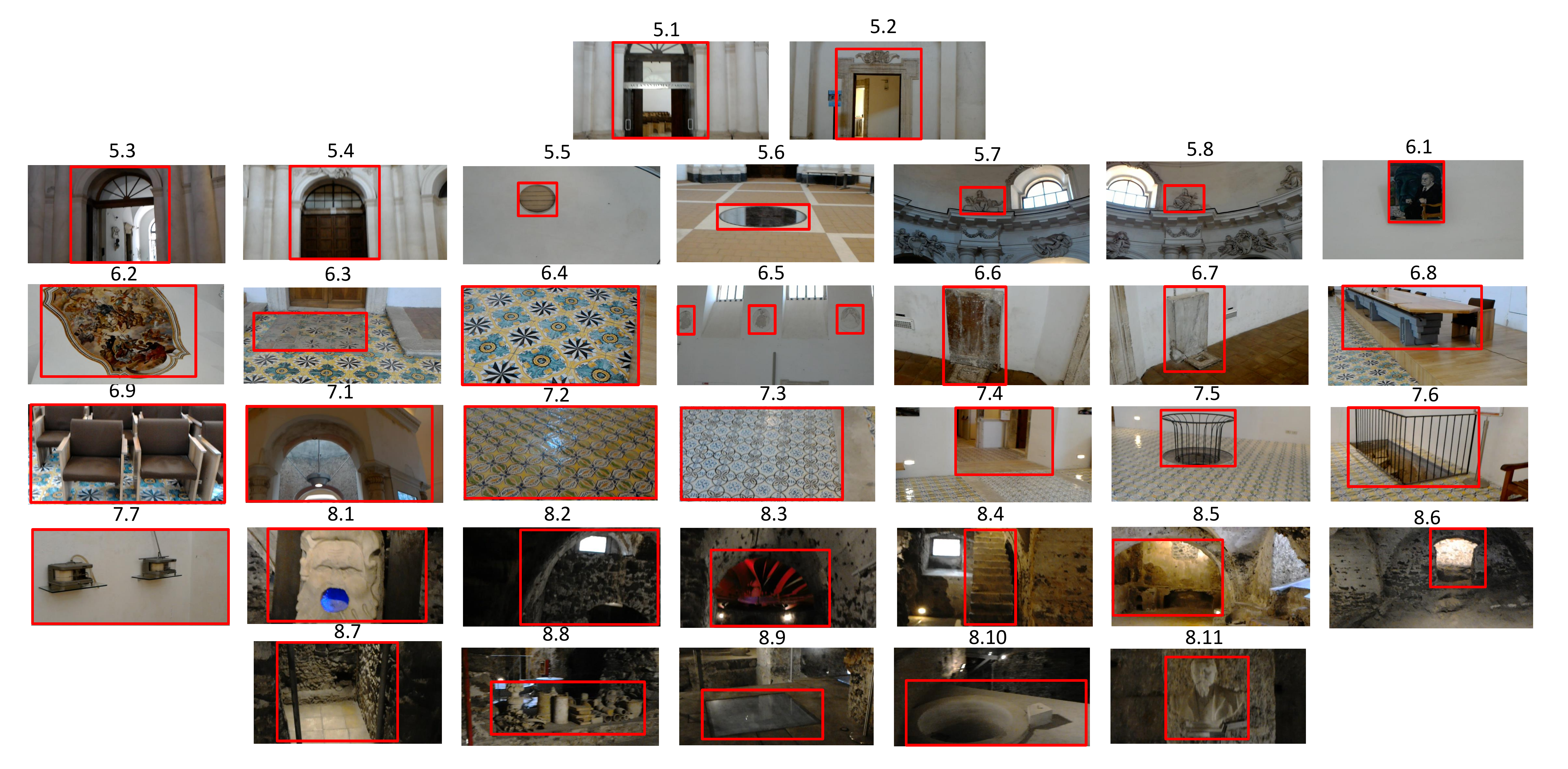}
	\caption{Sample frames from the 35 considered POIs of ``Monastero dei Benedettini'', with the related bounding box annotations.} \label{fig:monPOI}
\end{figure}

\begin{table*}[]
    \caption{List of training videos of ``Monastero dei Benedettini''}
		\label{tab:montrainstats}
	\centering	
	\begin{tabular}{l|c|c|c|c}
		\hline
		\multicolumn{1}{c|}{\textbf{Name}} & \textbf{Time (s)} & \textbf{Storage (MB)} & \textbf{\#frame} & \textbf{\%frame} \\ \hline
		5.0\_Antirefettorio       & 247        & 244.500     & 5933    & 1,21\%  \\
		5.0\_Antirefettorio1      & 263        & 260.395     & 6315    & 1,29\%  \\
		6.0\_SantoMazzarino       & 241        & 239.007     & 5800    & 1,18\%  \\
		7.0\_Cucina               & 239        & 237.268     & 5753    & 1,17\%  \\
		8.0\_Ventre               & 679        & 832.844     & 20385   & 4,15\%  \\
		5.1\_Antirefettorio\_PortaA.S.Mazz.Ap.    & 67         & 67.001      & 1630    &    0,33\%
	\\
	5.1\_Antirefettorio\_PortaA.S.Mazz.Ch.     & 74         & 73.589      & 1785    &    0,36\%
	\\
	5.2\_Antirefettorio\_PortaMuseoFab.Ap.     & 50         & 49.840      & 1211    &    0,25\%
	\\
	5.2\_Antirefettorio\_PortaMuseoFab.Ch.     & 62         & 61.846      & 1503    &   0,31\%
	\\
	5.3\_Antirefettorio\_PortaAntiref.           & 51         & 58.557      & 1537    &    0,31\%
	\\
	5.4\_Antirefettorio\_PortaRef.Piccolo           & 54         & 53.767      & 1306    &   0,27\%
	\\
	5.5\_Antirefettorio\_Cupola           & 57         & 56.089      & 1377    &   0,28\%
	\\
	5.6\_Antirefettorio\_AperturaPavimento        & 55         & 54.586      & 1322    &   0,27\%
	\\
	5.7\_Antirefettorio\_S.Agata          & 48         & 47.820      & 1165    &    0,24\%
	\\
	5.8\_Antirefettorio\_S.Scolastica           & 58         & 57.919      & 1407    &    0,29\%
	\\
	5.9\_Antirefettorio\_ArcoconFirma             & 62         & 76.700      & 1864    &    0,38\%
	\\
	5.10\_Antirefettorio\_BustoVaccarini           & 65         & 64.298      & 1563    &    0,32\%
	\\
	6.1\_SantoMazzarino\_QuadroS.Mazz.          & 71         & 70.401      & 1716    &     0,35\%
	\\
	6.2\_SantoMazzarino\_Affresco         & 213        & 211.424     & 5124    &   1,04\%
	\\
	6.3\_SantoMazzarino\_PavimentoOr.     & 99         & 98.691      & 2397    &     0,49\%
	\\
	6.4\_SantoMazzarino\_PavimentoRes.    & 69         & 69.148      & 1675    &    0,34\%
	\\
	6.5\_SantoMazzarino\_BassorilieviManc.     & 117        & 115.348     & 2823    &    0,57\%
	\\
	6.6\_SantoMazzarino\_LavamaniSx       & 151        & 149.882     & 3637    &     0,74\%
	\\
	6.7\_SantoMazzarino\_LavamaniDx       & 93         & 92.928      & 2256    &    0,46\%
	\\
	6.8\_SantoMazzarino\_TavoloRelatori      & 150        & 148.661     & 3603    &    0,73\%
	\\
	6.9\_SantoMazzarino\_Poltrone         & 108        & 107.374     & 2604    &    0,53\%
	\\
	7.1\_Cucina\_Edicola                  & 369        & 437.219     & 11086   &    2,26\%
	\\
	7.2\_Cucina\_PavimentoA               & 52         & 52.163      & 1268    &     0,26\%
	\\
	7.3\_Cucina\_PavimentoB               & 52         & 52.244      & 1266    &     0,26\%
	\\
	7.4\_Cucina\_PassavivandePavim.Orig.             & 81         & 80.733      & 1961    &     0,40\%
	\\
	7.5\_Cucina\_AperturaPav.1                    & 57         & 57.170      & 1385    &     0,28\%
	\\
	7.5\_Cucina\_AperturaPav.2                    & 53         & 52.875      & 1280    &     0,26\%
	\\
	7.5\_Cucina\_AperturaPav.3                    & 62         & 62.320      & 1509    &     0,31\%
	\\
	7.6\_Cucina\_Scala                    & 77         & 76.587      & 1856    &     0,38\%
	\\
	7.7\_Cucina\_SalaMetereologica         & 156        & 154.394     & 3748    &   0,76\%
	\\
	8.1\_Ventre\_Doccione                 & 103        & 102.683     & 2492    &      0,51\%
	\\
	8.2\_Ventre\_VanoRacc.Cenere               & 126        & 124.837     & 3026    &     0,62\%
	\\
	8.3\_Ventre\_SalaRossa                & 300        & 296.792     & 7202    &     1,47\%
	\\
	8.4\_Ventre\_ScalaCucina              & 214        & 212.372     & 5152    &     1,05\%
	\\
	8.5\_Ventre\_CucinaProvv.             & 148        & 146.379     & 3553    &    0,72\%
	\\
	8.6\_Ventre\_Ghiacciaia               & 69         & 68.562      & 1668    &      0,34\%
	\\
	8.6\_Ventre\_Ghiacciaia1              & 266        & 263.817     & 6398    &    1,30\%
	\\
	8.7\_Ventre\_Latrina                    & 102        & 100.542     & 2468    &     0,50\%
	\\
	8.8\_Ventre\_OssaScarti                     & 154        & 152.861     & 3713    &     0,76\%
	\\
	8.8\_Ventre\_OssaScarti1                    & 36         & 36.345      & 886     &    0,18\%
	\\
	8.9\_Ventre\_Pozzo                    & 300        & 297.167     & 7209    &    1,47\%
	\\
	8.10\_Ventre\_Cisterna                & 57         & 56.572      & 1384    &      0,28\%
	\\
	8.11\_Ventre\_BustoP.Tacchini           & 110        & 109.520     & 2662    &    0,54\%
	\\ \hline
	\textbf{AVG} & \textbf{133.06} & \textbf{137.38} & \textbf{3351.31} & \textbf{1.04\%} \\ \hline
	\end{tabular}
	
		\end{table*}

\begin{table*}[t]
\caption{List of validation videos of ``Monastero dei Benedettini''.}
\label{tab:monval}
\begin{tabular}{lcl|lc}
\textbf{Name}  & \multicolumn{1}{l}{\textbf{\#frame}} & \textbf{} & \textbf{Class}      & \multicolumn{1}{l}{\textbf{\#frame}} \\ \hline
Test1          & 4141                                 &           & 1 Antirefettorio    & 88613                                \\
Test3          & 18678                                &           & 2 Aula S. Mazzarino & 8395                                 \\
Test4          & 13731                                &           & 3 Cucina            & 9712                                 \\
Test5          & 15958                                &           & 4 Ventre            & 20513                                \\
Test7          & 1124                                 &           & Negatives           & 6399                                 \\ \hline
\textbf{Total} & \multicolumn{1}{l}{\textbf{53632}}   & \textbf{} & \textbf{Total}      & \multicolumn{1}{l}{\textbf{53632}}  
\end{tabular}
\end{table*}

Additionally, we collected $60$ test videos by asking real visitors to freely visit the cultural site. 
No specific instructions have been given to the visitors, who were free to explore the $4$ environments and the $35$ POIs.
The $60$ test videos have a resolution of $1408 \times 792$ pixels and a frame-rate of $30.03\ fps$. 
The average number of frames for each video is \textit{39296}.
We also include $35$ reference images related to the considered POIs to be used for one-shot image retrieval. 
Figure~\ref{fig:monHD} shows some example of reference images. 

	\begin{figure}[t]
	\centering
	\includegraphics[width=8cm]{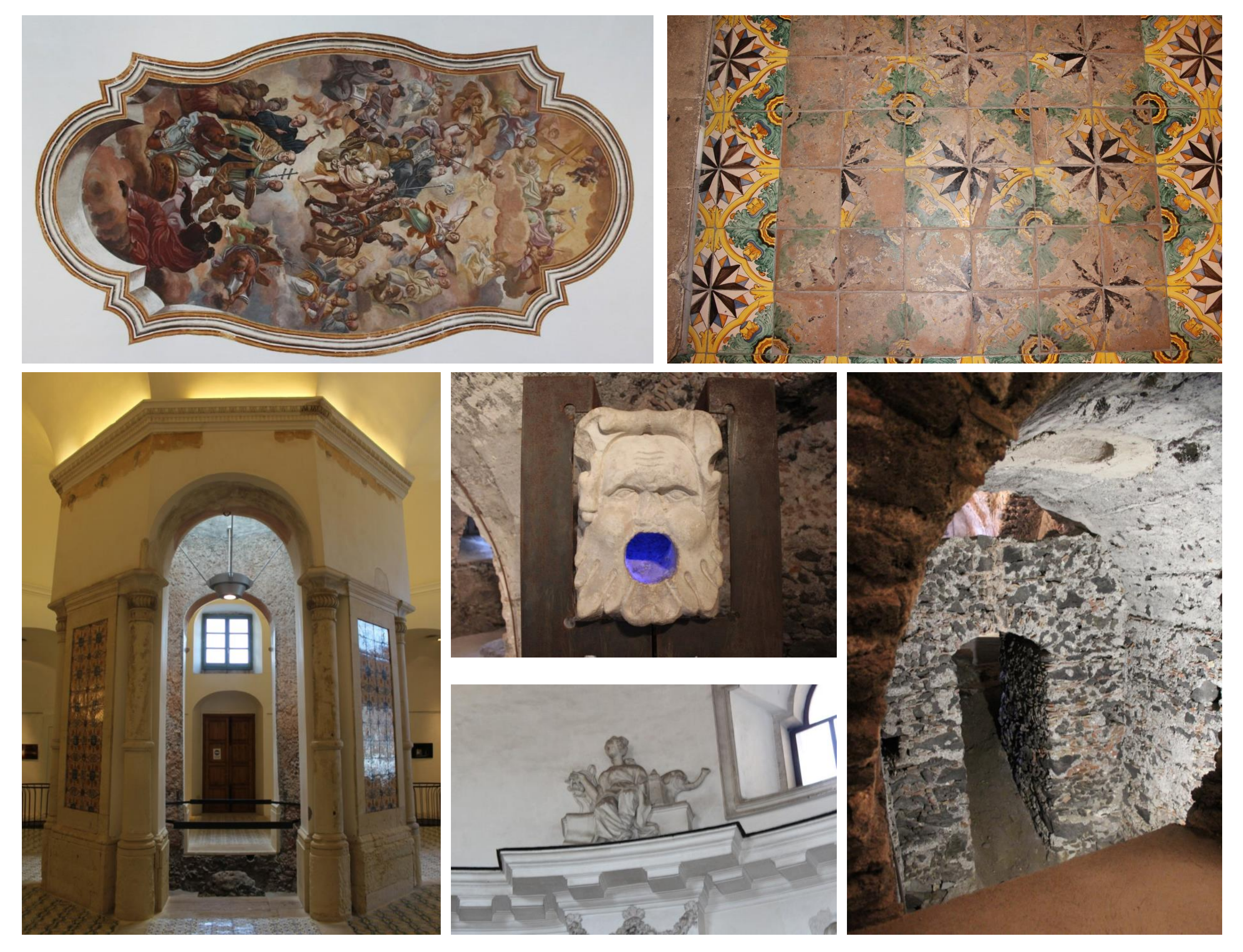}
	\caption{Sample references images related to the  ``Monastero dei Benedettini''.} \label{fig:monHD}
\end{figure}

\subsubsection{Annotations} 
\label{sec:data_annotations}
\paragraph{Temporal Labels} All test and validation videos have been temporally labeled to indicate in every frame the environment in which the visitor is located and the currently observed point of interest, if any. 
If the visitor is not located in any of the considered environments or they are not observing any of the considered POIs we mark as that frame as ``negative'' (Figure~\ref{fig:neg}). 

	\begin{figure}[t]
	\centering
	\includegraphics[width=8cm]{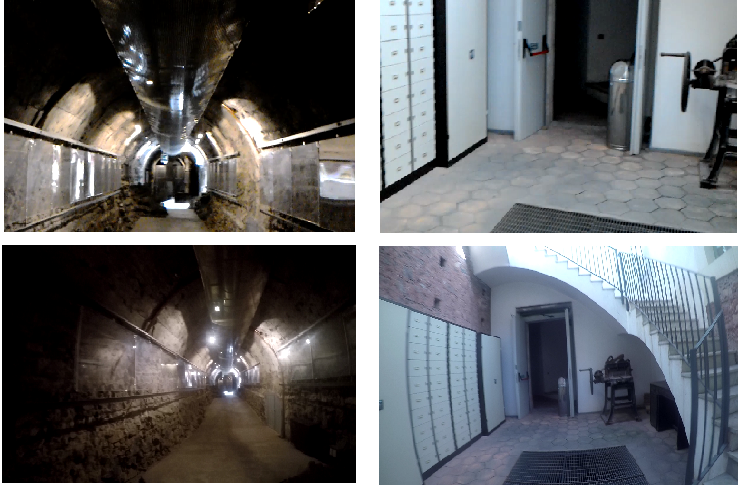}
	\caption{Sample frames from ``Monastero dei Benedettigni'' marked as ``negative locations''.} \label{fig:neg}
\end{figure}

\paragraph{Bounding Box Annotations}
A subset of frames from the dataset (sampled at at $1$ fps) has been labeled with bounding boxes indicating the presence and locations of all POIs. 
Figure~\ref{fig:monPOI} shows some example of labeled frames from the training set of the ``Monastero dei Benedettini''. 

The EGO-CH dataset is publicy available at our website: \textit{http://iplab.dmi.unict.it/EGO-CH/}. The dataset can be used only for reasearch purposes and is available upon request to the authors.

\subsection{\uppercase{Experimental Details}}
\label{sec:challenges}

\noindent
In this section, we present additional details on baseline experiments related to the $4$ tasks proposed in the paper.

\subsubsection{\textbf{Room-based Localization}}

\paragraph{1) Palazzo Bellomo}
We split the Training Set into two subsets to train and validate the VGG-19 for the ``Discrimination" stage (no ``negative" frames are used for training). 
Table~\ref{tab:bellomo_trainval} reports the number of frames belonging to the two subsets for each of the $22$ considered environments.

\begin{table}[t]
	\centering
	\caption{Number of frames belonging to the two subsets (Training/Validation) to train the CNN for ``Palazzo Bellomo''.}
	\label{tab:bellomo_trainval}
	\begin{tabular}{l|c|c}
		\hline
		& \multicolumn{1}{l|}{\textbf{Training}} & \multicolumn{1}{l}{\textbf{Validation}} \\ \hline
		1 Sala1           & 2605                                   & 1116                                    \\
		2 Sala2           & 5578                                   & 2390                                    \\
		3 Sala3           & 5800                                   & 2486                                    \\
		4 Sala4           & 8048                                   & 3449                                    \\
		5 Sala5           & 8023                                   & 3438                                    \\
		6 Sala6           & 1841                                   & 789                                     \\
		7 Sala7           & 7429                                   & 3184                                    \\
		8 Sala8           & 4837                                   & 2073                                    \\
		9 Sala9           & 7354                                   & 3152                                    \\
		10 Sala10         & 4081                                   & 1749                                    \\
		11 Sala11         & 5140                                   & 2203                                    \\
		12 Sala12         & 1724                                   & 739                                     \\
		13 Sala13         & 2128                                   & 912                                     \\
		14 Cortile degli Stemmi  & 4097                                   & 1756                                    \\
		15 Sala delle Carrozze   & 2281                                   & 978                                     \\
		16 Cortile Parisio & 7262                                   & 3112                                    \\
		17 Biglietteria   & 2869                                   & 1230                                    \\
		18 Portico        & 3991                                   & 1710                                    \\
		19 Scala Catalana  & 4479                                   & 1920                                    \\
		20 Loggetta       & 2918                                   & 1251                                    \\
		21 Box Sala8       & 3263                                   & 1398                                    \\
		22 Area Sosta      & 2454                                   & 1052                                    \\ \hline
		\textbf{Total}   & \textbf{98202}                         & \textbf{42084}                         
	\end{tabular}
\end{table}

We report the results obtained by the baseline on the $9$ test videos (``Test3'' has been used for validation) in Table~\ref{tab:HMM_FF1_Bellomo} considering the $FF1$ score metric, and in in Table~\ref{tab:HMM_ASF1_Bellomo} considering the $ASF1$ score.
As example, Figure~\ref{fig:seg_7} illustrate qualitatively the segmentation result of the baseline on ``Test7''. 
Figure~\ref{fig:bellomo_cf} reports the confusion matrix of the baseline on the test set.

\begin{table*}[]
	\centering
		\caption{Detailed results of the $9$ test videos of ``Palazzo Bellomo'' using the $FF_1$ score.
			The ``/'' sign indicates that no samples from that class was present in the test video.}
	\label{tab:HMM_FF1_Bellomo}
	\scriptsize
		\begin{tabular}{lcccccccccc}
			\multicolumn{11}{c}{\textbf{$FF_1$ score}}                                                                                                                                                                                                                                                                                                   \\ \hline
			\multicolumn{1}{l|}{Class}                 & \multicolumn{1}{l}{Test1} & \multicolumn{1}{l}{Test2} & \multicolumn{1}{l}{Test4} & \multicolumn{1}{l}{Test5} & \multicolumn{1}{l}{Test6} & \multicolumn{1}{l}{Test7} & \multicolumn{1}{l}{Test8} & \multicolumn{1}{l}{Test9} & \multicolumn{1}{l|}{Test10} & \multicolumn{1}{l}{AVG} \\ \hline
			\multicolumn{1}{l|}{1\_Sala1}               & 0,16                      & 0,00                      & 0,81                      & 0,96                      & 0,86                      & 0,96                      & 0,90                      & 0,85                      & \multicolumn{1}{c|}{0,92}   & 0,71                    \\
			\multicolumn{1}{l|}{2\_Sala2}               & 0,78                      & 0,67                      & 0,96                      & 0,96                      & 0,99                      & 0,99                      & 0,97                      & 0,97                      & \multicolumn{1}{c|}{0,96}   & 0,92                    \\
			\multicolumn{1}{l|}{3\_Sala3}               & 0,68                      & 0,75                      & 0,97                      & 0,87                      & 0,72                      & 0,96                      & 0,83                      & 0,89                      & \multicolumn{1}{c|}{0,91}   & 0,84                    \\
			\multicolumn{1}{l|}{4\_Sala4}               & 0,89                      & 0,96                      & 0,93                      & 0,91                      & /                         & 0,98                      & 0,86                      & 0,91                      & \multicolumn{1}{c|}{0,95}   & 0,92                    \\
			\multicolumn{1}{l|}{5\_Sala5}               & 0,90                      & 0,94                      & 0,98                      & 0,95                      & 0,89                      & 0,95                      & 0,95                      & 0,95                      & \multicolumn{1}{c|}{0,97}   & 0,94                    \\
			\multicolumn{1}{l|}{6\_Sala6}               & 0,84                      & 0,86                      & 0,80                      & 0,59                      & 0,76                      & 0,57                      & 0,93                      & 0,93                      & \multicolumn{1}{c|}{0,68}   & 0,77                    \\
			\multicolumn{1}{l|}{7\_Sala7}               & 0,99                      & 0,95                      & 0,99                      & 0,88                      & 0,84                      & 0,99                      & 0,92                      & 0,93                      & \multicolumn{1}{c|}{0,97}   & 0,94                    \\
			\multicolumn{1}{l|}{8\_Sala8}               & 0,85                      & 0,95                      & 0,96                      & 0,91                      & 0,67                      & 0,90                      & 0,84                      & 0,95                      & \multicolumn{1}{c|}{0,93}   & 0,89                    \\
			\multicolumn{1}{l|}{9\_Sala9}               & 0,86                      & 0,97                      & 0,95                      & 0,90                      & 0,76                      & 0,93                      & 0,94                      & 0,94                      & \multicolumn{1}{c|}{0,90}   & 0,91                    \\
			\multicolumn{1}{l|}{10\_Sala10}             & 0,87                      & 0,96                      & 0,96                      & 0,97                      & 0,00                      & 0,98                      & 0,95                      & 0,97                      & \multicolumn{1}{c|}{0,90}   & 0,84                    \\
			\multicolumn{1}{l|}{11\_Sala11}             & 0,86                      & 0,96                      & 0,97                      & 0,97                      & 0,00                      & 0,97                      & 0,94                      & 0,96                      & \multicolumn{1}{c|}{0,96}   & 0,84                    \\
			\multicolumn{1}{l|}{12\_Sala12}             & 0,82                      & 0,91                      & 0,86                      & 0,94                      & 0,00                      & 0,88                      & 0,96                      & 0,94                      & \multicolumn{1}{c|}{0,88}   & 0,80                    \\
			\multicolumn{1}{l|}{13\_Sala13}             & 0,74                      & 0,94                      & 0,91                      & 0,85                      & 0,00                      & 0,95                      & 0,97                      & 0,92                      & \multicolumn{1}{c|}{0,94}   & 0,80                    \\
			\multicolumn{1}{l|}{14\_CortiledegliStemmi} & 0,92                      & 0,80                      & 0,93                      & 0,89                      & 0,73                      & 0,92                      & 0,97                      & 0,94                      & \multicolumn{1}{c|}{0,57}   & 0,85                    \\
			\multicolumn{1}{l|}{15\_SalaCarrozze}       & 0,91                      & 0,89                      & 0,96                      & 0,93                      & 0,90                      & 0,90                      & 0,85                      & 0,96                      & \multicolumn{1}{c|}{}       & 0,91                    \\
			\multicolumn{1}{l|}{16\_CortileParisio}     & 0,74                      & 0,50                      & 0,92                      & 0,71                      & 0,48                      & 0,91                      & 0,99                      & 0,74                      & \multicolumn{1}{c|}{0,72}   & 0,75                    \\
			\multicolumn{1}{l|}{17\_Biglietteria}       & 0,64                      & 0,79                      & 0,81                      & 0,49                      & 0,74                      & 0,55                      & 0,66                      & 0,61                      & \multicolumn{1}{c|}{0,53}   & 0,65                    \\
			\multicolumn{1}{l|}{18\_Portico}            & 0,71                      & 0,42                      & 0,77                      & 0,70                      & 0,70                      & 0,73                      & 0,71                      & 0,75                      & \multicolumn{1}{c|}{0,72}   & 0,69                    \\
			\multicolumn{1}{l|}{19\_ScalaCatalana}      & 0,70                      & 0,78                      & 0,80                      & 0,77                      & 0,39                      & 0,86                      & 0,83                      & 0,95                      & \multicolumn{1}{c|}{0,76}   & 0,76                    \\
			\multicolumn{1}{l|}{20\_Loggetta}           & 0,62                      & 0,39                      & 0,75                      & 0,58                      & 0,67                      & 0,77                      & 0,84                      & 0,94                      & \multicolumn{1}{c|}{0,81}   & 0,71                    \\
			\multicolumn{1}{l|}{21\_BoxSala8}           & 0,97                      & 0,97                      & 0,98                      & /                         & 0,79                      & 0,97                      & 0,99                      & 0,94                      & \multicolumn{1}{c|}{0,94}   & 0,94                    \\
			\multicolumn{1}{l|}{22\_AreaSosta}          & 0,24                      & 0,81                      & 0,56                      & 0,46                      & 0,00                      & 0,87                      & 0,56                      & 0,78                      & \multicolumn{1}{c|}{0,77}   & 0,56                    \\ \hline
			\multicolumn{1}{c|}{mFF1}                   & 0,76                      & 0,78                      & 0,89                      & 0,82                      & 0,57                      & 0,89                      & 0,88                      & 0,90                      & \multicolumn{1}{c|}{0,84}   & 0,81                   
		\end{tabular}%

\end{table*}

\begin{table*}[]
	\centering
		\caption{Detailed results of the $9$ test videos of ``Palazzo Bellomo'' using the $ASF_1$ score.
		The ``/'' sign indicates that no samples from that class was present in the test video.}
	\label{tab:HMM_ASF1_Bellomo}
	\scriptsize
		\begin{tabular}{lcccccccccc}
			\multicolumn{11}{c}{\textbf{$ASF_1$}}                                                                                                                                                                                                                                                                                                  \\ \hline
			\multicolumn{1}{l|}{Class}                 & \multicolumn{1}{l}{Test1} & \multicolumn{1}{l}{Test2} & \multicolumn{1}{l}{Test4} & \multicolumn{1}{l}{Test5} & \multicolumn{1}{l}{Test6} & \multicolumn{1}{l}{Test7} & \multicolumn{1}{l}{Test8} & \multicolumn{1}{l}{Test9} & \multicolumn{1}{l|}{Test10} & \multicolumn{1}{l}{AVG} \\ \hline
			\multicolumn{1}{l|}{1\_Sala1}               & 0,07                      & 0,00                      & 0,25                      & 0,92                      & 0,15                      & 0,92                      & 0,62                      & 0,55                      & \multicolumn{1}{c|}{0,84}   & 0,48                    \\
			\multicolumn{1}{l|}{2\_Sala2}               & 0,32                      & 0,46                      & 0,92                      & 0,92                      & 0,97                      & 0,98                      & 0,65                      & 0,94                      & \multicolumn{1}{c|}{0,92}   & 0,79                    \\
			\multicolumn{1}{l|}{3\_Sala3}               & 0,18                      & 0,34                      & 0,66                      & 0,63                      & 0,19                      & 0,88                      & 0,16                      & 0,66                      & \multicolumn{1}{c|}{0,83}   & 0,50                    \\
			\multicolumn{1}{l|}{4\_Sala4}               & 0,58                      & 0,92                      & 0,41                      & 0,45                      & /                         & 0,95                      & 0,21                      & 0,28                      & \multicolumn{1}{c|}{0,89}   & 0,59                    \\
			\multicolumn{1}{l|}{5\_Sala5}               & 0,57                      & 0,56                      & 0,84                      & 0,77                      & 0,27                      & 0,72                      & 0,49                      & 0,64                      & \multicolumn{1}{c|}{0,94}   & 0,64                    \\
			\multicolumn{1}{l|}{6\_Sala6}               & 0,47                      & 0,61                      & 0,44                      & 0,35                      & 0,32                      & 0,40                      & 0,87                      & 0,80                      & \multicolumn{1}{c|}{0,44}   & 0,52                    \\
			\multicolumn{1}{l|}{7\_Sala7}               & 0,97                      & 0,91                      & 0,97                      & 0,28                      & 0,07                      & 0,97                      & 0,21                      & 0,16                      & \multicolumn{1}{c|}{0,94}   & 0,61                    \\
			\multicolumn{1}{l|}{8\_Sala8}               & 0,40                      & 0,86                      & 0,84                      & 0,59                      & 0,13                      & 0,68                      & 0,70                      & 0,74                      & \multicolumn{1}{c|}{0,85}   & 0,64                    \\
			\multicolumn{1}{l|}{9\_Sala9}               & 0,19                      & 0,93                      & 0,90                      & 0,45                      & 0,32                      & 0,18                      & 0,64                      & 0,14                      & \multicolumn{1}{c|}{0,49}   & 0,47                    \\
			\multicolumn{1}{l|}{10\_Sala10}             & 0,28                      & 0,92                      & 0,92                      & 0,93                      & 0,00                      & 0,96                      & 0,42                      & 0,94                      & \multicolumn{1}{c|}{0,81}   & 0,69                    \\
			\multicolumn{1}{l|}{11\_Sala11}             & 0,21                      & 0,92                      & 0,94                      & 0,94                      & 0,00                      & 0,94                      & 0,13                      & 0,21                      & \multicolumn{1}{c|}{0,92}   & 0,58                    \\
			\multicolumn{1}{l|}{12\_Sala12}             & 0,37                      & 0,82                      & 0,76                      & 0,89                      & 0,00                      & 0,79                      & 0,91                      & 0,63                      & \multicolumn{1}{c|}{0,78}   & 0,66                    \\
			\multicolumn{1}{l|}{13\_Sala13}             & 0,32                      & 0,88                      & 0,83                      & 0,74                      & 0,00                      & 0,90                      & 0,94                      & 0,40                      & \multicolumn{1}{c|}{0,89}   & 0,66                    \\
			\multicolumn{1}{l|}{14\_CortiledegliStemmi} & 0,78                      & 0,61                      & 0,83                      & 0,79                      & 0,16                      & 0,84                      & 0,72                      & 0,69                      & \multicolumn{1}{c|}{0,39}   & 0,64                    \\
			\multicolumn{1}{l|}{15\_SalaCarrozze}       & 0,49                      & 0,80                      & 0,92                      & 0,87                      & 0,37                      & 0,82                      & 0,17                      & 0,91                      & \multicolumn{1}{c|}{/}      & 0,67                    \\
			\multicolumn{1}{l|}{16\_CortileParisio}     & 0,31                      & 0,22                      & 0,65                      & 0,41                      & 0,16                      & 0,80                      & 0,98                      & 0,47                      & \multicolumn{1}{c|}{0,54}   & 0,50                    \\
			\multicolumn{1}{l|}{17\_Biglietteria}       & 0,41                      & 0,67                      & 0,69                      & 0,32                      & 0,45                      & 0,29                      & 0,38                      & 0,37                      & \multicolumn{1}{c|}{0,36}   & 0,44                    \\
			\multicolumn{1}{l|}{18\_Portico}            & 0,54                      & 0,28                      & 0,60                      & 0,43                      & 0,51                      & 0,58                      & 0,56                      & 0,49                      & \multicolumn{1}{c|}{0,58}   & 0,51                    \\
			\multicolumn{1}{l|}{19\_ScalaCatalana}      & 0,61                      & 0,48                      & 0,66                      & 0,64                      & 0,41                      & 0,72                      & 0,65                      & 0,91                      & \multicolumn{1}{c|}{0,62}   & 0,63                    \\
			\multicolumn{1}{l|}{20\_Loggetta}           & 0,43                      & 0,31                      & 0,37                      & 0,46                      & 0,23                      & 0,63                      & 0,70                      & 0,87                      & \multicolumn{1}{c|}{0,64}   & 0,51                    \\
			\multicolumn{1}{l|}{21\_BoxSala8}           & 0,65                      & 0,94                      & 0,96                      & /                         & 0,34                      & 0,94                      & 0,98                      & 0,64                      & \multicolumn{1}{c|}{0,88}   & 0,79                    \\
			\multicolumn{1}{l|}{22\_AreaSosta}          & 0,24                      & 0,69                      & 0,42                      & 0,47                      & 0,00                      & 0,76                      & 0,41                      & 0,64                      & \multicolumn{1}{c|}{0,62}   & 0,47                    \\ \hline
			\multicolumn{1}{c|}{mASF1}                  & 0,43                      & 0,64                      & 0,72                      & 0,63                      & 0,24                      & 0,76                      & 0,57                      & 0,59                      & \multicolumn{1}{c|}{0,72}   & 0,59                   
		\end{tabular}%

\end{table*}

\begin{figure}
	\centering
	\includegraphics[width=8cm]{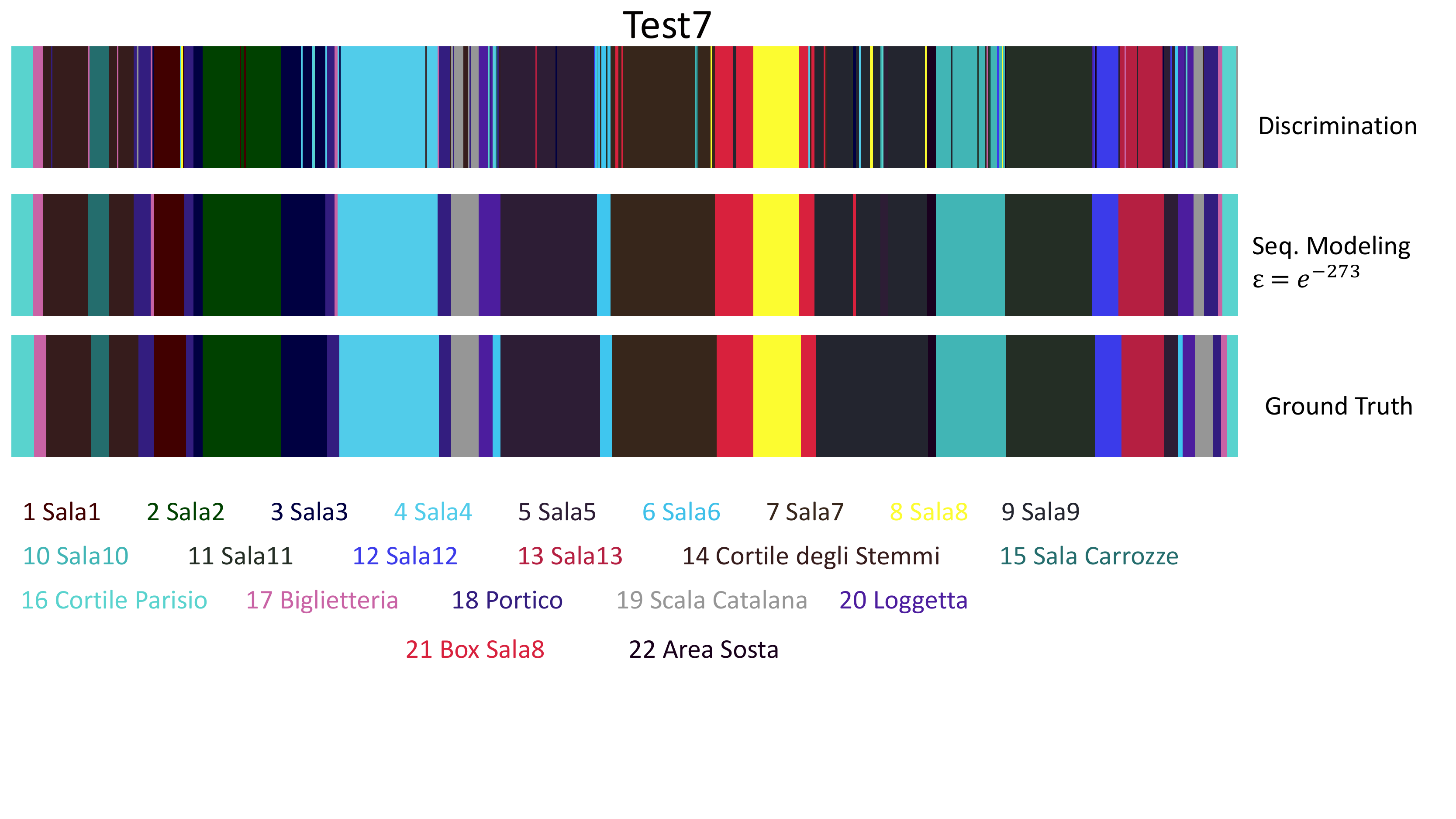}
	\caption{Color-coded segmentations for the test video "Test7" of ``Palazzo Bellomo''.}
	\label{fig:seg_7}
\end{figure}

\begin{figure*}[t]
	\centering
	\includegraphics[width=13cm]{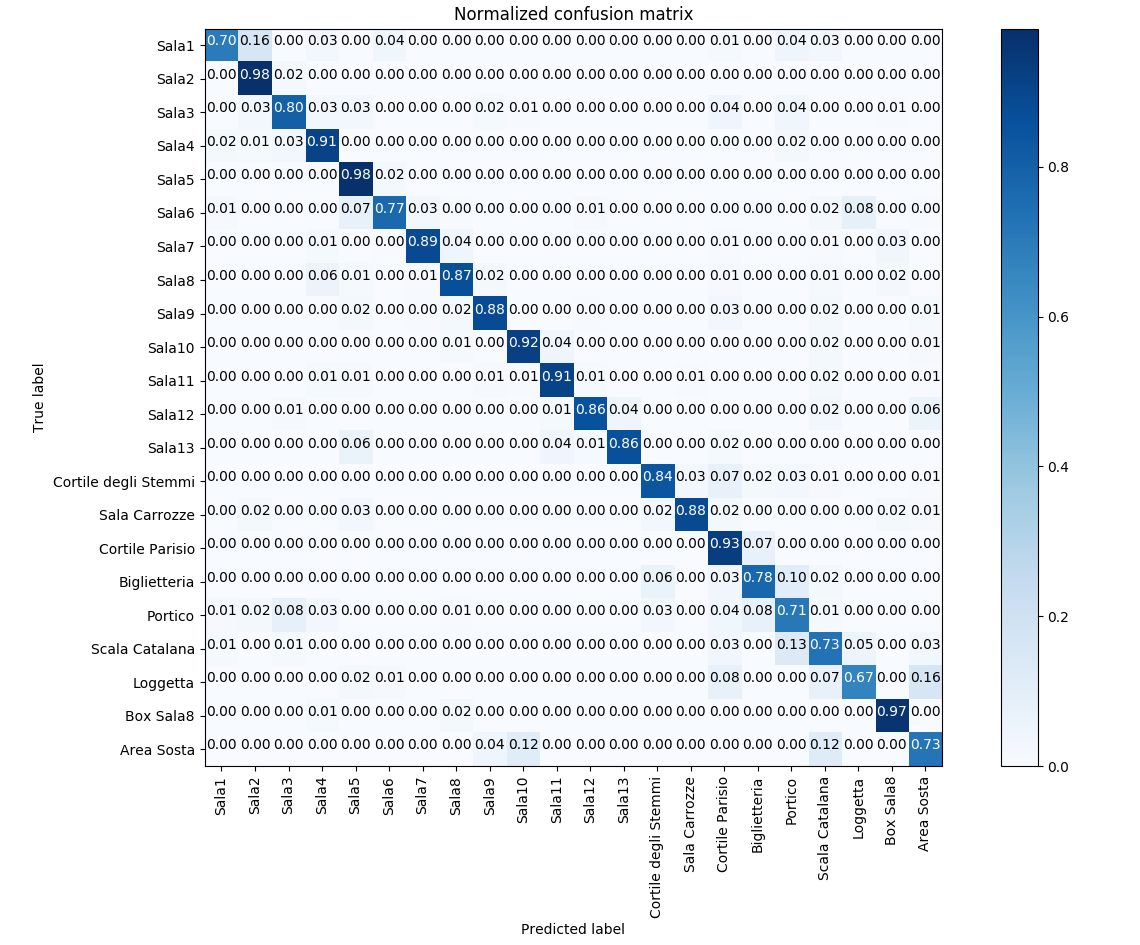}
	\caption{ Confusion matrix for localization in ``Palazzo Bellomo''.}
	\label{fig:bellomo_cf}
\end{figure*}

\begin{table*}[]
\centering
\caption{Detailed results of the $9$ test videos of ``Palazzo Bellomo'' using the $FF_1$ score. The backbone used is DenseNet.
			The ``/'' sign indicates that no samples from that class was present in the test video.}
\label{tab:F1_Dense}
\scriptsize
\begin{tabular}{lcccccccccc}
\multicolumn{11}{c}{\textbf{FF\_1 score}} \\ \hline
\multicolumn{1}{l|}{Class} & \multicolumn{1}{l}{Test1} & \multicolumn{1}{l}{Test2} & \multicolumn{1}{l}{Test4} & \multicolumn{1}{l}{Test5} & \multicolumn{1}{l}{Test6} & \multicolumn{1}{l}{Test7} & \multicolumn{1}{l}{Test8} & \multicolumn{1}{l}{Test9} & \multicolumn{1}{l|}{Test10} & \multicolumn{1}{l}{AVG} \\ \hline
\multicolumn{1}{l|}{1\_Sala1} & 0,00 & 0,96 & 0,96 & 0,95 & 0,99 & 0,94 & 0,97 & 0,97 & \multicolumn{1}{c|}{0,91} & 0,85 \\
\multicolumn{1}{l|}{2\_Sala2} & 0,82 & 0,91 & 0,96 & 0,95 & 0,99 & 0,99 & 0,99 & 0,97 & \multicolumn{1}{c|}{0,95} & 0,95 \\
\multicolumn{1}{l|}{3\_Sala3} & 0,65 & 0,84 & 0,84 & 0,83 & 0,94 & 0,88 & 0,61 & 0,85 & \multicolumn{1}{c|}{0,83} & 0,81 \\
\multicolumn{1}{l|}{4\_Sala4} & 0,84 & 0,96 & 0,98 & 0,93 & / & 0,96 & 0,78 & 0,97 & \multicolumn{1}{c|}{0,93} & 0,92 \\
\multicolumn{1}{l|}{5\_Sala5} & 0,98 & 0,88 & 0,96 & 0,91 & 0,65 & 0,91 & 0,97 & 0,97 & \multicolumn{1}{c|}{0,99} & 0,91 \\
\multicolumn{1}{l|}{6\_Sala6} & 0,84 & 0,80 & 0,43 & 0,00 & 0,86 & 0,00 & 0,18 & 0,83 & \multicolumn{1}{c|}{0,73} & 0,52 \\
\multicolumn{1}{l|}{7\_Sala7} & 0,67 & 0,93 & 0,22 & 0,00 & 0,38 & 0,88 & 0,88 & 0,40 & \multicolumn{1}{c|}{0,96} & 0,59 \\
\multicolumn{1}{l|}{8\_Sala8} & 0,64 & 0,75 & 0,60 & 0,52 & 0,56 & 0,56 & 0,64 & 0,60 & \multicolumn{1}{c|}{0,77} & 0,63 \\
\multicolumn{1}{l|}{9\_Sala9} & 0,90 & 0,89 & 0,73 & 0,88 & 0,41 & 0,92 & 0,98 & 0,99 & \multicolumn{1}{c|}{0,88} & 0,84 \\
\multicolumn{1}{l|}{10\_Sala10} & 0,96 & 0,98 & 0,92 & 0,72 & 0,00 & 0,98 & 0,85 & 0,97 & \multicolumn{1}{c|}{0,78} & 0,80 \\
\multicolumn{1}{l|}{11\_Sala11} & 0,93 & 0,96 & 0,96 & 0,97 & 0,00 & 0,97 & 0,97 & 0,98 & \multicolumn{1}{c|}{0,95} & 0,85 \\
\multicolumn{1}{l|}{12\_Sala12} & 0,82 & 0,88 & 0,87 & 0,90 & 0,00 & 0,87 & 0,85 & 0,92 & \multicolumn{1}{c|}{0,87} & 0,78 \\
\multicolumn{1}{l|}{13\_Sala13} & 0,96 & 0,95 & 0,95 & 0,76 & 0,00 & 0,93 & 0,95 & 0,94 & \multicolumn{1}{c|}{0,96} & 0,82 \\
\multicolumn{1}{l|}{14\_CortiledegliStemmi} & 0,78 & 0,68 & 0,74 & 0,88 & 0,17 & 0,90 & 0,79 & 0,92 & \multicolumn{1}{c|}{0,00} & 0,65 \\
\multicolumn{1}{l|}{15\_SalaCarrozze} & 0,84 & 0,89 & 0,95 & 0,93 & 0,91 & 0,84 & 0,97 & 0,95 & \multicolumn{1}{c|}{/} & 0,91 \\
\multicolumn{1}{l|}{16\_CortileParisio} & 0,33 & 0,51 & 0,52 & 0,45 & 0,35 & 0,60 & 0,81 & 0,91 & \multicolumn{1}{c|}{0,80} & 0,59 \\
\multicolumn{1}{l|}{17\_Biglietteria} & 0,25 & 0,56 & 0,00 & 0,26 & 0,00 & 0,00 & 0,00 & 0,36 & \multicolumn{1}{c|}{0,00} & 0,16 \\
\multicolumn{1}{l|}{18\_Portico} & 0,68 & 0,67 & 0,78 & 0,54 & 0,69 & 0,69 & 0,62 & 0,78 & \multicolumn{1}{c|}{0,62} & 0,67 \\
\multicolumn{1}{l|}{19\_ScalaCatalana} & 0,00 & 0,00 & 0,54 & 0,55 & 0,78 & 0,67 & 0,58 & 0,85 & \multicolumn{1}{c|}{0,49} & 0,50 \\
\multicolumn{1}{l|}{20\_Loggetta} & 0,00 & 0,50 & 0,80 & 0,44 & 0,72 & 0,53 & 0,55 & 0,83 & \multicolumn{1}{c|}{0,68} & 0,56 \\
\multicolumn{1}{l|}{21\_BoxSala8} & 0,96 & 0,97 & 0,97 & / & 0,00 & 0,97 & 0,99 & 0,81 & \multicolumn{1}{c|}{0,94} & 0,83 \\
\multicolumn{1}{l|}{22\_AreaSosta} & 0,74 & 0,85 & 0,45 & 0,83 & 0,00 & 0,56 & 0,72 & 0,85 & \multicolumn{1}{c|}{0,00} & 0,56 \\ \hline
\multicolumn{1}{c|}{mFF1} & 0,66 & 0,79 & 0,73 & 0,68 & 0,45 & 0,75 & 0,76 & 0,85 & \multicolumn{1}{c|}{0,72} & 0,71
\end{tabular}
\end{table*}

\begin{table*}[]
\centering
\caption{Detailed results of the $9$ test videos of ``Palazzo Bellomo'' using the $ASF_1$ score. The backbone used is DenseNet.
			The ``/'' sign indicates that no samples from that class was present in the test video.}
\label{tab:ASF1_Dense}
\scriptsize
\begin{tabular}{lcccccccccc}
\multicolumn{11}{c}{\textbf{ASF\_1 score}} \\ \hline
\multicolumn{1}{l|}{Class} & \multicolumn{1}{l}{Test1} & \multicolumn{1}{l}{Test2} & \multicolumn{1}{l}{Test4} & \multicolumn{1}{l}{Test5} & \multicolumn{1}{l}{Test6} & \multicolumn{1}{l}{Test7} & \multicolumn{1}{l}{Test8} & \multicolumn{1}{l}{Test9} & \multicolumn{1}{l|}{Test10} & \multicolumn{1}{l}{AVG} \\ \hline
\multicolumn{1}{l|}{1\_Sala1} & 0,00 & 0,93 & 0,93 & 0,91 & 0,97 & 0,89 & 0,94 & 0,94 & \multicolumn{1}{c|}{0,84} & 0,82 \\
\multicolumn{1}{l|}{2\_Sala2} & 0,64 & 0,83 & 0,91 & 0,90 & 0,97 & 0,97 & 0,97 & 0,94 & \multicolumn{1}{c|}{0,90} & 0,89 \\
\multicolumn{1}{l|}{3\_Sala3} & 0,36 & 0,53 & 0,51 & 0,53 & 0,61 & 0,63 & 0,29 & 0,59 & \multicolumn{1}{c|}{0,72} & 0,53 \\
\multicolumn{1}{l|}{4\_Sala4} & 0,39 & 0,92 & 0,96 & 0,86 & / & 0,92 & 0,31 & 0,94 & \multicolumn{1}{c|}{0,87} & 0,77 \\
\multicolumn{1}{l|}{5\_Sala5} & 0,94 & 0,64 & 0,66 & 0,62 & 0,34 & 0,63 & 0,54 & 0,95 & \multicolumn{1}{c|}{0,97} & 0,70 \\
\multicolumn{1}{l|}{6\_Sala6} & 0,46 & 0,43 & 0,18 & 0,00 & 0,42 & 0,00 & 0,09 & 0,59 & \multicolumn{1}{c|}{0,51} & 0,30 \\
\multicolumn{1}{l|}{7\_Sala7} & 0,50 & 0,86 & 0,13 & 0,00 & 0,10 & 0,78 & 0,28 & 0,25 & \multicolumn{1}{c|}{0,92} & 0,42 \\
\multicolumn{1}{l|}{8\_Sala8} & 0,36 & 0,44 & 0,42 & 0,35 & 0,35 & 0,41 & 0,49 & 0,53 & \multicolumn{1}{c|}{0,57} & 0,44 \\
\multicolumn{1}{l|}{9\_Sala9} & 0,81 & 0,80 & 0,51 & 0,78 & 0,22 & 0,84 & 0,95 & 0,97 & \multicolumn{1}{c|}{0,78} & 0,74 \\
\multicolumn{1}{l|}{10\_Sala10} & 0,92 & 0,96 & 0,85 & 0,19 & 0,00 & 0,95 & 0,27 & 0,94 & \multicolumn{1}{c|}{0,64} & 0,64 \\
\multicolumn{1}{l|}{11\_Sala11} & 0,53 & 0,92 & 0,91 & 0,93 & 0,00 & 0,93 & 0,65 & 0,95 & \multicolumn{1}{c|}{0,91} & 0,75 \\
\multicolumn{1}{l|}{12\_Sala12} & 0,38 & 0,79 & 0,76 & 0,81 & 0,00 & 0,77 & 0,30 & 0,85 & \multicolumn{1}{c|}{0,76} & 0,60 \\
\multicolumn{1}{l|}{13\_Sala13} & 0,93 & 0,90 & 0,90 & 0,62 & 0,00 & 0,87 & 0,91 & 0,89 & \multicolumn{1}{c|}{0,92} & 0,77 \\
\multicolumn{1}{l|}{14\_CortiledegliStemmi} & 0,57 & 0,53 & 0,54 & 0,77 & 0,08 & 0,81 & 0,49 & 0,67 & \multicolumn{1}{c|}{0,00} & 0,50 \\
\multicolumn{1}{l|}{15\_SalaCarrozze} & 0,72 & 0,80 & 0,91 & 0,87 & 0,83 & 0,72 & 0,94 & 0,90 & \multicolumn{1}{c|}{/} & 0,84 \\
\multicolumn{1}{l|}{16\_CortileParisio} & 0,30 & 0,63 & 0,43 & 0,65 & 0,38 & 0,47 & 0,59 & 0,83 & \multicolumn{1}{c|}{0,66} & 0,55 \\
\multicolumn{1}{l|}{17\_Biglietteria} & 0,26 & 0,43 & 0,00 & 0,11 & 0,00 & 0,00 & 0,00 & 0,39 & \multicolumn{1}{c|}{0,00} & 0,13 \\
\multicolumn{1}{l|}{18\_Portico} & 0,51 & 0,47 & 0,60 & 0,45 & 0,48 & 0,52 & 0,42 & 0,64 & \multicolumn{1}{c|}{0,47} & 0,51 \\
\multicolumn{1}{l|}{19\_ScalaCatalana} & 0,00 & 0,00 & 0,43 & 0,51 & 0,61 & 0,55 & 0,42 & 0,75 & \multicolumn{1}{c|}{0,39} & 0,41 \\
\multicolumn{1}{l|}{20\_Loggetta} & 0,00 & 0,31 & 0,54 & 0,28 & 0,46 & 0,38 & 0,48 & 0,72 & \multicolumn{1}{c|}{0,44} & 0,40 \\
\multicolumn{1}{l|}{21\_BoxSala8} & 0,92 & 0,95 & 0,95 & / & 0,00 & 0,93 & 0,98 & 0,67 & \multicolumn{1}{c|}{0,88} & 0,79 \\
\multicolumn{1}{l|}{22\_AreaSosta} & 0,59 & 0,73 & 0,29 & 0,71 & 0,00 & 0,75 & 0,57 & 0,74 & \multicolumn{1}{c|}{0,00} & 0,49 \\ \hline
\multicolumn{1}{c|}{mFF1} & 0,50 & 0,67 & 0,61 & 0,56 & 0,32 & 0,67 & 0,54 & 0,76 & \multicolumn{1}{c|}{0,63} & 0,58
\end{tabular}
\end{table*}

\paragraph{2) Monastero dei Benedettini}
Similarly to ``Palazzo Bellomo", we split the Training Set into two subsets to train and validate the VGG-19 (no ``negative" frames are used for training on this cultural site). 
Table~\ref{tab:train-val} reports the number of frames belonging to the two subsets for each of the 4 considered environments.
We report the results obtained by the baseline method over the $60$ test videos in Table~\ref{tab:monresultsF1} and Table~\ref{tab:monresultsF1_1} considering the $FF_1$ score metric, and in Table~\ref{tab:monresultsASF1} and Table~\ref{tab:monresultsASF1_1} considering the $ASF_1$ score.

\begin{table}[t]
	\centering
	\caption{Number of frames belonging to the two subsets (Training/Validation) to train the CNN for ``Monastero dei Benedettini''.}
	\label{tab:train-val}
	\begin{tabular}{l|c|c}
		\hline
		& \multicolumn{1}{l|}{\textbf{Training}} & \multicolumn{1}{l}{\textbf{Validation}} \\ \hline
		
		1 Antirefettorio       & 7000                                   & 3000                                    \\
		2 Aula Santo Mazzarino & 7000                                   & 3000                                    \\
		3 Cucina               & 7000                                   & 3000                                    \\
		4 Ventre               & 7000                                   & 3000                                    \\ \hline
		\textbf{Total}  & 28000                                  & 12000                                  
	\end{tabular}
\end{table}

\begin{table}[t]
\centering
\caption{Detailed results on the $60$ test videos of ``Monastero dei Benedettini'', considering the evaluation measure $FF_1$ score. The ``/'' sign indicates that no samples from that class were present in the test video. The four classes are: 1) Antirefettorio, 2) Aula S. Mazzarino, 3) Cucina, 3) Ventre, whereas Neg. represents the negatives.}
\label{tab:monresultsF1}
\scriptsize
\begin{tabular}{l|ccccc|c}
\textbf{ID\_Visit} & \textbf{1} & \textbf{2} & \textbf{3} & \textbf{4} & \textbf{Neg.} & \textbf{AVG} \\ \hline
4805               & 0,91                      & 0,33                         & 0,75              & 0,99              & 0,43               & 0,682        \\
1804               & 0,69                      & 0,43                         & 0,8               & 0,99              & 0,43               & 0,668        \\
4377               & 0,77                      & 0,47                         & 0,84              & 0,99              & 0,5                & 0,714        \\
1669               & 0,8                       & 0,51                         & 0,92              & 0,99              & 0,67               & 0,778        \\
1791               & 0,59                      & 0,22                         & 0,78              & 0,98              & 0,35               & 0,584        \\
3948               & 0,8                       & /                            & 0,66              & 0,99              & 0,61               & 0,765        \\
3152               & 0,71                      & 0,25                         & 0,86              & 0,98              & 0,75               & 0,71         \\
4361               & 0,89                      & 0,32                         & 0,81              & 0,099             & 0,62               & 0,5478       \\
3976               & 0,97                      & 0,62                         & 0,92              & 0,98              & 0,68               & 0,834        \\
3527               & 0,85                      & 0                            & 0,82              & 0,99              & 0,63               & 0,658        \\
4105               & 0,81                      & 0                            & 0,77              & 0,99              & 0,66               & 0,646        \\
1399               & 0,65                      & 0                            & 0,76              & 0,99              & 0,62               & 0,604        \\
3836               & 0,65                      & 0                            & 0,76              & 0,99              & 0,62               & 0,604        \\
4006               & 0,81                      & 0,82                         & 0,92              & 0,99              & 0,75               & 0,858        \\
4415               & 0,87                      & 0,49                         & 0,77              & 0,98              & 0,73               & 0,768        \\
3008               & 0,82                      & 0,2                          & 0,63              & 0,99              & 0,23               & 0,574        \\
4660               & 0,82                      & 0,2                          & 0,63              & 0,99              & 0,23               & 0,574        \\
2826               & 0,79                      & 0,57                         & 0,81              & 1                 & 0,41               & 0,716        \\
1099               & 0,77                      & 0,27                         & 0,64              & 0,98              & 0,5                & 0,632        \\
4391               & 0,8                       & 0,03                         & 0,79              & 0,98              & 0,55               & 0,63         \\
3929               & 0,94                      & 0                            & 0,84              & 0,99              & 0,67               & 0,688        \\
3362               & 0,46                      & 0,25                         & 0,76              & 0,99              & 0,46               & 0,584        \\
1379               & 0,84                      & 0                            & 0,75              & 0,98              & 0,33               & 0,58         \\
2600               & 0,74                      & 0                            & 0,78              & 0,99              & 0,59               & 0,62         \\
1430               & 0,57                      & 0,22                         & 0,72              & 0,96              & 0,47               & 0,588        \\
2956               & 0,53                      & 0,33                         & 0,73              & 0,96              & 0,8                & 0,67         \\
4742               & 0,14                      & 0,43                         & 0,92              & 0,99              & 0,67               & 0,63         \\
3651               & 0,94                      & 0,66                         & 0,82              & 0,99              & 0,49               & 0,78         \\
1064               & 0,77                      & 0,12                         & 0,88              & 0,99              & 0,31               & 0,614        \\
3818               & 0,93                      & 0,14                         & 0,71              & 0,99              & 0,51               & 0,656        \\ \hline
\end{tabular}
\end{table}

\begin{table}[t]
\centering
\caption{Continued from \tablename~\ref{tab:monresultsF1} }
\label{tab:monresultsF1_1}
\scriptsize
\begin{tabular}{lcccccc}
\multicolumn{1}{l|}{\textbf{ID\_Visit}} & \textbf{1} & \textbf{2} & \textbf{3} & \textbf{4} & \multicolumn{1}{c|}{\textbf{Neg.}} & \textbf{AVG}  \\ \hline
\multicolumn{1}{l|}{2043}               & 0,72                      & 0,47                         & 0,79              & 0,98              & \multicolumn{1}{c|}{0,66}               & 0,724         \\
\multicolumn{1}{l|}{3996}               & 0,47                      & 0,4                          & 0,72              & 0,98              & \multicolumn{1}{c|}{0,76}               & 0,666         \\
\multicolumn{1}{l|}{3455}               & 0,85                      & 0,33                         & 0,88              & 0,99              & \multicolumn{1}{c|}{0,61}               & 0,732         \\
\multicolumn{1}{l|}{4785}               & 0,03                      & 0                            & 0,73              & 0,96              & \multicolumn{1}{c|}{0,65}               & 0,474         \\
\multicolumn{1}{l|}{2047}               & 0,95                      & 0,55                         & 0,86              & 1                 & \multicolumn{1}{c|}{0,6}                & 0,792         \\
\multicolumn{1}{l|}{1912}               & 0,79                      & 0,44                         & 0,67              & 0,96              & \multicolumn{1}{c|}{0,62}               & 0,696         \\
\multicolumn{1}{l|}{3232}               & 0,89                      & 0,33                         & 0,82              & 0,99              & \multicolumn{1}{c|}{0,73}               & 0,752         \\
\multicolumn{1}{l|}{4442}               & 0,9                       & 0,37                         & 0,82              & 0,99              & \multicolumn{1}{c|}{0,54}               & 0,724         \\
\multicolumn{1}{l|}{3646}               & 0,67                      & 0,09                         & 0,8               & 1                 & \multicolumn{1}{c|}{0,34}               & 0,58          \\
\multicolumn{1}{l|}{4833}               & 0,78                      & 0,59                         & 0,86              & 0,95              & \multicolumn{1}{c|}{0,66}               & 0,768         \\
\multicolumn{1}{l|}{3478}               & 0,82                      & 0,51                         & 0,94              & 1                 & \multicolumn{1}{c|}{0,67}               & 0,788         \\
\multicolumn{1}{l|}{4396}               & 0,81                      & 0,53                         & 0,89              & 0,99              & \multicolumn{1}{c|}{0,38}               & 0,72          \\
\multicolumn{1}{l|}{2894}               & 0,67                      & 0                            & 0,87              & 0,97              & \multicolumn{1}{c|}{0,55}               & 0,612         \\
\multicolumn{1}{l|}{4414}               & 0,88                      & 0,51                         & 0,9               & 0,98              & \multicolumn{1}{c|}{0,58}               & 0,77          \\
\multicolumn{1}{l|}{4639}               & 0,73                      & 0,21                         & 0,83              & 0,99              & \multicolumn{1}{c|}{0,61}               & 0,674         \\
\multicolumn{1}{l|}{1004}               & 0,15                      & 0,44                         & 0,6               & 0,98              & \multicolumn{1}{c|}{0,48}               & 0,53          \\
\multicolumn{1}{l|}{1917}               & 0,44                      & 0,48                         & 0,51              & 1                 & \multicolumn{1}{c|}{0,42}               & 0,57          \\
\multicolumn{1}{l|}{1153}               & 0,78                      & 0,54                         & 0,81              & 0,98              & \multicolumn{1}{c|}{0,57}               & 0,736         \\
\multicolumn{1}{l|}{2244}               & 0,86                      & 0,3                          & 0,77              & 0,99              & \multicolumn{1}{c|}{0,32}               & 0,648         \\
\multicolumn{1}{l|}{2614}               & 0,97                      & 0,59                         & 0,84              & 0,99              & \multicolumn{1}{c|}{0,39}               & 0,756         \\
\multicolumn{1}{l|}{1624}               & 0,91                      & 0,71                         & 0,69              & 0,99              & \multicolumn{1}{c|}{0,48}               & 0,756         \\
\multicolumn{1}{l|}{3441}               & 0,82                      & 0,45                         & 0,79              & 0,99              & \multicolumn{1}{c|}{0,46}               & 0,702         \\
\multicolumn{1}{l|}{4793}               & 0,82                      & /                            & 0,74              & 0,99              & \multicolumn{1}{c|}{0,42}               & 0,7425        \\
\multicolumn{1}{l|}{4083}               & 0,84                      & /                            & 0,72              & 0,99              & \multicolumn{1}{c|}{0,73}               & 0,82          \\
\multicolumn{1}{l|}{4906}               & 0,77                      & 0,2                          & 0,74              & 0,99              & \multicolumn{1}{c|}{0,42}               & 0,624         \\
\multicolumn{1}{l|}{1160}               & 0,84                      & 0,66                         & 0,74              & 1                 & \multicolumn{1}{c|}{0,42}               & 0,732         \\
\multicolumn{1}{l|}{3416}               & 0,77                      & /                            & 0,82              & 0,99              & \multicolumn{1}{c|}{0,68}               & 0,815         \\
\multicolumn{1}{l|}{1051}               & 0,79                      & 0,43                         & 0,78              & 0,98              & \multicolumn{1}{c|}{0,4}                & 0,676         \\
\multicolumn{1}{l|}{2580}               & 0,73                      & 0,18                         & 0,89              & 0,99              & \multicolumn{1}{c|}{0,48}               & 0,654         \\
\multicolumn{1}{l|}{1109}               & 0,81                      & 0,28                         & 0,89              & 0,99              & \multicolumn{1}{c|}{0,43}               & 0,68          \\ \hline
\multicolumn{1}{c}{\textbf{AVG}}        & \textbf{0,75}             & \textbf{0,33}                & \textbf{0,79}     & \textbf{0,99}     & \textbf{0,54}                           & \textbf{0,68}
\end{tabular}
\end{table}

\begin{table}[t]
\centering
\caption{Detailed results on the $60$ test videos of ``Monastero dei Benedettini'', considering the evaluation measure $ASF_1$ score. The ``/'' sign indicates that no samples from that class were present in the test video. The four classes are: 1) Antirefettorio, 2) Aula S. Mazzarino, 3) Cucina, 3) Ventre, whereas Neg. represents the negatives.}
\label{tab:monresultsASF1}
\scriptsize
\begin{tabular}{l|ccccc|c}
\textbf{ID\_Visit} & \textbf{1} & \textbf{2} & \textbf{3} & \textbf{4} & \textbf{Neg.} & \textbf{AVG} \\ \hline
4805               & 0,71       & 0,06       & 0,2        & 0,31       & 0,22          & 0,3          \\
1804               & 0,55       & 0,06       & 0,26       & 0,15       & 0,14          & 0,232        \\
4377               & 0,55       & 0,05       & 0,18       & 0,98       & 0,25          & 0,402        \\
1669               & 0,4        & 0,22       & 0,27       & 0,93       & 0,49          & 0,462        \\
1791               & 0,46       & 0,14       & 0,21       & 0,25       & 0,19          & 0,25         \\
3948               & 0,67       & /          & 0,26       & 0,98       & 0,51          & 0,605        \\
3152               & 0,54       & 0,07       & 0,66       & 0,88       & 0,47          & 0,524        \\
4361               & 0,47       & 0,12       & 0,17       & 0,27       & 0,29          & 0,264        \\
3976               & 0,94       & 0,24       & 0,45       & 0,22       & 0,56          & 0,482        \\
3527               & 0,72       & 0          & 0,44       & 0,46       & 0,55          & 0,434        \\
4105               & 0,54       & 0          & 0,49       & 0,39       & 0,51          & 0,386        \\
1399               & 0,15       & 0          & 0,44       & 0,98       & 0,26          & 0,366        \\
3836               & 0,15       & 0          & 0,44       & 0,98       & 0,26          & 0,366        \\
4006               & 0,56       & 0,69       & 0,53       & 0,98       & 0,5           & 0,652        \\
4415               & 0,48       & 0,15       & 0,24       & 0,49       & 0,48          & 0,368        \\
3008               & 0,68       & 0,1        & 0,26       & 0,98       & 0,1           & 0,424        \\
4660               & 0,68       & 0,09       & 0,26       & 0,98       & 0,09          & 0,42         \\
2826               & 0,71       & 0,14       & 0,24       & 0,99       & 0,41          & 0,498        \\
1099               & 0,54       & 0,15       & 0,16       & 0,49       & 0,31          & 0,33         \\
4391               & 0,67       & 0,05       & 0,32       & 0,96       & 0,53          & 0,506        \\
3929               & 0,91       & 0          & 0,43       & 0,98       & 0,46          & 0,556        \\
3362               & 0,36       & 0,07       & 0,45       & 0,57       & 0,36          & 0,362        \\
1379               & 0,49       & 0          & 0,21       & 0,18       & 0,23          & 0,222        \\
2600               & 0,64       & 0          & 0,41       & 0,98       & 0,38          & 0,482        \\
1430               & 0,17       & 0,09       & 0,03       & 0,08       & 0,1           & 0,094        \\
2956               & 0,33       & 0,06       & 0,46       & 0,37       & 0,48          & 0,34         \\
4742               & 0,1        & 0,09       & 0,39       & 0,62       & 0,17          & 0,274        \\
3651               & 0,63       & 0,19       & 0,54       & 0,65       & 0,28          & 0,458        \\
1064               & 0,63       & 0,05       & 0,3        & 0,65       & 0,13          & 0,352        \\
3818               & 0,78       & 0,11       & 0,17       & 0,25       & 0,35          & 0,332        \\ \hline
\end{tabular}
\end{table}

\begin{table}[t]
\centering
\caption{Continued from \tablename~\ref{tab:monresultsASF1}}
\label{tab:monresultsASF1_1}
\scriptsize
\begin{tabular}{lcccccc}
\multicolumn{1}{l|}{\textbf{ID\_Visit}} & \textbf{1}    & \textbf{2}    & \textbf{3}    & \textbf{4}    & \multicolumn{1}{c|}{\textbf{Neg.}} & \textbf{AVG}  \\ \hline
\multicolumn{1}{l|}{2043}               & 0,4           & 0,22          & 0,2           & 0,27          & \multicolumn{1}{c|}{0,25}          & 0,268         \\
\multicolumn{1}{l|}{3996}               & 0,45          & 0,12          & 0,39          & 0,4           & \multicolumn{1}{c|}{0,49}          & 0,37          \\
\multicolumn{1}{l|}{3455}               & 0,69          & 0,16          & 0,39          & 0,96          & \multicolumn{1}{c|}{0,5}           & 0,54          \\
\multicolumn{1}{l|}{4785}               & 0,05          & 0,11          & 0,53          & 0,13          & \multicolumn{1}{c|}{0.33}             & 0,205         \\
\multicolumn{1}{l|}{2047}               & 0,9           & 0,38          & 0,22          & 0,99          & \multicolumn{1}{c|}{0,36}          & 0,57          \\
\multicolumn{1}{l|}{1912}               & 0,54          & 0,1           & 0,16          & 0,31          & \multicolumn{1}{c|}{0,39}          & 0,3           \\
\multicolumn{1}{l|}{3232}               & 0,73          & 0,04          & 0,49          & 0,98          & \multicolumn{1}{c|}{0,53}          & 0,554         \\
\multicolumn{1}{l|}{4442}               & 0,59          & 0,285         & 0,22          & 0,27          & \multicolumn{1}{c|}{0,25}          & 0,323         \\
\multicolumn{1}{l|}{3646}               & 0,4           & 0,13          & 0,2           & 0,66          & \multicolumn{1}{c|}{0,34}          & 0,346         \\
\multicolumn{1}{l|}{4833}               & 0,54          & 0,15          & 0,53          & 0,19          & \multicolumn{1}{c|}{0,36}          & 0,354         \\
\multicolumn{1}{l|}{3478}               & 0,71          & 0,29          & 0,54          & 0,99          & \multicolumn{1}{c|}{0,46}          & 0,598         \\
\multicolumn{1}{l|}{4396}               & 0,35          & 0,02          & 0,05          & 0,18          & \multicolumn{1}{c|}{0,08}          & 0,136         \\
\multicolumn{1}{l|}{2894}               & 0,43          & 0             & 0,46          & 0,35          & \multicolumn{1}{c|}{0,18}          & 0,284         \\
\multicolumn{1}{l|}{4414}               & 0,75          & 0,11          & 0,31          & 0,24          & \multicolumn{1}{c|}{0,3}           & 0,342         \\
\multicolumn{1}{l|}{4639}               & 0,61          & 0,12          & 0,44          & 0,39          & \multicolumn{1}{c|}{0,43}          & 0,398         \\
\multicolumn{1}{l|}{1004}               & 0,22          & 0,11          & 0,27          & 0,21          & \multicolumn{1}{c|}{0,38}          & 0,238         \\
\multicolumn{1}{l|}{1917}               & 0,41          & 0,11          & 0,15          & 0,31          & \multicolumn{1}{c|}{0,26}          & 0,248         \\
\multicolumn{1}{l|}{1153}               & 0,54          & 0,11          & 0,33          & 0,59          & \multicolumn{1}{c|}{0,26}          & 0,366         \\
\multicolumn{1}{l|}{2244}               & 0,58          & 0,16          & 0,45          & 0,97          & \multicolumn{1}{c|}{0,26}          & 0,484         \\
\multicolumn{1}{l|}{2614}               & 0,65          & 0,05          & 0,39          & 0,98          & \multicolumn{1}{c|}{0,3}           & 0,474         \\
\multicolumn{1}{l|}{1624}               & 0,71          & 0,29          & 0,35          & 0,39          & \multicolumn{1}{c|}{0,23}          & 0,394         \\
\multicolumn{1}{l|}{3441}               & 0,59          & 0,13          & 0,26          & 0,88          & \multicolumn{1}{c|}{0,25}          & 0,422         \\
\multicolumn{1}{l|}{4793}               & 0,68          & /             & 0,52          & 0,98          & \multicolumn{1}{c|}{0,38}          & 0,64          \\
\multicolumn{1}{l|}{4083}               & 0,65          & /             & 0,31          & 0,35          & \multicolumn{1}{c|}{0,59}          & 0,475         \\
\multicolumn{1}{l|}{4906}               & 0,57          & 0,06          & 0,33          & 0,59          & \multicolumn{1}{c|}{0,33}          & 0,376         \\
\multicolumn{1}{l|}{1160}               & 0,7           & 0,15          & 0,34          & 0,99          & \multicolumn{1}{c|}{0,29}          & 0,494         \\
\multicolumn{1}{l|}{3416}               & 0,45          & /             & 0,24          & 0,97          & \multicolumn{1}{c|}{0,19}          & 0,4625        \\
\multicolumn{1}{l|}{1051}               & 0,55          & 0,28          & 0,33          & 0,4           & \multicolumn{1}{c|}{0,27}          & 0,366         \\
\multicolumn{1}{l|}{2580}               & 0,37          & 0,1           & 0,68          & 0,21          & \multicolumn{1}{c|}{0,37}          & 0,346         \\
\multicolumn{1}{l|}{1109}               & 0,66          & 0,16          & 0,63          & 0,98          & \multicolumn{1}{c|}{0,36}          & 0,558         \\ \hline
\multicolumn{1}{c}{\textbf{AVG}}        & \textbf{0,54} & \textbf{0,12} & \textbf{0,34} & \textbf{0,60} & \textbf{0,33}                      & \textbf{0,40}
\end{tabular}
\end{table}

\paragraph{\textbf{DenseNet Backbone}}
We performed experiments using another backbone in the same pipeline to solve the first task. We used DenseNet\cite{huang2017densely}, a densely convolutional connect network which connects each layer to every other layer in a feed-forward fashion. Table~\ref{tab:F1_Dense} and Table~\ref{tab:ASF1_Dense} report the results obtained with DenseNet at the end of the Sequential Modeling step. We have evaluated the model using F1 score and Asf1 score. As shown the tables we did not obtain an improvement respect the results obtained with the backbone VGG. In particular for the "Palazzo Bellomo" we obtained a $FF_1$ score of $0.71$ and a $ASF_1$ score of 0.58 which are lower respect the scores obtained with VGG ($FF_1 = 0.82$, $ASF_1 = 0.59$). From Table~\ref{tab:monresultsF1} to Table~\ref{tab:monresultsASF1_1} we report the results obtained in the "Monastero dei Benedettini" using the backbone DenseNet. As shown, neither considering this cultural site we obtained an improvement of $FF_1$ and $ASF_1$ scores.

\begin{table}[]
\centering
\caption{Detailed results on the $60$ test videos of ``Monastero dei Benedettini'', considering the evaluation measure $FF_1$ score. We used the backbone DenseNet. The ``/'' sign indicates that no samples from that class were present in the test video. The four classes are: 1) Antirefettorio, 2) Aula S. Mazzarino, 3) Cucina, 3) Ventre, whereas Neg. represents the negatives.}
\label{tab:monDenseF1_1}
\scriptsize
\begin{tabular}{l|ccccc|c}
\textbf{ID\_Visit} & \textbf{1} & \textbf{2} & 3 & 4 & Neg. & AVG \\ \hline
4805 & 0,79 & 0,82 & 0,74 & 0,95 & 0,45 & 0,75 \\
1804 & 0,32 & 0,36 & 0,79 & 0,98 & 0,55 & 0,6 \\
4377 & 0,46 & 0,53 & 0,8 & 0,96 & 0,32 & 0,614 \\
1669 & 0,75 & 0,72 & 0,9 & 0,99 & 0,45 & 0,762 \\
1791 & 0,49 & 0,5 & 0,84 & 0,95 & 0,43 & 0,642 \\
3948 & 0 & / & 0,59 & 0,98 & 0,49 & 0,515 \\
3152 & 0,39 & 0,72 & 0,87 & 0,97 & 0,76 & 0,742 \\
4361 & 0,55 & 0,78 & 0,82 & 0,97 & 0,28 & 0,68 \\
3976 & 0,87 & 0,43 & 0,81 & 0,93 & 0,66 & 0,74 \\
3527 & 0,86 & 0,79 & 0,85 & 0,99 & 0,56 & 0,81 \\
4105 & 0,62 & 0 & 0,77 & 0,97 & 0,09 & 0,49 \\
1399 & 0,46 & 0,23 & 0,79 & 0,98 & 0,43 & 0,578 \\
3836 & 0,51 & 0 & 0,72 & 0,99 & 0,55 & 0,554 \\
4006 & 0,57 & 0,78 & 0,78 & 0,99 & 0,37 & 0,698 \\
4415 & 0,86 & 0,61 & 0,82 & 0,97 & 0,35 & 0,722 \\
3008 & 0,69 & 0 & 0,7 & 0,99 & 0,51 & 0,578 \\
4660 & 0,62 & 0,81 & 0,83 & 0,98 & 0,57 & 0,762 \\
2826 & 0,31 & 0,52 & 0,76 & 0,97 & 0,61 & 0,634 \\
1099 & 0,62 & 0,42 & 0,85 & 0,98 & 0,49 & 0,672 \\
4391 & 0,74 & 0,72 & 0,72 & 0,98 & 0,33 & 0,698 \\
3929 & 0,26 & 0 & 0,8 & 0,99 & 0,43 & 0,496 \\
3362 & 0,34 & 0,68 & 0,68 & 0,95 & 0,21 & 0,572 \\
1379 & 0,41 & 0 & 0,81 & 0,96 & 0,45 & 0,526 \\
2600 & 0,26 & 0 & 0,63 & 0,96 & 0,3 & 0,43 \\
1430 & 0,94 & 0 & 0,69 & 0,87 & 0,58 & 0,616 \\
2956 & 0,32 & 0,45 & 0,33 & 0,91 & 0,65 & 0,532 \\
4742 & 0,13 & 0,58 & 0,59 & 0,84 & 0,5 & 0,528 \\
3651 & 0,77 & 0,41 & 0,88 & 0,98 & 0,41 & 0,69 \\
1064 & 0,77 & 0,23 & 0,83 & 0,99 & 0,22 & 0,608 \\
3818 & 0,68 & 0,64 & 0,73 & 0,99 & 0,47 & 0,702 \\ \hline
\end{tabular}
\end{table}

\begin{table}[]
\centering
\caption{Continued from \tablename~\ref{tab:monDenseF1_1}}
\label{tab:monDenseF1_2}
\scriptsize
\begin{tabular}{lcccccc}
\multicolumn{1}{l|}{\textbf{ID\_Visit}} & \textbf{1} & \textbf{2} & \textbf{3} & \textbf{4} & \multicolumn{1}{c|}{\textbf{Neg.}} & \textbf{AVG} \\ \hline
\multicolumn{1}{l|}{2043} & 0,52 & 0,44 & 0,62 & 0,91 & \multicolumn{1}{c|}{0,33} & 0,564 \\
\multicolumn{1}{l|}{3996} & 0,17 & 0,68 & 0,57 & 0,95 & \multicolumn{1}{c|}{0,72} & 0,618 \\
\multicolumn{1}{l|}{3455} & 0,78 & 0,65 & 0,8 & 0,99 & \multicolumn{1}{c|}{0,55} & 0,754 \\
\multicolumn{1}{l|}{4785} & 0,02 & 0 & 0,64 & 0,95 & \multicolumn{1}{c|}{0,33} & 0,388 \\
\multicolumn{1}{l|}{2047} & 0,95 & 0,84 & 0,84 & 0,98 & \multicolumn{1}{c|}{0,38} & 0,798 \\
\multicolumn{1}{l|}{1912} & 0,66 & 0,51 & 0,69 & 0,94 & \multicolumn{1}{c|}{0,46} & 0,652 \\
\multicolumn{1}{l|}{3232} & 0,66 & 0,53 & 0,85 & 0,99 & \multicolumn{1}{c|}{0,49} & 0,704 \\
\multicolumn{1}{l|}{4442} & 0,8 & 0,77 & 0,82 & 0,98 & \multicolumn{1}{c|}{0,23} & 0,72 \\
\multicolumn{1}{l|}{3646} & 0,63 & 0,29 & 0,66 & 0,97 & \multicolumn{1}{c|}{0,4} & 0,59 \\
\multicolumn{1}{l|}{4833} & 0,67 & 0,82 & 0,67 & 0,88 & \multicolumn{1}{c|}{0,24} & 0,656 \\
\multicolumn{1}{l|}{3478} & 0,71 & 0,75 & 0,84 & 0,98 & \multicolumn{1}{c|}{0,37} & 0,73 \\
\multicolumn{1}{l|}{4396} & 0,74 & 0,69 & 0,89 & 0,96 & \multicolumn{1}{c|}{0,26} & 0,708 \\
\multicolumn{1}{l|}{2894} & 0,83 & 0,8 & 0,66 & 0,92 & \multicolumn{1}{c|}{0,48} & 0,738 \\
\multicolumn{1}{l|}{4414} & 0,75 & 0,7 & 0,91 & 0,95 & \multicolumn{1}{c|}{0,22} & 0,706 \\
\multicolumn{1}{l|}{4639} & 0,57 & 0,52 & 0,85 & 0,99 & \multicolumn{1}{c|}{0,39} & 0,664 \\
\multicolumn{1}{l|}{1004} & 0,08 & 0,72 & 0,49 & 0,97 & \multicolumn{1}{c|}{0,26} & 0,504 \\
\multicolumn{1}{l|}{1917} & 0,41 & 0,71 & 0,36 & 0,87 & \multicolumn{1}{c|}{0,29} & 0,528 \\
\multicolumn{1}{l|}{1153} & 0,62 & 0,7 & 0,71 & 0,91 & \multicolumn{1}{c|}{0,32} & 0,652 \\
\multicolumn{1}{l|}{2244} & 0,94 & 0,56 & 0,74 & 0,99 & \multicolumn{1}{c|}{0,43} & 0,732 \\
\multicolumn{1}{l|}{2614} & 0,88 & 0,56 & 0,83 & 0,99 & \multicolumn{1}{c|}{0,4} & 0,732 \\
\multicolumn{1}{l|}{1624} & 0,33 & 0,8 & 0,62 & 0,99 & \multicolumn{1}{c|}{0,33} & 0,614 \\
\multicolumn{1}{l|}{3441} & 0,61 & 0,25 & 0,84 & 0,99 & \multicolumn{1}{c|}{0,41} & 0,62 \\
\multicolumn{1}{l|}{4793} & 0,52 & / & 0,68 & 0,99 & \multicolumn{1}{c|}{0,33} & 0,63 \\
\multicolumn{1}{l|}{4083} & 0,93 & / & 0,73 & 0,99 & \multicolumn{1}{c|}{0,71} & 0,84 \\
\multicolumn{1}{l|}{4906} & 0,46 & 0,36 & 0,59 & 0,94 & \multicolumn{1}{c|}{0,31} & 0,532 \\
\multicolumn{1}{l|}{1160} & 0,88 & 0,84 & 0,72 & 0,98 & \multicolumn{1}{c|}{0,46} & 0,776 \\
\multicolumn{1}{l|}{3416} & 0,56 & / & 0,5 & 0,82 & \multicolumn{1}{c|}{0,2} & 0,52 \\
\multicolumn{1}{l|}{1051} & 0,78 & 0,76 & 0,64 & 0,95 & \multicolumn{1}{c|}{0,57} & 0,74 \\
\multicolumn{1}{l|}{2580} & 0,71 & 0,45 & 0,63 & 0,96 & \multicolumn{1}{c|}{0,44} & 0,638 \\
\multicolumn{1}{l|}{1109} & 0,91 & 0,28 & 0,85 & 0,97 & \multicolumn{1}{c|}{0,36} & 0,674 \\ \hline
\multicolumn{1}{c}{mFF1} & \textbf{0,59} & \textbf{0,51} & \textbf{0,73} & \textbf{0,96} & \textbf{0,42} & \textbf{0,64}
\end{tabular}
\end{table}

\begin{table}[]
\centering
\caption{Detailed results on the $60$ test videos of ``Monastero dei Benedettini'', considering the evaluation measure $ASF_1$ score. We used the backbone DenseNet. The ``/'' sign indicates that no samples from that class were present in the test video. The four classes are: 1) Antirefettorio, 2) Aula S. Mazzarino, 3) Cucina, 3) Ventre, whereas Neg. represents the negatives.}
\label{tab:monASF1Dense}
\scriptsize
\begin{tabular}{l|ccccc|c}
\textbf{ID\_visit} & \textbf{1} & \textbf{2} & \textbf{3} & \textbf{4} & \textbf{Neg.} & \textbf{AVG} \\ \hline
4805 & 0,55 & 0,62 & 0,4 & 0,12 & 0,36 & 0,41 \\
1804 & 0,32 & 0,1 & 0,39 & 0,05 & 0,21 & 0,214 \\
4377 & 0,31 & 0,12 & 0,26 & 0,08 & 0,27 & 0,208 \\
1669 & 0,68 & 0,34 & 0,67 & 0,9 & 0,33 & 0,584 \\
1791 & 0,41 & 0,41 & 0,25 & 0,06 & 0,22 & 0,27 \\
3948 & 0 & / & 0,33 & 0,26 & 0,49 & 0,27 \\
3152 & 0,39 & 0,43 & 0,64 & 0,44 & 0,58 & 0,496 \\
4361 & 0,26 & 0,35 & 0,32 & 0,08 & 0,2 & 0,242 \\
3976 & 0,68 & 0,26 & 0,35 & 0,03 & 0,5 & 0,364 \\
3527 & 0,65 & 0,37 & 0,38 & 0,28 & 0,44 & 0,424 \\
4105 & 0,49 & 0 & 0,49 & 0,16 & 0,06 & 0,24 \\
1399 & 0,55 & 0,1 & 0,33 & 0,23 & 0,29 & 0,3 \\
3836 & 0,47 & 0 & 0,35 & 0,98 & 0,34 & 0,428 \\
4006 & 0,47 & 0,49 & 0,49 & 0,55 & 0,26 & 0,452 \\
4415 & 0,74 & 0,07 & 0,35 & 0,14 & 0,23 & 0,306 \\
3008 & 0,57 & 0 & 0,5 & 0,99 & 0,41 & 0,494 \\
4660 & 0,45 & 0,52 & 0,17 & 0,05 & 0,27 & 0,292 \\
2826 & 0,31 & 0,29 & 0,33 & 0,06 & 0,25 & 0,248 \\
1099 & 0,52 & 0,24 & 0,55 & 0,96 & 0,17 & 0,488 \\
4391 & 0,54 & 0,13 & 0,14 & 0,2 & 0,24 & 0,25 \\
3929 & 0,15 & 0 & 0,31 & 0,22 & 0,39 & 0,214 \\
3362 & 0,2 & 0,35 & 0,38 & 0,05 & 0,22 & 0,24 \\
1379 & 0,22 & 0 & 0,43 & 0,05 & 0,35 & 0,21 \\
2600 & 0,32 & 0 & 0,39 & 0,31 & 0,26 & 0,256 \\
1430 & 0,43 & 0 & 0,1 & 0,02 & 0,21 & 0,152 \\
2956 & 0,41 & 0,2 & 0,16 & 0,29 & 0,31 & 0,274 \\
4742 & 0,15 & 0,18 & 0,25 & 0,11 & 0,17 & 0,172 \\
3651 & 0,32 & 0,25 & 0,74 & 0,48 & 0,28 & 0,414 \\
1064 & 0,63 & 0,09 & 0,52 & 0,97 & 0,19 & 0,48 \\
3818 & 0,43 & 0,49 & 0,2 & 0,06 & 0,29 & 0,294 \\ \hline
\end{tabular}
\end{table}

\begin{table}[]
\centering
\caption{Continued from \tablename~\ref{tab:monASF1Dense}}
\label{tab:monASF!Dense1}
\scriptsize
\begin{tabular}{lcccccc}
\multicolumn{1}{l|}{\textbf{ID\_Visit}} & \textbf{1} & \textbf{2} & \textbf{3} & \textbf{4} & \multicolumn{1}{c|}{\textbf{Neg.}} & \textbf{AVG} \\ \hline
\multicolumn{1}{l|}{2043} & 0,4 & 0,27 & 0,28 & 0,13 & \multicolumn{1}{c|}{0,27} & 0,27 \\
\multicolumn{1}{l|}{3996} & 0,23 & 0,3 & 0,22 & 0,05 & \multicolumn{1}{c|}{0,38} & 0,236 \\
\multicolumn{1}{l|}{3455} & 0,63 & 0,38 & 0,65 & 0,99 & \multicolumn{1}{c|}{0,48} & 0,626 \\
\multicolumn{1}{l|}{4785} & 0,06 & 0 & 0,25 & 0,45 & \multicolumn{1}{c|}{0,07} & 0,166 \\
\multicolumn{1}{l|}{2047} & 0,9 & 0,72 & 0,72 & 0,37 & \multicolumn{1}{c|}{0,34} & 0,61 \\
\multicolumn{1}{l|}{1912} & 0,42 & 0,25 & 0,2 & 0,08 & \multicolumn{1}{c|}{0,3} & 0,25 \\
\multicolumn{1}{l|}{3232} & 0,52 & 0,1 & 0,34 & 0,13 & \multicolumn{1}{c|}{0,35} & 0,288 \\
\multicolumn{1}{l|}{4442} & 0,61 & 0,22 & 0,33 & 0,06 & \multicolumn{1}{c|}{0,12} & 0,268 \\
\multicolumn{1}{l|}{3646} & 0,62 & 0,3 & 0,26 & 0,19 & \multicolumn{1}{c|}{0,22} & 0,318 \\
\multicolumn{1}{l|}{4833} & 0,54 & 0,55 & 0,26 & 0,1 & \multicolumn{1}{c|}{0,21} & 0,332 \\
\multicolumn{1}{l|}{3478} & 0,57 & 0,4 & 0,47 & 0,14 & \multicolumn{1}{c|}{0,24} & 0,364 \\
\multicolumn{1}{l|}{4396} & 0,41 & 0,1 & 0,2 & 0,07 & \multicolumn{1}{c|}{0,12} & 0,18 \\
\multicolumn{1}{l|}{2894} & 0,55 & 0,67 & 0,36 & 0,13 & \multicolumn{1}{c|}{0,31} & 0,404 \\
\multicolumn{1}{l|}{4414} & 0,59 & 0,26 & 0,83 & 0,11 & \multicolumn{1}{c|}{0,08} & 0,374 \\
\multicolumn{1}{l|}{4639} & 0,5 & 0,35 & 0,67 & 0,39 & \multicolumn{1}{c|}{0,27} & 0,436 \\
\multicolumn{1}{l|}{1004} & 0,22 & 0,47 & 0,17 & 0,11 & \multicolumn{1}{c|}{0,17} & 0,228 \\
\multicolumn{1}{l|}{1917} & 0,39 & 0,19 & 0,03 & 0,03 & \multicolumn{1}{c|}{0,21} & 0,17 \\
\multicolumn{1}{l|}{1153} & 0,47 & 0,33 & 0,21 & 0,14 & \multicolumn{1}{c|}{0,16} & 0,262 \\
\multicolumn{1}{l|}{2244} & 0,77 & 0,35 & 0,49 & 0,35 & \multicolumn{1}{c|}{0,32} & 0,456 \\
\multicolumn{1}{l|}{2614} & 0,43 & 0,06 & 0,39 & 0,99 & \multicolumn{1}{c|}{0,26} & 0,426 \\
\multicolumn{1}{l|}{1624} & 0,3 & 0,33 & 0,24 & 0,29 & \multicolumn{1}{c|}{0,29} & 0,29 \\
\multicolumn{1}{l|}{3441} & 0,49 & 0,11 & 0,44 & 0,88 & \multicolumn{1}{c|}{0,42} & 0,468 \\
\multicolumn{1}{l|}{4793} & 0,35 & / & 0,36 & 0,28 & \multicolumn{1}{c|}{0,37} & 0,34 \\
\multicolumn{1}{l|}{4083} & 0,88 & / & 0,41 & 0,25 & \multicolumn{1}{c|}{0,48} & 0,505 \\
\multicolumn{1}{l|}{4906} & 0,36 & 0,1 & 0,31 & 0,06 & \multicolumn{1}{c|}{0,21} & 0,208 \\
\multicolumn{1}{l|}{1160} & 0,79 & 0,5 & 0,39 & 0,14 & \multicolumn{1}{c|}{0,28} & 0,42 \\
\multicolumn{1}{l|}{3416} & 0,32 & / & 0,25 & 0,42 & \multicolumn{1}{c|}{0,34} & 0,3325 \\
\multicolumn{1}{l|}{1051} & 0,5 & 0,31 & 0,21 & 0,05 & \multicolumn{1}{c|}{0,4} & 0,294 \\
\multicolumn{1}{l|}{2580} & 0,36 & 0,29 & 0,33 & 0,05 & \multicolumn{1}{c|}{0,28} & 0,262 \\
\multicolumn{1}{l|}{1109} & 0,82 & 0,17 & 0,46 & 0,11 & \multicolumn{1}{c|}{0,28} & 0,368 \\ \hline
\multicolumn{1}{c}{mASF1} & 0,46 & 0,26 & 0,37 & 0,28 & 0,28 & \textbf{0,40}
\end{tabular}
\end{table}

\subsubsection{\textbf{Points of Interest Recognition}}

\paragraph{1) Palazzo Bellomo}
We used the subset of the Training set considered in the first task annotated with bounding box for a total of \textit{56686} frames. 
We split this subset in training/validation sets to train and validate the object detector. 
In particular, we used \textit{41111} frames as training set and \textit{15575} as validation set.
To find the optimal detection threshold, we used ``Test3" as validation video. We obtaining the value of \textit{0.05} maximizing mAP over the validations et. 
In Table~\ref{tab:bellomo_map} we report the results on the $9$ test videos (excluding ``Test3") of the object detector when using the mean Average Precision (mAP) as evaluation metric. 
The table also reports the number of frames annotated with bounding box for each test video.
Per-class AP values are reported in Table~\ref{tab:AP-class}.
\begin{table}[t]
\centering
\caption{Detailed results obtained using the YoloV3 object detector Yolov3 on the $9$ test videos of ``Palazzo Bellomo''. The last column reports the number of frames belonging to each test video. The last row indicates the average of mAP score obtained for each test video.}
\label{tab:bellomo_map}
\scriptsize
\begin{tabular}{lcc}
\multicolumn{1}{l|}{}                & \multicolumn{1}{l|}{\textbf{mAP}} & \multicolumn{1}{l}{\textbf{\#images}} \\ \hline
\multicolumn{1}{l|}{\textbf{Test1}}  & \multicolumn{1}{c|}{12,72}        & 1644                                  \\
\multicolumn{1}{l|}{\textbf{Test2}}  & \multicolumn{1}{c|}{13,61}        & 1238                                  \\
\multicolumn{1}{l|}{\textbf{Test4}}  & \multicolumn{1}{c|}{12,31}        & 1398                                  \\
\multicolumn{1}{l|}{\textbf{Test5}}  & \multicolumn{1}{c|}{8,65}         & 848                                   \\
\multicolumn{1}{l|}{\textbf{Test6}}  & \multicolumn{1}{c|}{8,9}          & 1453                                  \\
\multicolumn{1}{l|}{\textbf{Test7}}  & \multicolumn{1}{c|}{10,29}        & 1200                                  \\
\multicolumn{1}{l|}{\textbf{Test8}}  & \multicolumn{1}{c|}{10,98}        & 2826                                  \\
\multicolumn{1}{l|}{\textbf{Test9}}  & \multicolumn{1}{c|}{9,97}         & 2004                                  \\
\multicolumn{1}{l|}{\textbf{Test10}} & \multicolumn{1}{c|}{7,85}         & 791                                   \\ \hline
\textbf{AVG}                         & \textbf{10,59}                    & \textbf{13402}                       
\end{tabular}
\end{table}

\begin{table*}[]
\centering
\caption{Per-class AP values obtained on the 9 test videos. The ``/'' sign indicates that no samples from that class were present in the test videos.}
\label{tab:AP-class}
\scriptsize
\begin{tabular}{l|c|ll|c|ll|c|llc}
\textbf{Class} & \textbf{AP} &  & \textbf{Class} & \textbf{AP} & \multicolumn{1}{c}{\textbf{}} & \textbf{Class} & \textbf{AP} & \multicolumn{1}{c|}{\textbf{}} & \multicolumn{1}{l|}{\textbf{Class}} & \textbf{AP}          \\ \cline{1-2} \cline{4-5} \cline{7-8} \cline{10-11} 
0              & 10,53       &  & 50             & 6,67        &                               & 100            & 5,34        & \multicolumn{1}{l|}{}          & \multicolumn{1}{l|}{150}            & 0,07                 \\
1              & 41,45       &  & 51             & 0,00        &                               & 101            & 9,45        & \multicolumn{1}{l|}{}          & \multicolumn{1}{l|}{151}            & 2,18                 \\
2              & 49,60       &  & 52             & 0,60        &                               & 102            & 5,44        & \multicolumn{1}{l|}{}          & \multicolumn{1}{l|}{152}            & 17,34                \\
3              & 41,79       &  & 53             & 0,00        &                               & 103            & 10,69       & \multicolumn{1}{l|}{}          & \multicolumn{1}{l|}{153}            & 15,24                \\
4              & 13,27       &  & 54             & 0,00        &                               & 104            & 10,77       & \multicolumn{1}{l|}{}          & \multicolumn{1}{l|}{154}            & 35,10                \\
5              & 66,73       &  & 55             & 1,85        &                               & 105            & 2,71        & \multicolumn{1}{l|}{}          & \multicolumn{1}{l|}{155}            & 0,00                 \\
6              & 66,97       &  & 56             & 0,00        &                               & 106            & 2,67        & \multicolumn{1}{l|}{}          & \multicolumn{1}{l|}{156}            & 0,71                 \\
7              & 72,62       &  & 57             & 0,00        &                               & 107            & 0,00        & \multicolumn{1}{l|}{}          & \multicolumn{1}{l|}{157}            & 3,90                 \\
8              & 52,41       &  & 58             & 0,00        &                               & 108            & 3,80        & \multicolumn{1}{l|}{}          & \multicolumn{1}{l|}{158}            & 1,92                 \\
9              & 68,21       &  & 59             & 0,00        &                               & 109            & 6,11        & \multicolumn{1}{l|}{}          & \multicolumn{1}{l|}{159}            & 0,00                 \\
10             & 2,69        &  & 60             & 1,85        &                               & 110            & 16,50       & \multicolumn{1}{l|}{}          & \multicolumn{1}{l|}{160}            & 0,63                 \\
11             & 14,79       &  & 61             & 0,13        &                               & 111            & 0,00        & \multicolumn{1}{l|}{}          & \multicolumn{1}{l|}{161}            & 18,92                \\
12             & 2,19        &  & 62             & 6,05        &                               & 112            & 12,32       & \multicolumn{1}{l|}{}          & \multicolumn{1}{l|}{162}            & 11,44                \\
13             & 44,65       &  & 63             & 0,00        &                               & 113            & 0,00        & \multicolumn{1}{l|}{}          & \multicolumn{1}{l|}{163}            & 18,31                \\
14             & 35,27       &  & 64             & 2,34        &                               & 114            & 0,00        & \multicolumn{1}{l|}{}          & \multicolumn{1}{l|}{164}            & 12,00                \\
15             & 16,58       &  & 65             & 2,22        &                               & 115            & 21,75       & \multicolumn{1}{l|}{}          & \multicolumn{1}{l|}{165}            & 26,78                \\
16             & 61,05       &  & 66             & 0,74        &                               & 116            & 9,50        & \multicolumn{1}{l|}{}          & \multicolumn{1}{l|}{166}            & 11,97                \\
17             & 28,68       &  & 67             & 18,36       &                               & 117            & 4,98        & \multicolumn{1}{l|}{}          & \multicolumn{1}{l|}{167}            & 11,18                \\
18             & 46,37       &  & 68             & 9,19        &                               & 118            & 1,97        & \multicolumn{1}{l|}{}          & \multicolumn{1}{l|}{168}            & 1,04                 \\
19             & 9,68        &  & 69             & 5,70        &                               & 119            & 1,64        & \multicolumn{1}{l|}{}          & \multicolumn{1}{l|}{169}            & 23,41                \\
20             & 51,04       &  & 70             & 1,14        &                               & 120            & 33,36       & \multicolumn{1}{l|}{}          & \multicolumn{1}{l|}{170}            & 11,95                \\
21             & 11,11       &  & 71             & 2,64        &                               & 121            & 0,39        & \multicolumn{1}{l|}{}          & \multicolumn{1}{l|}{171}            & 0,52                 \\
22             & 45,00       &  & 72             & 9,19        &                               & 122            & 9,74        & \multicolumn{1}{l|}{}          & \multicolumn{1}{l|}{172}            & 2,82                 \\
23             & 48,80       &  & 73             & 11,26       &                               & 123            & 3,62        & \multicolumn{1}{l|}{}          & \multicolumn{1}{l|}{173}            & 5,12                 \\
24             & 10,40       &  & 74             & 0,11        &                               & 124            & 17,70       & \multicolumn{1}{l|}{}          & \multicolumn{1}{l|}{174}            & 37,13                \\
25             & 0,00        &  & 75             & 34,85       &                               & 125            & 0,00        & \multicolumn{1}{l|}{}          & \multicolumn{1}{l|}{175}            & 30,37                \\
26             & 17,47       &  & 76             & 6,56        &                               & 126            & 0,96        & \multicolumn{1}{l|}{}          & \multicolumn{1}{l|}{176}            & 18,87                \\
27             & 0,00        &  & 77             & 0,75        &                               & 127            & 1,05        & \multicolumn{1}{l|}{}          & \multicolumn{1}{l|}{177}            & /                    \\
28             & 14,01       &  & 78             & 10,02       &                               & 128            & 4,66        & \multicolumn{1}{l|}{}          & \multicolumn{1}{l|}{178}            & 0,00                 \\
29             & 0,00        &  & 79             & 5,16        &                               & 129            & 12,58       & \multicolumn{1}{l|}{}          & \multicolumn{1}{l|}{179}            & 0,00                 \\
30             & 3,11        &  & 80             & 16,57       &                               & 130            & 15,71       & \multicolumn{1}{l|}{}          & \multicolumn{1}{l|}{180}            & 3,43                 \\
31             & 16,71       &  & 81             & 17,89       &                               & 131            & 6,43        & \multicolumn{1}{l|}{}          & \multicolumn{1}{l|}{181}            & 0,00                 \\
32             & 0,61        &  & 82             & /           &                               & 132            & 2,74        & \multicolumn{1}{l|}{}          & \multicolumn{1}{l|}{182}            & 0,43                 \\
33             & 2,76        &  & 83             & 25,18       &                               & 133            & 0,00        & \multicolumn{1}{l|}{}          & \multicolumn{1}{l|}{183}            & 0,10                 \\
34             & 0,99        &  & 84             & 0,86        &                               & 134            & 0,00        & \multicolumn{1}{l|}{}          & \multicolumn{1}{l|}{184}            & 0,00                 \\
35             & 0,00        &  & 85             & 1,00        &                               & 135            & 0,00        & \multicolumn{1}{l|}{}          & \multicolumn{1}{l|}{185}            & 0,04                 \\
36             & 0,56        &  & 86             & 0,00        &                               & 136            & 5,02        & \multicolumn{1}{l|}{}          & \multicolumn{1}{l|}{186}            & 0,00                 \\
37             & 4,60        &  & 87             & 15,17       &                               & 137            & 4,49        & \multicolumn{1}{l|}{}          & \multicolumn{1}{l|}{187}            & 0,00                 \\
38             & 16,01       &  & 88             & 12,64       &                               & 138            & 0,30        & \multicolumn{1}{l|}{}          & \multicolumn{1}{l|}{188}            & 3,86                 \\
39             & 3,38        &  & 89             & 8,99        &                               & 139            & 0,00        & \multicolumn{1}{l|}{}          & \multicolumn{1}{l|}{189}            & 0,00                 \\
40             & 14,20       &  & 90             & 25,49       &                               & 140            & 0,83        & \multicolumn{1}{l|}{}          & \multicolumn{1}{l|}{190}            & 10,42                \\ \cline{10-11} 
41             & 0,00        &  & 91             & 0,29        &                               & 141            & 11,66       &                                & \textbf{mAP}                        & \textbf{10.66}       \\
42             & 0,00        &  & 92             & 0,00        &                               & 142            & 0,00        &                                &                                     & \multicolumn{1}{l}{} \\
43             & 25,33       &  & 93             & 8,84        &                               & 143            & 1,25        &                                &                                     & \multicolumn{1}{l}{} \\
44             & 0,11        &  & 94             & 11,98       &                               & 144            & 6,56        &                                &                                     & \multicolumn{1}{l}{} \\
45             & 0,00        &  & 95             & 4,79        &                               & 145            & 25,27       &                                &                                     & \multicolumn{1}{l}{} \\
46             & 0,00        &  & 96             & 0,00        &                               & 146            & 2,01        &                                &                                     & \multicolumn{1}{l}{} \\
47             & 0,00        &  & 97             & 0,00        &                               & 147            & 31,66       &                                &                                     & \multicolumn{1}{l}{} \\
48             & 20,07       &  & 98             & 0,00        &                               & 148            & 2,15        &                                &                                     & \multicolumn{1}{l}{} \\
49             & 12,90       &  & 99             & 0,00        &                               & 149            & 0,23        &                                &                                     & \multicolumn{1}{l}{} \\ \cline{1-2} \cline{4-5} \cline{7-8}
\end{tabular}
\end{table*}

\paragraph{2) Monastero dei Benedettini}
A  subset  of  frames  from  the dataset (sampled at at 1 fps) has been labeled with bounding boxes.
The annotated frames with bounding box of the Training set used in the first task are \textit{33366}. 
We split this set into training/validation sets to train and validate the object detector. 
In particular, we used \textit{23363} frames as training set and \textit{10003} as validation set.
We used the $10$ validation videos (\textit{2235} frames) to find the optimal threshold of the object detector. 
Table~\ref{tab:monvalmap} reports the AP values obtained for each of the $35$ considered points of interest belonging to the validation set, using the optimal threshold found through validation ($0.001$).
We annotated the $60$ real visits with bounding boxes for a total of \textit{71310} images. 
We tested the object detector on these frames. The results are shown in Table~\ref{tab:montestmap}.
Also, for each test video, we report the number of frames annotated with bounding boxes.
Per-class AP values are reported in Table~\ref{tab:APMonastero}. 

\begin{table}[t]
\centering
\caption{Per-class AP values obtained on the validation set using the optimal threshold of $0.001$.}
\label{tab:monvalmap}
\begin{tabular}{l|c}
\textbf{Class}                      & \textbf{AP}                        \\ \hline
5.1  PortaAulaS.Mazzarino           & 36,05                              \\
5.2  PortaIngressoMuseoFabbrica     & 37,99                              \\
5.3  PortaAntirefettorio            & 20,44                              \\
5.4  PortaIngressoRef.Piccolo       & 26,07                              \\
5.5 Cupola                          & 73,98                              \\
5.6  AperturaPavimento              & 80,52                              \\
5.7 S.Agata                         & 74,89                              \\
5.8  S.Scolastica                   & 66,84                              \\
6.1  QuadroSantoMazzarino           & 76,89                              \\
6.2 Affresco                        & 60,39                              \\
6.3  PavimentoOriginale             & 37,41                              \\
6.4  PavimentoRestaurato            & 13,11                              \\
6.5  BassorilieviMancanti           & 25,7                               \\
6.6 LavamaniSx                      & 41,55                              \\
6.7 LavamaniDx                      & 25,8                               \\
6.8  TavoloRelatori                 & 14,59                              \\
6.9 Poltrone                        & 23,48                              \\
7.1 Edicola                         & 42,57                              \\
7.2 PavimentoA                      & 6,03                               \\
7.3 PavimentoB                      & 0,93                               \\
7.4  PassavivandePavimentoOriginale & 44,44                              \\
7.5  AperturaPavimento              & 33,12                              \\
7.6 Scala                           & 46,49                              \\
7.7  SalaMetereologica              & 30,83                              \\
8.1 Doccione                        & 46,65                              \\
8.2  VanoRaccoltaCenere             & 75,6                               \\
8.3 SalaRossa                       & 18,52                              \\
8.4  ScalaCucina                    & 42,48                              \\
8.5  CucinaProvv.                   & 48,26                              \\
8.6 Ghiacciaia                      & 47,7                               \\
8.7 Latrina                         & 72,86                              \\
8.8  OssaeScarti                    & 50,36                              \\
8.9 Pozzo                           & 51,74                              \\
8.10 Cisterna                       & 17,03                              \\
8.11  BustoPietroTacchini           & 41,15                              \\
Negatives                           & 5,8                                \\ \hline
\textbf{mAP}                        & \multicolumn{1}{l}{\textbf{40.51}}
\end{tabular}
\end{table}

\begin{table}[t]
\centering
\caption{Detailed results of the YoloV3 object detector on the the $60$ real visits. The second column reports the number of frames annotated with bounding box contained in each visit.}
\label{tab:montestmap}
\begin{tabular}{lccllcc}
\multicolumn{1}{l|}{\textbf{ID\_Visit}} & \multicolumn{1}{c|}{\textbf{\#images}} & \multicolumn{1}{l}{\textbf{mAP}} &  & \multicolumn{1}{l|}{\textbf{ID\_Visit}} & \multicolumn{1}{l|}{\textbf{\#images}} & \multicolumn{1}{l}{\textbf{mAP}} \\ \cline{1-3} \cline{5-7} 
\multicolumn{1}{l|}{156}                & \multicolumn{1}{c|}{1288}              & 19,89                            &  & \multicolumn{1}{l|}{117}                & \multicolumn{1}{c|}{1349}              & 14,11                            \\
\multicolumn{1}{l|}{154}                & \multicolumn{1}{c|}{2443}              & 18,73                            &  & \multicolumn{1}{l|}{115}                & \multicolumn{1}{c|}{1511}              & 20,64                            \\
\multicolumn{1}{l|}{153}                & \multicolumn{1}{c|}{1620}              & 20,34                            &  & \multicolumn{1}{l|}{135}                & \multicolumn{1}{c|}{1396}              & 14,73                            \\
\multicolumn{1}{l|}{155}                & \multicolumn{1}{c|}{786}               & 18,58                            &  & \multicolumn{1}{l|}{137}                & \multicolumn{1}{c|}{2484}              & 12,04                            \\
\multicolumn{1}{l|}{110}                & \multicolumn{1}{c|}{1671}              & 18,36                            &  & \multicolumn{1}{l|}{136}                & \multicolumn{1}{c|}{566}               & 16,99                            \\
\multicolumn{1}{l|}{109}                & \multicolumn{1}{c|}{679}               & 19,96                            &  & \multicolumn{1}{l|}{132}                & \multicolumn{1}{c|}{1233}              & 17,74                            \\
\multicolumn{1}{l|}{108}                & \multicolumn{1}{c|}{1065}              & 13,84                            &  & \multicolumn{1}{l|}{134}                & \multicolumn{1}{c|}{1177}              & 18,41                            \\
\multicolumn{1}{l|}{107}                & \multicolumn{1}{c|}{1728}              & 15,24                            &  & \multicolumn{1}{l|}{130}                & \multicolumn{1}{c|}{1401}              & 15,54                            \\
\multicolumn{1}{l|}{158}                & \multicolumn{1}{c|}{1660}              & 13,37                            &  & \multicolumn{1}{l|}{105}                & \multicolumn{1}{c|}{1434}              & 15,04                            \\
\multicolumn{1}{l|}{157}                & \multicolumn{1}{c|}{874}               & 12,40                            &  & \multicolumn{1}{l|}{124}                & \multicolumn{1}{c|}{936}               & 20,68                            \\
\multicolumn{1}{l|}{160}                & \multicolumn{1}{c|}{660}               & 19,20                            &  & \multicolumn{1}{l|}{123}                & \multicolumn{1}{c|}{1571}              & 15,15                            \\
\multicolumn{1}{l|}{159}                & \multicolumn{1}{c|}{654}               & 16,08                            &  & \multicolumn{1}{l|}{103}                & \multicolumn{1}{c|}{2411}              & 11,65                            \\
\multicolumn{1}{l|}{129}                & \multicolumn{1}{c|}{751}               & 14,88                            &  & \multicolumn{1}{l|}{104}                & \multicolumn{1}{c|}{1492}              & 10,43                            \\
\multicolumn{1}{l|}{125}                & \multicolumn{1}{c|}{544}               & 19,51                            &  & \multicolumn{1}{l|}{122}                & \multicolumn{1}{c|}{1794}              & 14,36                            \\
\multicolumn{1}{l|}{126}                & \multicolumn{1}{c|}{892}               & 21,75                            &  & \multicolumn{1}{l|}{120}                & \multicolumn{1}{c|}{939}               & 15,88                            \\
\multicolumn{1}{l|}{163}                & \multicolumn{1}{c|}{683}               & 12,75                            &  & \multicolumn{1}{l|}{140}                & \multicolumn{1}{c|}{1050}              & 10,68                            \\
\multicolumn{1}{l|}{165}                & \multicolumn{1}{c|}{1689}              & 17,91                            &  & \multicolumn{1}{l|}{139}                & \multicolumn{1}{c|}{1454}              & 16,62                            \\
\multicolumn{1}{l|}{161}                & \multicolumn{1}{c|}{1563}              & 22,52                            &  & \multicolumn{1}{l|}{138}                & \multicolumn{1}{c|}{1736}              & 13,27                            \\
\multicolumn{1}{l|}{162}                & \multicolumn{1}{c|}{979}               & 20,02                            &  & \multicolumn{1}{l|}{145}                & \multicolumn{1}{c|}{1343}              & 11,25                            \\
\multicolumn{1}{l|}{166}                & \multicolumn{1}{c|}{597}               & 10,80                            &  & \multicolumn{1}{l|}{146}                & \multicolumn{1}{c|}{1370}              & 15,12                            \\
\multicolumn{1}{l|}{164}                & \multicolumn{1}{c|}{1197}              & 13,98                            &  & \multicolumn{1}{l|}{114}                & \multicolumn{1}{c|}{868}               & 11,45                            \\
\multicolumn{1}{l|}{142}                & \multicolumn{1}{c|}{1161}              & 17,27                            &  & \multicolumn{1}{l|}{112}                & \multicolumn{1}{c|}{840}               & 9,49                             \\
\multicolumn{1}{l|}{144}                & \multicolumn{1}{c|}{868}               & 11,61                            &  & \multicolumn{1}{l|}{111}                & \multicolumn{1}{c|}{726}               & 13,79                            \\
\multicolumn{1}{l|}{143}                & \multicolumn{1}{c|}{824}               & 9,64                             &  & \multicolumn{1}{l|}{113}                & \multicolumn{1}{c|}{851}               & 11,29                            \\
\multicolumn{1}{l|}{101}                & \multicolumn{1}{c|}{1894}              & 13,52                            &  & \multicolumn{1}{l|}{149}                & \multicolumn{1}{c|}{1612}              & 11,69                            \\
\multicolumn{1}{l|}{102}                & \multicolumn{1}{c|}{824}               & 10,50                            &  & \multicolumn{1}{l|}{148}                & \multicolumn{1}{c|}{847}               & 8,84                             \\
\multicolumn{1}{l|}{100}                & \multicolumn{1}{c|}{1343}              & 19,07                            &  & \multicolumn{1}{l|}{147}                & \multicolumn{1}{c|}{450}               & 19,92                            \\
\multicolumn{1}{l|}{119}                & \multicolumn{1}{c|}{564}               & 17,43                            &  & \multicolumn{1}{l|}{152}                & \multicolumn{1}{c|}{1519}              & 18,71                            \\
\multicolumn{1}{l|}{118}                & \multicolumn{1}{c|}{740}               & 16,74                            &  & \multicolumn{1}{l|}{150}                & \multicolumn{1}{c|}{1437}              & 13,82                            \\
\multicolumn{1}{l|}{116}                & \multicolumn{1}{c|}{1618}              & 16,55                            &  & \multicolumn{1}{l|}{151}                & \multicolumn{1}{c|}{942}               & 16,71                            \\ \cline{1-3} \cline{5-7} 
                                        &                                        &                                  &  & \textbf{Tot./AVG}                      & \textbf{71310}                         & \textbf{15,45}                  
\end{tabular}
\end{table}

\begin{table}[]
\centering
\caption{Per-class AP values obtained on the 60 real visits.}
\label{tab:APMonastero}
\begin{tabular}{l|c}
\textbf{Class}                      & \textbf{AP}                        \\ \hline
5.1  PortaAulaS.Mazzarino           & 37,86                              \\
5.2  PortaIngressoMuseoFabbrica     & 27,18                              \\
5.3  PortaAntirefettorio            & 2,23                               \\
5.4  PortaIngressoRef.Piccolo       & 15,44                              \\
5.5 Cupola                          & 65,80                              \\
5.6  AperturaPavimento              & 0,89                               \\
5.7 S.Agata                         & 41,82                              \\
5.8  S.Scolastica                   & 31,64                              \\
6.1  QuadroSantoMazzarino           & 13,22                              \\
6.2 Affresco                        & 53,69                              \\
6.3  PavimentoOriginale             & 6,95                               \\
6.4  PavimentoRestaurato            & 4,35                               \\
6.5  BassorilieviMancanti           & 16,57                              \\
6.6 LavamaniSx                      & 0,83                               \\
6.7 LavamaniDx                      & 0,58                               \\
6.8  TavoloRelatori                 & 4,99                               \\
6.9 Poltrone                        & 9,39                               \\
7.1 Edicola                         & 34,01                              \\
7.2 PavimentoA                      & 1,59                               \\
7.3 PavimentoB                      & 3,26                               \\
7.4  PassavivandePavimentoOriginale & 9,79                               \\
7.5  AperturaPavimento              & 22,77                              \\
7.6 Scala                           & 20,44                              \\
7.7  SalaMetereologica              & 11,71                              \\
8.1 Doccione                        & 13,78                              \\
8.2  VanoRaccoltaCenere             & 16,47                              \\
8.3 SalaRossa                       & 17,35                              \\
8.4  ScalaCucina                    & 13,36                              \\
8.5  CucinaProvv.                   & 16,33                              \\
8.6 Ghiacciaia                      & 3,98                               \\
8.7 Latrina                         & 22,29                              \\
8.8  OssaeScarti                    & 29,39                              \\
8.9 Pozzo                           & 13,43                              \\
8.10 Cisterna                       & 5,45                               \\
8.11  BustoPietroTacchini           & 23,13                              \\
Negatives                           & 16,81                              \\ \hline
\textbf{mAP}                        & \multicolumn{1}{l}{\textbf{17,47}}
\end{tabular}
\end{table}

\subsubsection{\textbf{Object Retrieval}}
To address this task, we extracted image patches from the bounding box annotations of the dataset. %

\paragraph{1) Palazzo Bellomo}
In Table~\ref{tab:bellomoretr}, we report the number of image patches which have been extracted for each test video. 
For one-shot learning, we have used reference images for training and all the image patches for testing, whereas to perform many-shot learning, we used the patches belonging to test videos $1-7$ for training (\textit{15185} patches) and the others to test (\textit{8542} patches). 

\begin{table}[t]
\centering
\caption{Number of patches extracted from each of the $10$ test videos of ``Palazzo Bellomo''.}
\label{tab:bellomoretr}
\begin{tabular}{lc}
\multicolumn{1}{l|}{\textbf{Test video}} & \textbf{\#images}                  \\ \hline
\multicolumn{1}{l|}{Test1}               & 2568                               \\
\multicolumn{1}{l|}{Test2}               & 2048                               \\
\multicolumn{1}{l|}{Test3}               & 2672                               \\
\multicolumn{1}{l|}{Test4}               & 2224                               \\
\multicolumn{1}{l|}{Test5}               & 1439                               \\
\multicolumn{1}{l|}{Test6}               & 2086                               \\
\multicolumn{1}{l|}{Test7}               & 2148                               \\
\multicolumn{1}{l|}{Test8}               & 4108                               \\
\multicolumn{1}{l|}{Test9}               & 2914                               \\
\multicolumn{1}{l|}{Test10}              & 1520                               \\ \hline
\textbf{Total}                           & \multicolumn{1}{l}{\textbf{23727}}
\end{tabular}
\end{table}

\paragraph{2) Monastero dei Benedettini}
In Table~\ref{tab:monretrieval}, we report the number of image patches extracted from the $60$ test videos. 
One-shot learning has been performed using reference images from training and all extracted patches for testing. 
For many-shot learning, we used \textit{30497} image patches images belonging to the visits with IDs from $100$ to $147$ for training, and \textit{14551} patches belonging to the visits with ID from $148$ to $166$ for testing.

\begin{table}[t]
\centering
\caption{Number of image patches extracted from the $60$ test videos of ``Monastero dei Benedettini''.}
\label{tab:monretrieval}
\begin{tabular}{lcllc}
\multicolumn{1}{l|}{\textbf{ID\_Visit}} & \multicolumn{1}{c|}{\textbf{\#images}} & \multicolumn{1}{l|}{} & \multicolumn{1}{l|}{\textbf{ID\_Visit}} & \multicolumn{1}{l}{\textbf{\#images}} \\ \cline{1-2} \cline{4-5} 
\multicolumn{1}{l|}{100}                & \multicolumn{1}{c|}{770}               & \multicolumn{1}{l|}{} & \multicolumn{1}{l|}{135}                & 765                                   \\
\multicolumn{1}{l|}{101}                & \multicolumn{1}{c|}{696}               & \multicolumn{1}{l|}{} & \multicolumn{1}{l|}{136}                & 414                                   \\
\multicolumn{1}{l|}{102}                & \multicolumn{1}{c|}{613}               & \multicolumn{1}{l|}{} & \multicolumn{1}{l|}{137}                & 354                                   \\
\multicolumn{1}{l|}{103}                & \multicolumn{1}{c|}{1733}              & \multicolumn{1}{l|}{} & \multicolumn{1}{l|}{138}                & 824                                   \\
\multicolumn{1}{l|}{104}                & \multicolumn{1}{c|}{768}               & \multicolumn{1}{l|}{} & \multicolumn{1}{l|}{139}                & 707                                   \\
\multicolumn{1}{l|}{105}                & \multicolumn{1}{c|}{929}               & \multicolumn{1}{l|}{} & \multicolumn{1}{l|}{140}                & 494                                   \\
\multicolumn{1}{l|}{107}                & \multicolumn{1}{c|}{1011}              & \multicolumn{1}{l|}{} & \multicolumn{1}{l|}{142}                & 770                                   \\
\multicolumn{1}{l|}{108}                & \multicolumn{1}{c|}{659}               & \multicolumn{1}{l|}{} & \multicolumn{1}{l|}{143}                & 536                                   \\
\multicolumn{1}{l|}{109}                & \multicolumn{1}{c|}{234}               & \multicolumn{1}{l|}{} & \multicolumn{1}{l|}{144}                & 598                                   \\
\multicolumn{1}{l|}{110}                & \multicolumn{1}{c|}{918}               & \multicolumn{1}{l|}{} & \multicolumn{1}{l|}{145}                & 897                                   \\
\multicolumn{1}{l|}{111}                & \multicolumn{1}{c|}{365}               & \multicolumn{1}{l|}{} & \multicolumn{1}{l|}{146}                & 1307                                  \\
\multicolumn{1}{l|}{112}                & \multicolumn{1}{c|}{727}               & \multicolumn{1}{l|}{} & \multicolumn{1}{l|}{147}                & 173                                   \\
\multicolumn{1}{l|}{113}                & \multicolumn{1}{c|}{288}               & \multicolumn{1}{l|}{} & \multicolumn{1}{l|}{148}                & 692                                   \\
\multicolumn{1}{l|}{114}                & \multicolumn{1}{c|}{561}               & \multicolumn{1}{l|}{} & \multicolumn{1}{l|}{149}                & 954                                   \\
\multicolumn{1}{l|}{115}                & \multicolumn{1}{c|}{623}               & \multicolumn{1}{l|}{} & \multicolumn{1}{l|}{150}                & 609                                   \\
\multicolumn{1}{l|}{116}                & \multicolumn{1}{c|}{968}               & \multicolumn{1}{l|}{} & \multicolumn{1}{l|}{151}                & 652                                   \\
\multicolumn{1}{l|}{117}                & \multicolumn{1}{c|}{810}               & \multicolumn{1}{l|}{} & \multicolumn{1}{l|}{152}                & 699                                   \\
\multicolumn{1}{l|}{118}                & \multicolumn{1}{c|}{907}               & \multicolumn{1}{l|}{} & \multicolumn{1}{l|}{153}                & 1244                                  \\
\multicolumn{1}{l|}{119}                & \multicolumn{1}{c|}{545}               & \multicolumn{1}{l|}{} & \multicolumn{1}{l|}{154}                & 1691                                  \\
\multicolumn{1}{l|}{120}                & \multicolumn{1}{c|}{669}               & \multicolumn{1}{l|}{} & \multicolumn{1}{l|}{155}                & 666                                   \\
\multicolumn{1}{l|}{121}                & \multicolumn{1}{c|}{774}               & \multicolumn{1}{l|}{} & \multicolumn{1}{l|}{156}                & 709                                   \\
\multicolumn{1}{l|}{122}                & \multicolumn{1}{c|}{1156}              & \multicolumn{1}{l|}{} & \multicolumn{1}{l|}{157}                & 515                                   \\
\multicolumn{1}{l|}{123}                & \multicolumn{1}{c|}{957}               & \multicolumn{1}{l|}{} & \multicolumn{1}{l|}{158}                & 846                                   \\
\multicolumn{1}{l|}{124}                & \multicolumn{1}{c|}{652}               & \multicolumn{1}{l|}{} & \multicolumn{1}{l|}{159}                & 544                                   \\
\multicolumn{1}{l|}{125}                & \multicolumn{1}{c|}{491}               & \multicolumn{1}{l|}{} & \multicolumn{1}{l|}{160}                & 327                                   \\
\multicolumn{1}{l|}{126}                & \multicolumn{1}{c|}{702}               & \multicolumn{1}{l|}{} & \multicolumn{1}{l|}{161}                & 985                                   \\
\multicolumn{1}{l|}{129}                & \multicolumn{1}{c|}{587}               & \multicolumn{1}{l|}{} & \multicolumn{1}{l|}{162}                & 902                                   \\
\multicolumn{1}{l|}{130}                & \multicolumn{1}{c|}{820}               & \multicolumn{1}{l|}{} & \multicolumn{1}{l|}{164}                & 693                                   \\
\multicolumn{1}{l|}{132}                & \multicolumn{1}{c|}{771}               & \multicolumn{1}{l|}{} & \multicolumn{1}{l|}{165}                & 1133                                  \\
\multicolumn{1}{l|}{134}                & \multicolumn{1}{c|}{884}               & \multicolumn{1}{l|}{} & \multicolumn{1}{l|}{166}                & 690                                   \\ \cline{1-2} \cline{4-5} 
                                        & \multicolumn{1}{l}{}                   &                       & \multicolumn{1}{c}{\textbf{Total}}      & \textbf{44978}                       
\end{tabular}
\end{table}

\paragraph{Object Retrieval with DenseNet}
We tried to extract features using a different backbone respect to the proposed baseline based on VGG-19. We extracted the features from the FC7 layer of DenseNet\cite{huang2017densely}. In this way, we obtained for each image a fixed-size vector of 1024 values. We have followed the same pipeline used with VGG to perform object retrieval. Table~\ref{tab:retr_Dense} shows the results obtained using DenseNet as backbone. The results show that using DenseNet we didn't obtained an improvement considering the $F_1$ score as evaluation measure in both variants (One-shot and Many-shots) for both cultural sites.

\begin{table}[]
\centering
\caption{Results using Densenet.}
\label{tab:retr_Dense}
\scriptsize
\begin{tabular}{ccccc}
\multicolumn{5}{c}{\textbf{Points of Interest Retrieval}} \\ \hline
\multicolumn{1}{l}{} &  & \multicolumn{1}{l}{} & \multicolumn{1}{l}{} & \multicolumn{1}{l}{} \\
\multicolumn{5}{c}{\textbf{1) Palazzo Bellomo}} \\ \hline
\multicolumn{1}{c|}{\textbf{Variant}} & \multicolumn{1}{c|}{\textbf{K}} & \multicolumn{1}{c|}{\textbf{Precision}} & \multicolumn{1}{c|}{\textbf{Recall}} & \textbf{F1 score} \\ \hline
\multicolumn{1}{c|}{1 - One Shot} & \multicolumn{1}{c|}{1} & \multicolumn{1}{c|}{0.02} & \multicolumn{1}{c|}{0.01} & 0.00 \\ \hline
\multicolumn{1}{c|}{} & \multicolumn{1}{c|}{1} & \multicolumn{1}{c|}{\textbf{0.62}} & \multicolumn{1}{c|}{\textbf{0.59}} & \textbf{0.6} \\
\multicolumn{1}{c|}{} & \multicolumn{1}{c|}{3} & \multicolumn{1}{c|}{0.62} & \multicolumn{1}{c|}{0.56} & 0.56 \\
\multicolumn{1}{c|}{2 - Many Shots} & \multicolumn{1}{c|}{5} & \multicolumn{1}{c|}{0.62} & \multicolumn{1}{c|}{0.56} & 0.56 \\
\multicolumn{1}{l|}{} & \multicolumn{1}{c|}{7} & \multicolumn{1}{c|}{0,61} & \multicolumn{1}{c|}{0,56} & 0,56 \\
\multicolumn{1}{l|}{} & \multicolumn{1}{c|}{9} & \multicolumn{1}{c|}{0,61} & \multicolumn{1}{c|}{0,55} & 0,56 \\
\multicolumn{1}{l|}{} & \multicolumn{1}{c|}{11} & \multicolumn{1}{c|}{0,61} & \multicolumn{1}{c|}{0,55} & 0,55 \\
\multicolumn{1}{l}{} & \multicolumn{1}{l}{} & \multicolumn{1}{l}{} & \multicolumn{1}{l}{} & \multicolumn{1}{l}{} \\
\multicolumn{5}{c}{\textbf{2) Monastero dei Benedettini}} \\ \hline
\multicolumn{1}{c|}{\textbf{Variant}} & \multicolumn{1}{c|}{\textbf{K}} & \multicolumn{1}{c|}{\textbf{Precision}} & \multicolumn{1}{c|}{\textbf{Recall}} & \textbf{F1 score} \\ \hline
\multicolumn{1}{l|}{1 - One shot} & \multicolumn{1}{l|}{1} & \multicolumn{1}{c|}{0.38} & \multicolumn{1}{c|}{0.07} & 0.09 \\ \hline
\multicolumn{1}{c|}{} & \multicolumn{1}{c|}{1} & \multicolumn{1}{c|}{0,83} & \multicolumn{1}{c|}{0,83} & 0,83 \\
\multicolumn{1}{c|}{} & \multicolumn{1}{c|}{3} & \multicolumn{1}{c|}{0,84} & \multicolumn{1}{c|}{0,83} & 0,83 \\
\multicolumn{1}{c|}{2 - Many Shots} & \multicolumn{1}{c|}{5} & \multicolumn{1}{c|}{\textbf{0,84}} & \multicolumn{1}{c|}{\textbf{0,84}} & \textbf{0,83} \\
\multicolumn{1}{l|}{} & \multicolumn{1}{c|}{7} & \multicolumn{1}{c|}{0,84} & \multicolumn{1}{c|}{0,83} & 0,83 \\
\multicolumn{1}{l|}{} & \multicolumn{1}{c|}{9} & \multicolumn{1}{c|}{0,84} & \multicolumn{1}{c|}{0,83} & 0,83 \\
\multicolumn{1}{l|}{} & \multicolumn{1}{c|}{11} & \multicolumn{1}{c|}{0,83} & \multicolumn{1}{c|}{0,83} & 0,82
\end{tabular}
\end{table}

\subsubsection{\textbf{Survey Prediction}}
Table~\ref{tab:KNN} shows the results obtained on binary classification using a KNN with different values of \textit{k}. The method is evaluated using precision, recall and $F_1$ score. $K=9$ gives the best results.
Table~\ref{tab:KNN_multi} shows the results of multi-class classification using a KNN classifier with different values of $K$. We obtained the best results for $K=9$.

\begin{table}[t]
\centering
\caption{Results of the binary classifier obtained using a KNN with different values of $K$.}
\label{tab:KNN}
\begin{tabular}{l|c|c|c|c}
\textbf{K} & \multicolumn{1}{l|}{\textbf{Precision}} & \multicolumn{1}{l|}{\textbf{Recall}} & \multicolumn{1}{l|}{\textbf{F1 score}} & \multicolumn{1}{l}{\textbf{Support}} \\ \hline
\textbf{1} & 0,62                                    & 0,61                                 & 0,62                                   & \multirow{5}{*}{1980}                \\
\textbf{3} & 0,62                                    & 0,65                                 & 0,63                                   &                                      \\
\textbf{5} & 0,63                                    & 0,67                                 & 0,64                                   &                                      \\
\textbf{7} & 0,64                                    & 0,69                                 & 0,65                                   &                                      \\
\textbf{9} & \textbf{0,65}                           & \textbf{0,7}                         & \textbf{0,66}                          &                                     
\end{tabular}
\end{table}

\begin{table}[t]
\centering
\caption{Results of the multi-class classifier obtained using a KNN with different values of $K$.}
\label{tab:KNN_multi}
\begin{tabular}{l|c|c|c|c}
\textbf{K} & \multicolumn{1}{l|}{\textbf{Precision}} & \multicolumn{1}{l|}{\textbf{Recall}} & \multicolumn{1}{l|}{\textbf{F1 score}} & \multicolumn{1}{l}{\textbf{Support}} \\ \hline
\textbf{1} & 0,2                                     & 0,2                                  & 0,2                                    & \multirow{5}{*}{1980}                \\
\textbf{3} & 0,2                                     & 0,23                                 & 0,19                                   &                                      \\
\textbf{5} & 0,2                                     & 0,24                                 & 0,21                                   &                                      \\
\textbf{7} & 0,22                                    & 0,24                                 & 0,22                                   &                                      \\
\textbf{9} & \textbf{0,23}                           & \textbf{0,27}                        & \textbf{0,23}                          &                                     
\end{tabular}
\end{table}

\end{document}